\documentclass{article} 
\usepackage{iclr2025_conference,times}

\usepackage{amsmath,amsfonts,bm}

\def\eqref#1{equation~\ref{#1}}

\def\1{\bm{1}}

\DeclareMathAlphabet{\mathsfit}{\encodingdefault}{\sfdefault}{m}{sl}
\SetMathAlphabet{\mathsfit}{bold}{\encodingdefault}{\sfdefault}{bx}{n}

\usepackage[utf8]{inputenc} 
\usepackage[T1]{fontenc}    
\usepackage{hyperref}       
\usepackage{url}            
\usepackage{booktabs}       
\usepackage{amsfonts}       
\usepackage{nicefrac}       
\usepackage{microtype}      
\usepackage{xcolor}         
\usepackage{amsmath}
\usepackage{amssymb}
\usepackage{subcaption}
\usepackage{bm}
\usepackage{leftindex}
\usepackage{graphicx}
\title{Analyzing Neural Scaling Laws in Two-Layer Networks with Power-Law Data Spectra}

\author{Roman Worschech \\
Institut für Theoretische Physik \\
Universität Leipzig\\
Brüderstraße 16, 04103 Leipzig, Germany \\[5pt]
Max Planck Institute for Mathematics in the Sciences \\
Inselstraße 22, 04103 Leipzig, Germany \\
\texttt{roman.worschech@uni-leipzig.de}
\And
Bernd Rosenow \\
Institut für Theoretische Physik \\
Universität Leipzig \\
Brüderstraße 16, 04103 Leipzig, Germany \\
}

\iclrfinalcopy 
\begin{document}

\maketitle

\begin{abstract}

Neural scaling laws describe how the performance of deep neural networks scales with key factors such as training data size, model complexity, and training time, often following power-law behaviors over multiple orders of magnitude. Despite their empirical observation, the theoretical understanding of these scaling laws remains limited.  In this work, we employ techniques from statistical mechanics to analyze one-pass stochastic gradient descent within a student-teacher framework, where both the student and teacher are two-layer neural networks. Our study primarily focuses on the generalization error and its behavior in response to data covariance matrices that exhibit power-law spectra. 
For linear activation functions, we derive analytical expressions for the generalization error, exploring different learning regimes and identifying conditions under which power-law scaling emerges. Additionally, we extend our analysis to non-linear activation functions in the feature learning regime, investigating how power-law spectra in the data covariance matrix impact learning dynamics. Importantly, we find that the length of the symmetric plateau depends on the number of distinct eigenvalues of the data covariance matrix and the number of hidden units, demonstrating how these plateaus behave under various configurations. In addition, our results reveal a transition from exponential to power-law convergence in the specialized phase when the data covariance matrix possesses a power-law spectrum. This work contributes to the theoretical understanding of neural scaling laws and provides insights into optimizing learning performance in practical scenarios involving complex data structures.
\end{abstract}

\section{Introduction}

Recent empirical studies have revealed that the performance of state-of-the-art deep neural networks, trained on large-scale real-world data, can be predicted by simple phenomenological functions \cite{hestness2017deep,kaplan2020scaling,porian2024resolvingdiscrepanciescomputeoptimalscaling}. Specifically, the network's error decreases in a power-law fashion with respect to the number of training examples, model size, or training time, spanning many orders of magnitude. 
This observed phenomenon is encapsulated by neural scaling laws, which describe how neural network performance varies as key scaling factors change. Interestingly, the performance improvement due to one scaling factor is often limited by another, suggesting the presence of bottleneck effects. 
Understanding these scaling laws theoretically is crucial for practical applications such as optimizing architectural design and selecting appropriate hyperparameters. However, the fundamental reasons behind the emergence of neural scaling laws have mainly been explored in the context of linear models \cite{lin2024} and random feature models \cite{bahri2021explaining}, and a more comprehensive theoretical framework is still absent. \\

\textbf{Scope of Study}. In this work, we employ techniques from statistical mechanics to analyze one-pass stochastic gradient descent within a student-teacher framework. Both networks are two-layer neural networks: the student has $K$ hidden neurons, the teacher has $M$, and we train only the student's input-to-hidden weights, realizing a so-called committee machine \cite{M_Biehl_1995}. We begin our analysis with linear activation functions for both networks and then extend it to non-linear activation functions, focusing on the feature learning regime where the student weights undergo significant changes during training. Our primary focus is on analyzing the generalization error $\epsilon_g$ by introducing order parameters that elucidate the relationships between the student and teacher weights. Despite the diversity of datasets across various learning domains, a critical commonality is that their feature-feature covariance matrices often exhibit power-law spectra \cite{maloney2022solvable}. To model realistic data, we therefore utilize Gaussian-distributed inputs with covariance matrices that display power-law spectra. \\

\begin{figure}
  \centering
  \includegraphics[width=0.5\linewidth]{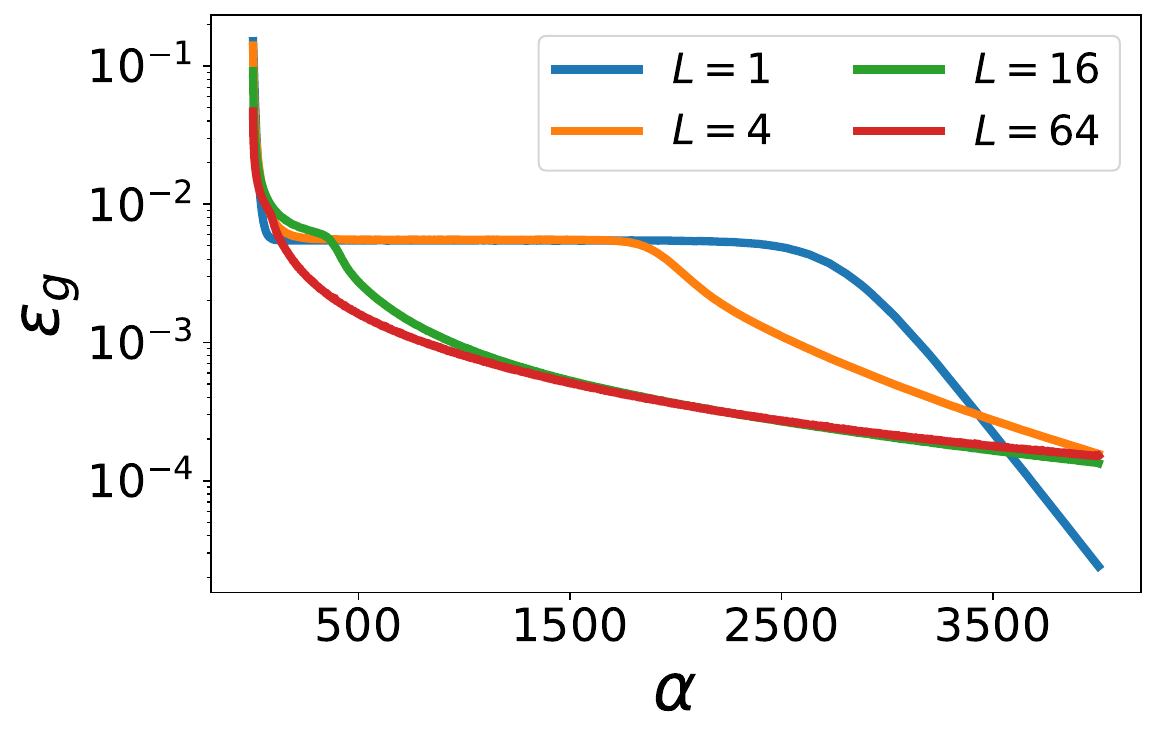}
  \caption{ Generalization error $\epsilon_g$ as a function of $\alpha$ for $K = M = 2$, $\eta = 0.1$, $\beta = 1$, $\sigma_J = 0.01$, and $N = 1024$, with varying numbers of distinct eigenvalues $L$. As $L$ increases, the plateau length decreases until it disappears. Additionally, with increasing $L$, the convergence of the asymptotic generalization error slows down, transitioning from exponential to power-law scaling in the early asymptotic phase.}
  \label{fig: emerging power-law}
\end{figure}

\textbf{White Noise vs.\ Power-Law Spectra}. \hspace{3pt} The student-teacher setup with isotropic input data has been extensively studied and is well-understood in the literature \cite{PhysRevE.52.4225}. In the realizable scenario where $K=M$, the generalization error typically undergoes three distinct phases: a rapid learning phase, a plateau phase, and an exponentially decaying phase with time $\alpha$. Introducing a power-law spectrum in the covariance matrix leads to observable changes in the plateau's height and duration, along with a slowdown in the convergence towards zero generalization error.
Notably, as the number of distinct eigenvalues $L$ in the data covariance spectrum increases, the plateau shortens, and the convergence to perfect learning becomes progressively slower, as depicted in Figure \ref{fig: emerging power-law}. This observation indicates a potential transition from exponential decay to power-law scaling in the generalization error over time. Identifying and understanding this transition is a critical focus of our investigation.
Our main \textbf{contributions} are:
\begin{itemize}
    \item For linear activation functions, we derive an exact analytical expression for the generalization error as a function of training time $\alpha$ and the power-law exponent $\beta$ of the covariance matrix. We characterize different learning regimes for the generalization error and analyze the conditions under which power-law scaling emerges.

 \item In addition, for linear activation functions, we demonstrate a scaling law in the number of trainable student parameters, effectively reducing the input dimension of the network. This power-law is different from the power-law characterizing the training time dependence.
 
 \item We derive an analytical formula for the dependence of the plateau length on the number of distinct eigenvalues and the power-law exponent $\beta$ of the covariance matrix, illustrating how these plateaus behave under different configurations.
 
\item We investigate the asymptotic learning regime for non-linear activation functions and find that, in the realizable case with $M=K$, the convergence to perfect learning shifts from an exponential to a power-law regime when the data covariance matrix has a power-law spectrum.

\end{itemize}

\section{Related work}
\textbf{Theory of Neural Scaling Laws for Linear Activation Functions}. \hspace{3pt} Previous studies on neural scaling laws have primarily focused on random feature models or linear (ridge) regression with power-law features \cite{wei22}. In particular, \cite{maloney2022solvable,paquette20244+} and \cite{atanasov2024} analyzed random feature models for linear features and ridge regression, employing techniques from random matrix theory. \cite{bahri2021explaining} examined random feature models for kernel ridge regression within a student-teacher framework using techniques from statistical mechanics. In their analysis, either the number of parameters or the training dataset size was considered infinite, leading to scaling laws in the test loss with respect to the remaining finite quantity. \cite{bordelon24} studied random feature models with randomly projected features and momentum, trained using gradient flow. Using a dynamical mean field theory approach, they derived a "bottleneck scaling" where only one of time, dataset size, or model size is finite while the other two quantities approach infinity. Additionally, \cite{hutter2021} investigated a binary toy model and found non-trivial scaling laws with respect to the number of training examples. \\

\cite{bordelon2022learning} studied one-pass stochastic gradient descent for random feature models, deriving a scaling law for the test error over time in the small learning rate regime. Similarly, \cite{lin2024} investigated infinite-dimensional linear regression under one-pass stochastic gradient descent, providing insights through a statistical learning theory framework. They derived upper and lower bounds for the test error, demonstrating scaling laws with respect to the number of parameters and dataset size under different scaling exponents. \\

 Building upon the work of \cite{lin2024} and \cite{bordelon2022learning}, we also consider one-pass stochastic gradient descent.   However, our study extends to both linear and non-linear neural networks where we train the weights used in the pre-activations (i.e., feature learning), and use fixed hidden-to-output connections. Unlike \cite{bordelon2022learning}, we extend the analysis for linear activation functions to general learning rates and varying numbers of input neurons.  
Additionally, we derive upper and lower bounds for the time interval over which the generalization error exhibits power-law behavior. A significant difference from previous works is our focus on feature learning, where all pre-activation weights are trainable. In this regime, certain groups of student weights, organized by student vectors, begin to imitate teacher vectors during the late training phase, leading to specialization.

Other theoretical studies have explored different aspects of scaling laws. Some have focused on learnable network skills or abilities that drive the decay of the loss \cite{arora2023,michaud2023,caballero2023,nam2024}. Others have compared the influence of synthetic data with real data \cite{jain2024} or investigated model collapse phenomena \cite{dohmatob2024, dohmatob2024model}. 
Further works studying correlated and realistic input data are \cite{PhysRevX.10.041044,loureiro2021learning,PhysRevX.14.031001,cagnetta2024theorystructurelanguageacquired}.
 
\textbf{Statistical Mechanics Approach}. \hspace{3pt} Analytical studies using the statistical mechanics framework for online learning have traditionally focused on uncorrelated input data or white noise.   \cite{PhysRevE.52.4225} first introduced differential equations for two-layer neural networks trained via stochastic gradient descent on such data. Building upon this, \cite{NEURIPS2019_287e03db} recently expanded these models to include Gaussian-correlated input patterns, deriving a set of closed-form differential equations.
  Their research primarily involved numerically solving these equations for covariance matrices with up to two distinct eigenvalues, exploring how the magnitudes of the eigenvalues affect the plateau's length and height. In our study, we extend this hierarchy of differential equations to investigate the dynamics of order parameters for data covariance matrices with power-law spectra, considering $L$ distinct eigenvalues.

\section{Setup}
\label{Setup section}
\textbf{Dataset}. \hspace{3pt} We consider a student network trained on outputs generated by a teacher network, using $p$ input examples $\boldsymbol{\xi}^\mu \in \mathbb{R}^N$, where $\mu = 1, \dots, p$. Each input $\boldsymbol{\xi}^\mu$ is drawn from a correlated Gaussian distribution $\mathcal{N}(0, \bm{\Sigma})$, with covariance matrix $\bm{\Sigma} \in \mathbb{R}^{N \times N}$.  Although the covariance matrix generally has $N$ eigenvalues, we assume it has only $L$ distinct eigenvalues, each occurring with multiplicity $N/L$, where $1 \leq L \leq N$ and $N/L$ is an integer. The eigenvalues follow a power-law distribution:
\begin{equation}
\lambda_l= \frac{\lambda_{+}}{l^{1+\beta}}
\label{data generating model}
\end{equation}
where $\beta > 0$ is the power-law exponent of the covariance matrix, $\lambda_{+} = \lambda_1$ is the largest eigenvalue, and $l \in \{1, \dots, L\}$. We choose $\lambda_{+}$ such that the total variance satisfies $\sum_{l=1}^L \left( \frac{N}{L} \lambda_l \right) = N$, ensuring that the pre-activations of the hidden neurons remain of order one in our setup.

\textbf{Student-Teacher Setup}. \hspace{3pt} The student is a soft committee machine -- a two-layer neural network with an input layer of $N$ neurons, a hidden layer of $K$ neurons, and an output layer with a single neuron.
 In the statistical mechanics framework, we represent the weights between the input layer and the hidden layer as vectors. Specifically, the connection between the input layer and the $i$-th hidden neuron is represented by the student vector $\boldsymbol{J}_i \in \mathbb{R}^N$.
Thus, we have $K$ student vectors $\boldsymbol{J}_i$, each representing the weights connecting the entire input layer to one of the hidden neurons. The pre-activation received by the $i$-th hidden neuron is defined  as $x_i = \frac{1}{\sqrt{N}} \boldsymbol{\xi}^\mu \cdot \boldsymbol{J}_i$. 
The overall output of the student is given by
\begin{equation}
\sigma(\boldsymbol{J},\boldsymbol{\xi})=\frac{\sqrt{M}}{K}\sum_{i=1}^{K} g\left(x_i\right),
\label{output student}
\end{equation}
where $g(x_i)$ is the activation function, and the output weights are set to $\sqrt{M}/K$. In this setup, we train the student vectors $\boldsymbol{J}_i$ and keep the hidden-to-output weights fixed. The teacher network has the same architecture but with $M$ hidden neurons, and its weights are characterized by the teacher vectors $\boldsymbol{B}_n \in \mathbb{R}^N$. The pre-activations for the teacher are
$y_n = \frac{1}{\sqrt{N}} \boldsymbol{\xi}^\mu \cdot \boldsymbol{B}_n$, and its overall output is $\zeta(\boldsymbol{B}, \boldsymbol{\xi}) = \sum_{n=1}^{M} g\left(y_n\right)$. We initialize the student and teacher vectors from normal distributions: $J_{ia} \sim \mathcal{N}(0, \sigma_J^2)$ and $B_{na} \sim \mathcal{N}(0, 1)$, where $\sigma_J^2$ is the variance of the student weights and $a \in \{1, \dots, N\}$. To quantify the discrepancy between the student's output and the teacher's output, we use the squared loss function $\epsilon = \frac{1}{2} [\zeta - \sigma]^2$. Our main focus is the generalization error $\epsilon_g = \left< \epsilon(\boldsymbol{\xi}) \right>_{\boldsymbol{\xi}}$, which measures the typical error of the student on new inputs. Throughout this work, we consider the error function as our non-linear activation function $g\hspace{-2pt}\left(x\right)= \mathrm{erf}\hspace{-2pt}\left({\frac{x}{\sqrt{2}}}\right)$.

\textbf{Transition from Microscopic to Macroscopic Formalism}. \hspace{3pt} Rather than computing expectation values directly over the input distribution, we consider higher-order pre-activations defined as $x_i^{(l)}=\bm{\xi}^\mu \left( \bm{\Sigma}\right)^{l}\bm{J}_i/\sqrt{N} $ and $y_n^{(l)}=\bm{\xi}^\mu \left( \bm{\Sigma}\right)^{l} \bm{B}_n/\sqrt{N}$, as suggested in \cite{NEURIPS2019_287e03db}.
Here, $\left( \bm{\Sigma} \right)^l$ denotes the $l$-th power of the covariance matrix, and we define $\left( \bm{\Sigma} \right)^0 = \bm{I}$. In the thermodynamic limit $N \rightarrow \infty$, these higher-order pre-activations become Gaussian random variables with zero mean and covariances given by: $\left<x_i^{(l)} x_j^{(l)} \right> =\frac{\boldsymbol{J}_i\left( \bm{\Sigma}\right)^{l}\boldsymbol{J}_j}{N} :=Q_{ij}^{(l)} $, $\left<x_i^{(l)} y_n^{(l)}\right>=\frac{\boldsymbol{J}_i\left( \bm{\Sigma}\right)^{l}\boldsymbol{B}_n}{N} :=R_{in}^{(l)} $ and $\left<y_n^{(l)} y_m^{(l)}\right> =\frac{\boldsymbol{B}_n\left( \bm{\Sigma}\right)^{l}\boldsymbol{B}_m}{N} := T_{nm}^{(l)}$. 
This property, where the pre-activations become Gaussian in the thermodynamic limit, is known as the Gaussian equivalence property \cite{PhysRevX.10.041044,goldt2022gaussian}.
The higher-order order parameters $Q_{ij}^{(l)}$, $R_{in}^{(l)}$, and $T_{nm}^{(l)}$ capture the relationships between the student and teacher weights at different levels. 
By expressing the generalization error as a function of these order parameters, we transition from a microscopic view -- focused on individual weight components -- to a macroscopic perspective that centers on the relationships between the student and teacher vectors without detailing their exact components. Understanding the dynamics of these order parameters allows us to effectively analyze the behavior of the generalization error.

\textbf{Dynamical Equations}. \hspace{3pt} During the learning process, we update the student vectors $\boldsymbol{J}_i$ using stochastic gradient descent after each presentation of an input example:
\begin{align}
\bm{J}_i^{\mu+1}-\bm{J}_i^{\mu}=-\eta \nabla_{\bm{J}_i} \epsilon \left(\bm{J}_i^\mu,\bm{\xi}^\mu\right),
\label{difference eq}
\end{align}
where $\eta$ is the learning rate. In the thermodynamic limit, as $p, N \to \infty$ while maintaining a finite ratio $\alpha = p/N$, \cite{NEURIPS2019_287e03db} derived a set of hierarchical differential equations describing the dynamics of the order parameters under stochastic gradient descent.
Applying these findings to our specific setup, we obtain the following differential equations:
\begin{align}
\frac{d\bm{R}^{(l)}}{d{\alpha}} &= \frac{\eta}{K} F_1\left(\bm{R}^{(1)},\bm{Q}^{(1)}, \bm{R}^{(l+1)}, \bm{Q}^{(l+1)}\right) \nonumber \\
\frac{d\bm{Q}^{(l)}}{d{\alpha}} &=  \frac{\eta}{K} F_2\left(\bm{R}^{(1)},\bm{Q}^{(1)}, \bm{R}^{(l+1)}, \bm{Q}^{(l+1)}\right) +  \frac{\eta^2}{K^2} \nu_{l+1} F_3\left(\bm{R}^{(1)},\bm{Q}^{(1)}\right) \ , 
\label{odes op}
\end{align}
where $\nu_{l} = \frac{1}{N} \sum_{k=1}^N \lambda_k^l$.
The functions $F_1$, $F_2$, and $F_3$ are defined in Appendix \ref{odes}. The transition from Eq.~(\ref{difference eq}) to Eq.~(\ref{odes op}) represents a shift from discrete-time updates indexed by $\mu$ to a continuous-time framework where $\alpha$ serves as a continuous time variable.

At this stage, the differential equations are not closed because the left-hand sides of Eqs.~(\ref{odes op}) involve derivatives of the $l$-th order parameters, while the right-hand sides depend on the next higher-order parameters $\boldsymbol{R}^{(l+1)}$ and $\boldsymbol{Q}^{(l+1)}$. To close the system of equations, we employ the Cayley–Hamilton theorem, which states that every square matrix satisfies its own characteristic equation. Specifically, for the covariance matrix $\boldsymbol{\Sigma}$, the characteristic polynomial is given by
$P(\boldsymbol{\Sigma}) := \prod_{k=1}^{L} \left( \boldsymbol{\Sigma} - \lambda_k \boldsymbol{I} \right) = \sum_{k=0}^{L} c_k \boldsymbol{\Sigma}^{k} = 0$, where $c_k$ are the coefficients of the polynomial, and $\lambda_k$ are the distinct eigenvalues of $\boldsymbol{\Sigma}$. Consequently, we can express the highest-order order parameters in terms of lower-order ones:
 $\bm{R}^{(L)} = -\sum_{l=0}^{L-1} c_l \bm{R}^{(l)}$, $\bm{Q}^{(L)} = -\sum_{l=0}^{L-1} c_l \bm{Q}^{(l)}$, and $\bm{T}^{(L)} = -\sum_{l=0}^{L-1} c_l \bm{T}^{(l)}$. 
 By substituting these expressions back into the differential equations, we close the system, resulting in $(KM + K^2) \times L$ coupled differential equations.
 Further details on the derivation of these differential equations are provided in Appendix \ref{odes}.

\section{Linear Activation function}
\subsection{Solution of order parameters} 
For the linear activation function, a significant simplification occurs: the generalization error becomes independent of the sizes of the student and teacher networks.
Specifically, we can replace the student and teacher vectors with their weighted sums, effectively acting as single resultant vectors. By defining
 $\tilde{\bm{B}}=\frac{1}{\sqrt{M}}\sum_n^M \bm{B}_n$, the student effectively learns this combined teacher vector.  
Consequently, we focus on the case where $K = M = 1$. In this scenario, the generalization error simplifies to
  $\epsilon_g= \frac{1}{2} \left( Q^{(1)}-2 R^{(1)}+T^{(1)}\right)$, which depends only on the first-order order parameters. Therefore, our main interest lies in solving the dynamics of these first-order parameters.
  Since we have only one student and one teacher vector, we represent the order parameters in vector form 
  $\bm{R}=\left(R^{(0)},R^{(1)},...,R^{(L-1)}\right)^\top $, $\bm{Q}=\left(Q^{(0)},Q^{(1)},...,Q^{(L-1)}\right)^\top$ and $\bm{T}=\left(T^{(0)},T^{(1)},...,T^{(L-1)}\right)^\top$. 
  Using this setup and notation, along with Eq.~(\ref{odes op}), we derive the following dynamical equation:

\begin{align}
\frac{d}{d\alpha}\begin{pmatrix}
\bm{R} \\ \bm{Q}
\end{pmatrix}= \eta \begin{pmatrix}
\bm{A}_1 & \bm{0}_{L \times L}   \\
-2 \bm{A}_1 -2\eta \bm{U} & 2\bm{A}_1 +\eta \bm{U}
\end{pmatrix} \begin{pmatrix}
\bm{R} \\ \bm{Q}
\end{pmatrix} + \eta \begin{pmatrix}
\bm{u} \\ \eta \bm{u}
\end{pmatrix},
\label{ode linear}
\end{align}
where $\boldsymbol{u} = \left( T^{(1)}, T^{(2)}, \dots, T^{(L)} \right)^\top$, $\boldsymbol{U} = \boldsymbol{u} \boldsymbol{e}_2^\top$, and $\boldsymbol{e}_2 = \left( 0, 1, 0, \dots, 0 \right)^\top$. The matrix $\boldsymbol{A}_1 \in \mathbb{R}^{L \times L}$ is defined in Appendix \ref{Solution of order parameters}.

From Eq.~(\ref{ode linear}), we observe that the differential equations governing the higher-order student-teacher order parameters $\boldsymbol{R}$ can be solved independently of the student-student parameters $\boldsymbol{Q}$. Therefore, to understand the dynamical behavior of $\boldsymbol{R}(\alpha)$, we need to determine the eigenvalues of $\boldsymbol{A}_1$, and for the asymptotic solution, we require its inverse. 
Additionally, the solution for the student-student order parameters $\boldsymbol{Q}(\alpha)$ depends on $\boldsymbol{R}(\alpha)$ and the spectrum of $\boldsymbol{A}_1 + \eta \boldsymbol{U}$.
In Appendix~\ref{Solution of order parameters}, we derive an expression for the generalization error averaged over the teacher and initial student entries $B_a$ and $J_a^0$:
\begin{align}
\langle \epsilon_g \rangle_{J_{a}^0, B_a} =  \frac{\left(1+\sigma_J^2\right) }{2L} \sum_{k=1}^L b_k \tilde{\lambda}_k \exp(-2\eta \tilde{\lambda}_k \alpha)\ ,
\label{eps perceptron time full}
\end{align}
where $\tilde{\lambda}_k$ are the eigenvalues of $\boldsymbol{A}_1 + \eta \boldsymbol{U}$, $b_k = \sum_{l=1}^L \left( \boldsymbol{W}^{-1} \right)_{kl} T^{(l)}$, and $\boldsymbol{W}$ contains the eigenvectors of $\boldsymbol{A}_1 + \eta \boldsymbol{U}$.

\begin{figure}
  \centering
  \includegraphics[width=0.49\linewidth]{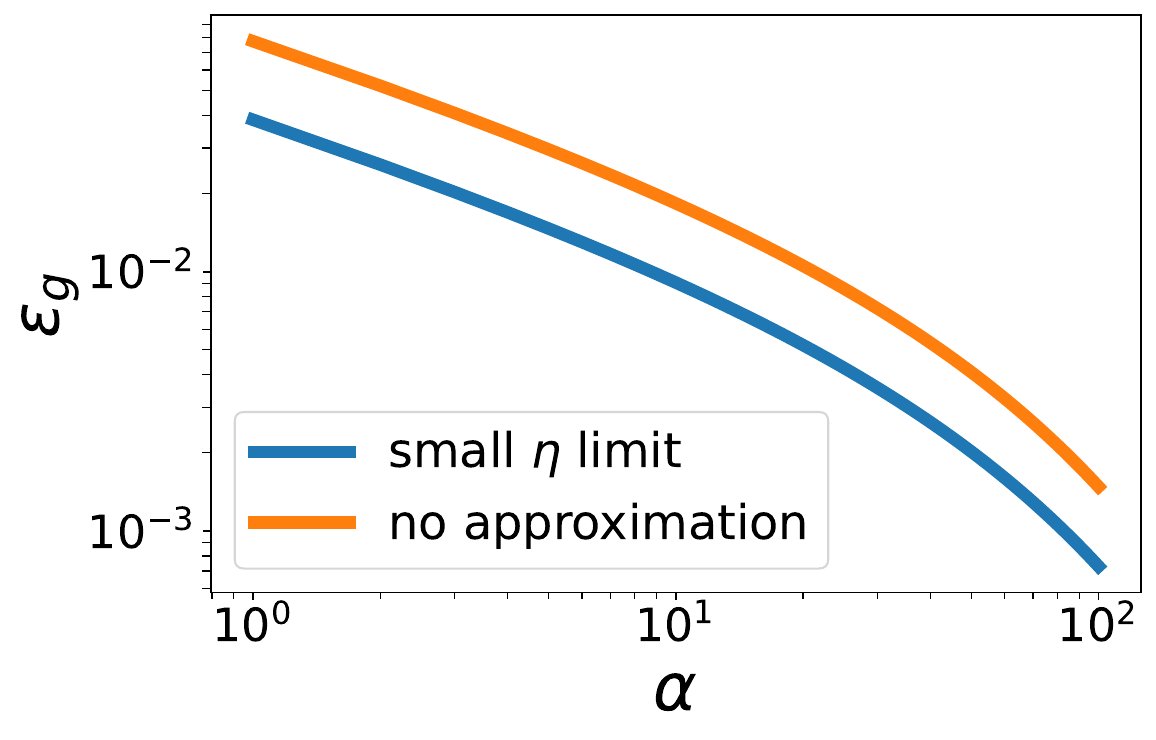}
  \includegraphics[width=0.49\linewidth]{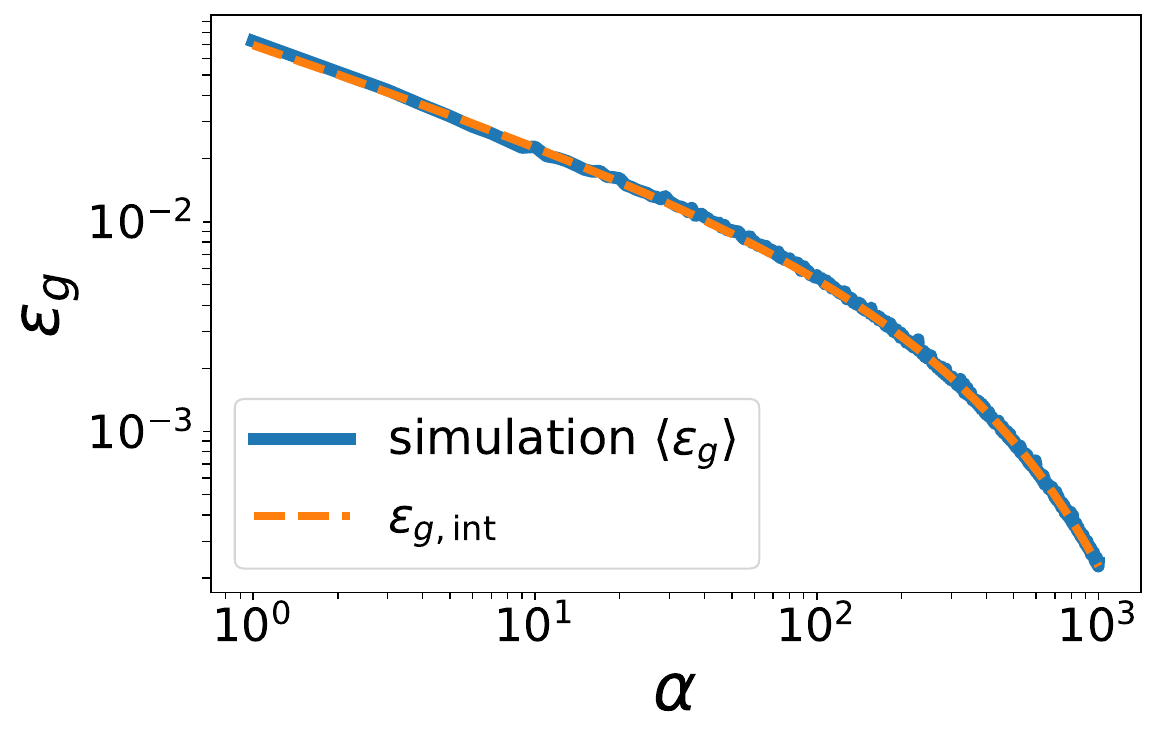}
  \caption{Generatlization error $\epsilon_g$ for linear activation function.
Left: $\epsilon_g$ evaluated using Eq.~(\ref{eps perceptron time}) (blue) and Eq.~(\ref{eps perceptron time full}) (orange) for $N = 128$, $K = M = 1$, $\sigma_J^2 = 1$, $\beta = 1$, and $\eta = 1$. Right: $\epsilon_g$ evaluated using Eq.~(\ref{linear time evolution integral}) (dashed orange) compared to simulations of a student-teacher setup averaged over 5 random initializations (solid blue), with $N = L = 256$, $\beta = 0.75$, $\eta = 0.1$, and $\sigma_J = 0.01$.}
  \label{fig: eg0 vs eg 1}
\end{figure}

This equation generally requires numerical evaluation.
Although $\boldsymbol{U}$ is a rank-$1$ matrix, standard perturbation methods are not applicable to find the eigenvalues of the shifted matrix $\boldsymbol{A}_1 + \eta \boldsymbol{U}$ because $\boldsymbol{U}$ may have a large eigenvalue, making it unsuitable as a small perturbation.
However, in the regime of small learning rates $\eta$, where we retain terms up to $\mathcal{O}(\eta)$ in Eq.~(\ref{ode linear}), we can determine the spectra of all involved matrices analytically and solve the differential equations.
The solutions for the first-order order parameters are then given by
\begin{align}
\langle R^{(1)}\rangle_{J_{a}^0, B_a}   &= 1-\frac{1}{L} \sum_k^L \lambda_k \exp\left(-\eta\lambda_k\alpha\right) , \nonumber \\
\langle Q^{(1)} \rangle_{J_{a}^0, B_a}  &= 1+\frac{1+\sigma_J^2}{L} \sum_k^L \lambda_k\exp\left(-2\eta\lambda_k\alpha\right)  - \frac{2}{L} \sum_k^L \lambda_k \exp\left(-\eta\lambda_k\alpha\right) \ ,
\label{sol lin first order RQ }
\end{align}
and the generalization error becomes
\begin{align}
\langle \epsilon_{g} \rangle_{J_{a}^0, B_a} \underset{\eta \to 0}{=}  \frac{1+\sigma_J^2}{2L} \sum_{k=1}^L {\lambda}_k \exp(-2\eta {\lambda}_k \alpha)\ .
\label{eps perceptron time}
\end{align}
Here, $\lambda_k$ are the distinct eigenvalues of the data covariance matrix as defined in Eq.~(\ref{data generating model}). Figure~\ref{fig: eg0 vs eg 1} compares the generalization error obtained from the exact solution in Eq.~(\ref{eps perceptron time full}) with the small learning rate approximation in Eq.~(\ref{eps perceptron time}). 
We observe that the generalization error without approximations consistently lies above the small learning rate solution. This discrepancy arises from the fluctuations in the stochastic gradient descent trajectory, which become more pronounced at larger learning rates.
\subsection{Scaling with time}
To evaluate the sum on the right-hand side of Eq.~(\ref{eps perceptron time}), we employ the Euler-Maclaurin approximation, which allows us to approximate the sum by an integral. In Appendix~\ref{Euler-Maclaurin with time}, we derive the following approximation for the generalization error:
  \begin{align}
 \langle \epsilon_{g}(\alpha) \rangle_{J_{a}^0, B_a} \underset{\eta \to 0}{\approx}  \lambda_+ \frac{1+\sigma_J^2}{2L}\frac{\left(2\eta\lambda_+ \alpha\right)^{-\frac{\beta}{1+\beta}}}{1+\beta} \Biggl[ \Gamma \left(\frac{\beta}{1+\beta}, \frac{2\eta\lambda_+ \alpha}{L^{\beta+1}}\right)-\Gamma \left(\frac{\beta}{1+\beta}, 2\eta\lambda_+ \alpha\right)\Biggr] \ , 
 \label{linear time evolution integral}
 \end{align}
where $\Gamma(s, x)$ is the incomplete gamma function. 
This expression reveals that the generalization error exhibits a power-law scaling within the time window 
 $\frac{1}{2\eta \lambda_+}<\alpha<\frac{1}{2\eta \lambda_+} \Gamma\left(\frac{2\beta+1}{1+\beta}\right)^{\frac{1+\beta}{\beta}} L^{1+\beta}$. 
In this regime, the generalization error scales as $\epsilon_g(\alpha) \propto \alpha^{-\frac{\beta}{1+\beta}}$, aligning with the results of \cite{bordelon2022learning} and \cite{bahri2021explaining} for the random feature model. The right panel of Figure~\ref{fig: eg0 vs eg 1} illustrates our analytical prediction from Eq.~(\ref{linear time evolution integral}), alongside the generalization error observed in a student neural network trained on Gaussian input data with a power-law spectrum.
Additional numerical analyses are provided in Appendix~\ref{Euler-Maclaurin with time}.
 
 \subsection{Feature scaling}
We first consider a diagonal covariance matrix $\bm{\Sigma}$, such that each entry of the student vector directly corresponds to an eigenvalue (see Appendix~\ref{Feature scaling appendix}). 
We later generalize to a non-diagonal covariance matrix and find that the same scaling law is obtained. 

\begin{figure}
  \centering
  \includegraphics[width=0.49\linewidth]{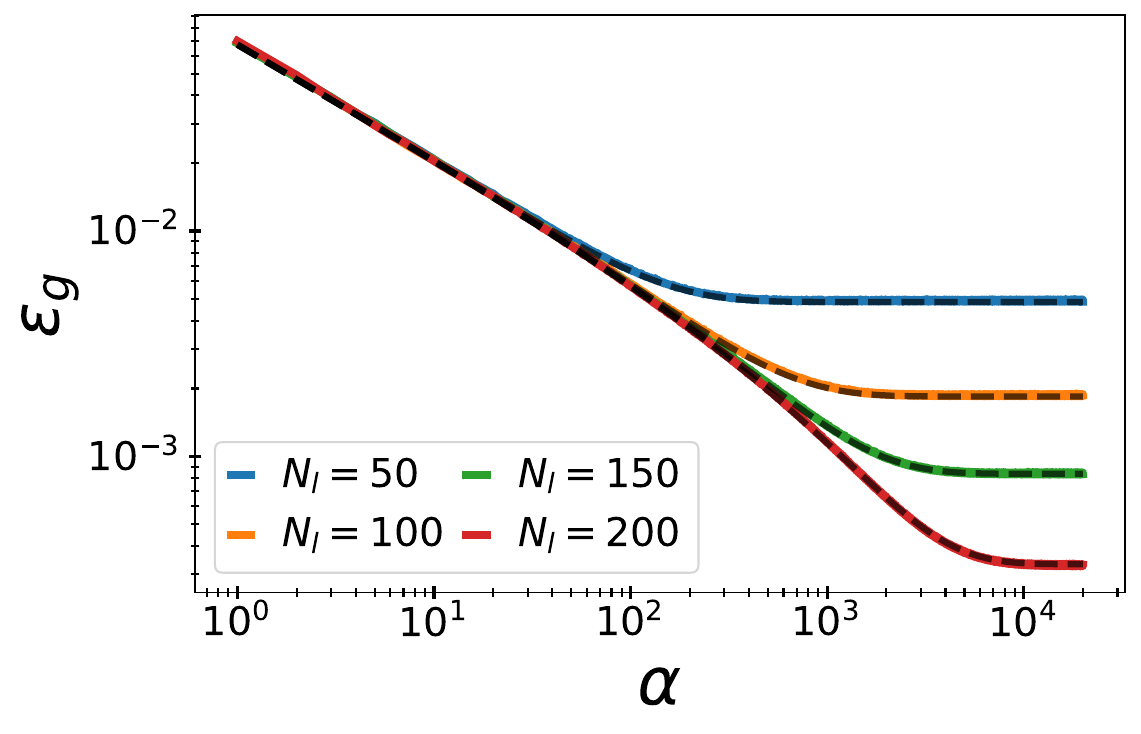}
    \includegraphics[width=0.49\linewidth]{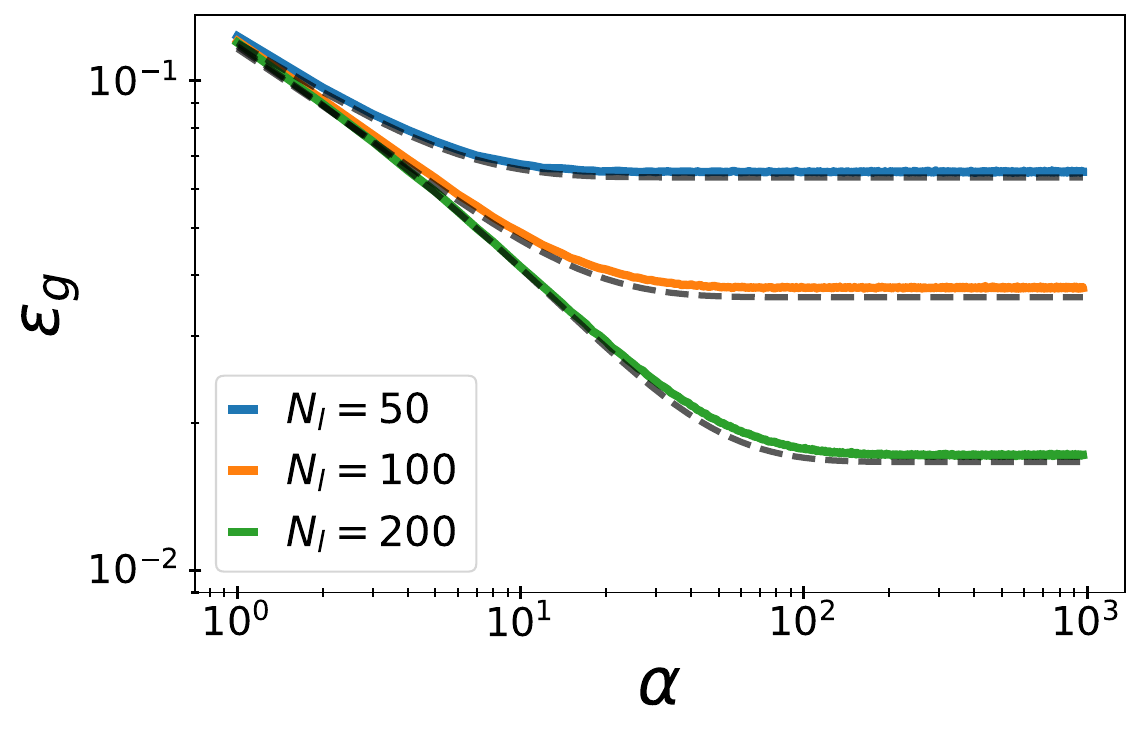}
  \caption{ Generalization error $\epsilon_g$ for different trainable input dimensions $N_l$ of the student network. Left: $\epsilon_g$ as a function of $\alpha$ for various $N_l$, with $L = N = 256$, $K = M = 1$, $\sigma_J = 0.01$, $\eta = 0.05$, and $\beta = 1$. The student network is trained on synthetic data and the teacher's outputs.
  Right: $\epsilon_g$ as a function of $\alpha$, with $L = N = 1024$, $K = M = 1$, $\sigma_J = 0.01$, and $\eta = 0.05$. The student network is trained on the CIFAR-5m dataset \cite{nakkiran2021the} using the teacher's outputs. We estimate the scaling exponent $\beta \approx 0.3$ for this dataset. For the theoretical predictions, the empirical data spectrum is used to evaluate Eq.~(\ref{eg min Nl}). Both plots compare the simulation results (solid curves) to the theoretical prediction from Eq.~(\ref{eg min Nl}) (black dashed lines). For both plots, the generalization error is averaged over 50 random initializations of the student and teacher vectors.
}
  \label{fig: eg lin Nl}
\end{figure}

Students typically learn directions associated with the largest eigenvalues of the data covariance matrix more rapidly \cite{ADVANI2020428}. To model this behavior, we assume the student can learn at most $N_l \leq L=N$ distinct eigenvalues of the data covariance matrix. Consequently, only the first $N_l$ entries of the student vector are trainable, while the remaining $N - N_l$ entries remain fixed at their initial random values.
 Our objective is to examine how the generalization error scales as the student explores more eigendirections of the data covariance matrix.
Figure~\ref{fig: eg lin Nl} displays the generalization error as a function of $\alpha$ for various values of $N_l$.
We observe that the generalization error approaches a limiting asymptotic value $\epsilon_{g,\mathrm{asymp}}$.
In Appendix~\ref{Feature scaling appendix}, we derive the following expression for the expected generalization error in this model: 
\begin{align}
\langle \epsilon_g \rangle_{J_{k}^0, B_k} \underset{\eta \to 0}{=} \frac{1+\sigma_J^2}{2L}\left[\sum_{k=1}^{N_l} \lambda_k \exp\left(-2\eta \lambda_{k} \alpha \right) +\sum_{k=N_l+1}^{L} \lambda_k \right] \ .
\label{eg final form params}
\end{align}
Using the Euler-Maclaurin formula, we approximate the sums by integrals and find:
\begin{align}
\langle \epsilon_{g}\left(N_l,\alpha \right) \rangle_{J_{a}^0, B_a}\underset{\eta \to 0}{\approx}  \frac{1+\sigma^2}{2}\frac{\lambda_+}{\beta L}\left(\frac{1}{N_l^\beta}-\frac{1}{L^\beta}\right) + \langle \epsilon_{g}(\alpha) \rangle_{J_{a}^0, B_a} \ .
\label{eg min Nl}
\end{align}
From this, we derive the asymptotic generalization error as  $\epsilon_{g,\mathrm{asymp}} \approx \frac{1+\sigma^2}{2}\frac{\lambda_+}{\beta L}\left(\frac{1}{N_l^\beta}-\frac{1}{L^\beta}\right)$. Thus, when $L^\beta > N_l^\beta$, we find a power-law scaling of the asymptotic generalization error with respect to the number of learned features: $\epsilon_{g,\mathrm{asymp}} \sim \frac{1}{N_l^\beta}$. A similar scaling result for feature scaling is presented in \cite{maloney2022solvable} for random feature models. However, our scaling exponent for the dataset size (parameterized by $\alpha$) differs from that for the number of features. In Appendix~\ref{Feature scaling appendix}, we analyze the student network trained with a non-diagonal data covariance matrix. In this setting, we find the same power-law exponent $\epsilon_{g,\mathrm{asymp}} \sim \frac{1}{N_l^\beta}$.
\section{Non-linear activation function }
\subsection{Plateau}
As discussed in the introduction and illustrated in Fig.~\ref{fig: emerging power-law}, both the length and height of the plateau are influenced by the number of distinct eigenvalues in the data covariance matrix. Specifically, as the number of distinct eigenvalues increases, the plateau becomes shorter and can eventually disappear. In this section, we present our findings that explain the underlying causes of this behavior for the case where $K = M$. \cite{Biehl_1996} derived a formula to estimate the plateau length $\alpha_P$ for a soft committee machine trained via stochastic gradient descent with randomly initialized student vectors. We adopt their heuristically derived formula for our setup, which takes the form
\begin{align}
\alpha_P-\alpha_0= \tau_{\mathrm{esc}} \left(D-\frac{1}{2}\ln\left(\sigma_J^2\right)+\frac{1}{2}\ln\left(N\right)\right)\ ,
\label{Biehl plateau}
\end{align}
where $D$ is a constant of order $\mathcal{O}(1)$ that depends on the variances at initialization and during the plateau phase, $\alpha_0$ is an arbitrary starting point on the plateau, and $\tau_{\mathrm{esc}}$ represents the escape time from the plateau. Our goal is to show how the escape time $\tau_{\mathrm{esc}}$ is modified when the dataset has a power-law spectrum.

However, there is not a single plateau or plateau length. As shown numerically by \cite{Biehl_1996}, multiple plateaus can exist, and their number depends on factors such as the network sizes $K$ and $M$, as well as hyperparameters like the learning rate $\eta$.
 To investigate how the plateau lengths depend on the number of distinct eigenvalues, we focus on the differential equations for the error function activation up to order $\mathcal{O}\left( \eta \right )$.  This corresponds to the small learning rate regime, although the associated plateau behavior can occur for intermediate learning rates as well.

\subsubsection{Plateau height}
In contrast to isotropic input correlations, the higher-order order parameters in our setup are no longer self-averaging, resulting in more complex learning dynamics. For the teacher-teacher order parameters, we find the expectation value 
 $\left\langle T_{nm}^{(l)} \right\rangle = \delta_{nm} \frac{1}{N} \operatorname{Tr}(\Sigma^l) $ and the variance $\mathrm{Var}\left( T_{nm}^{(l)} \right) =  \frac{\left(\delta_{nm}+1\right) }{N^2} \sum_k^N \lambda_k^2 $.
 Therefore, for both the diagonal and off-diagonal terms of the higher-order teacher-teacher order parameters, the variance does not decrease with increasing input dimension $N$ when $l > 0$. Consequently, the plateau height and length can fluctuate between different initializations, even in the small learning rate regime. 
 \begin{figure}[t]
  \centering
    \includegraphics[width=0.329\linewidth]{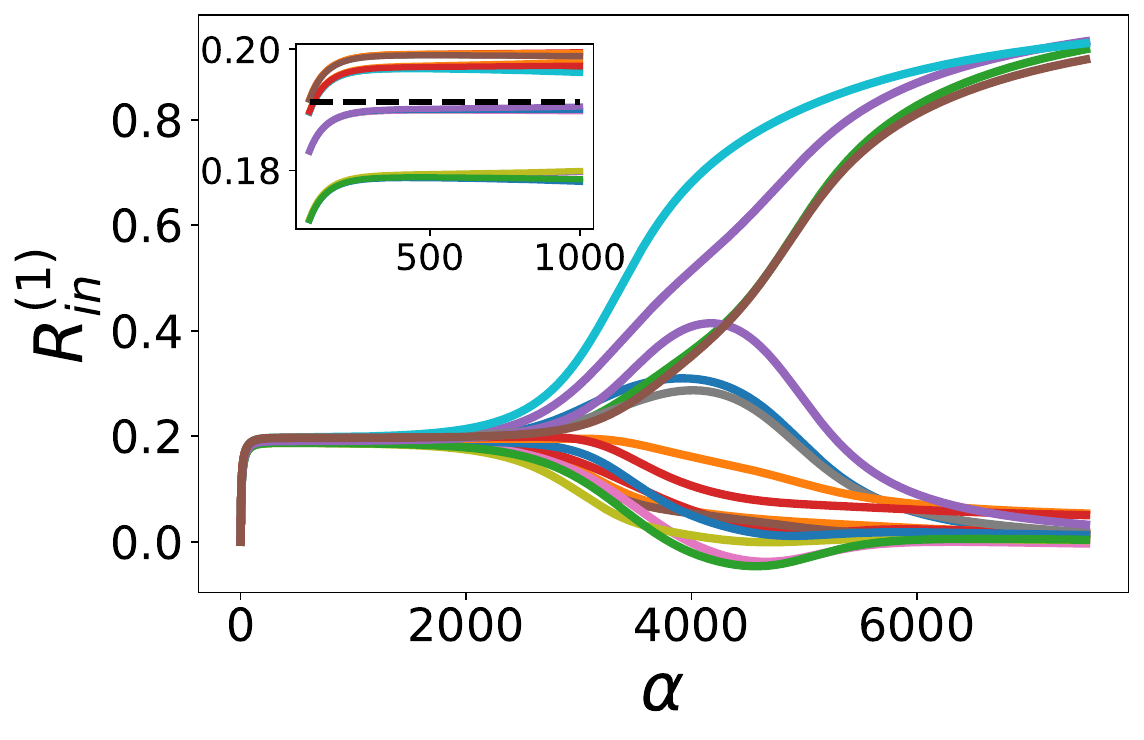}
  \includegraphics[width=0.329\linewidth]{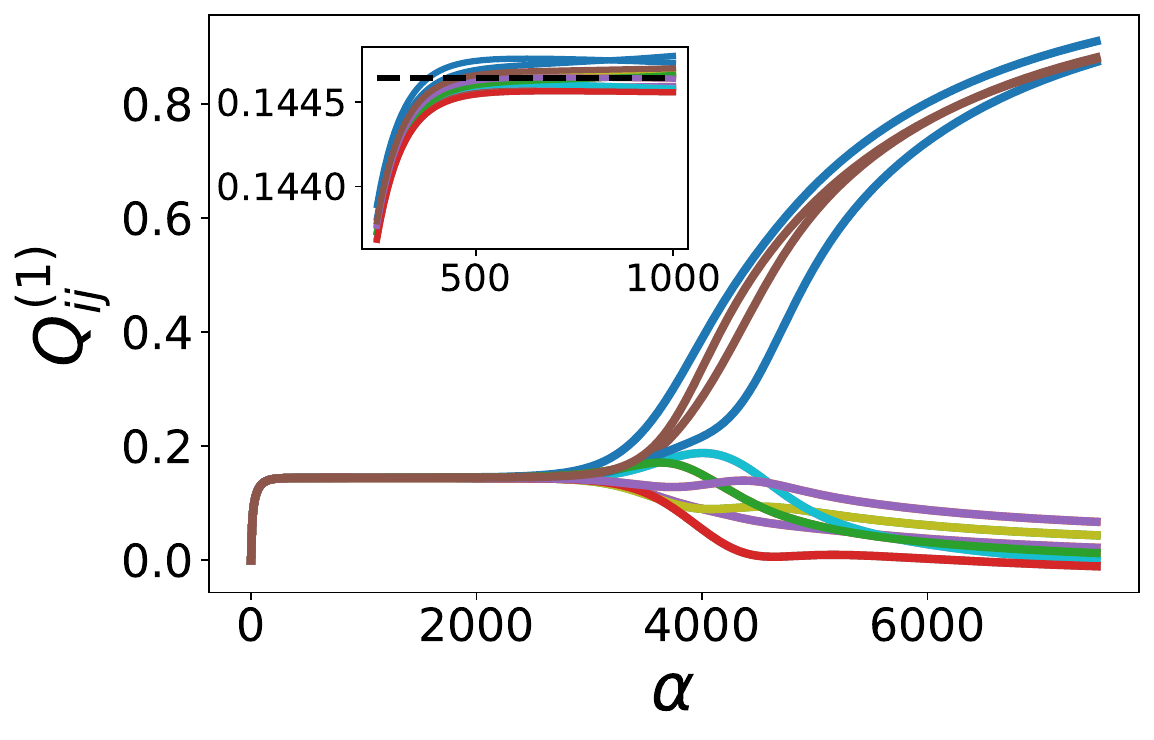} 
  \includegraphics[width=0.329\linewidth]{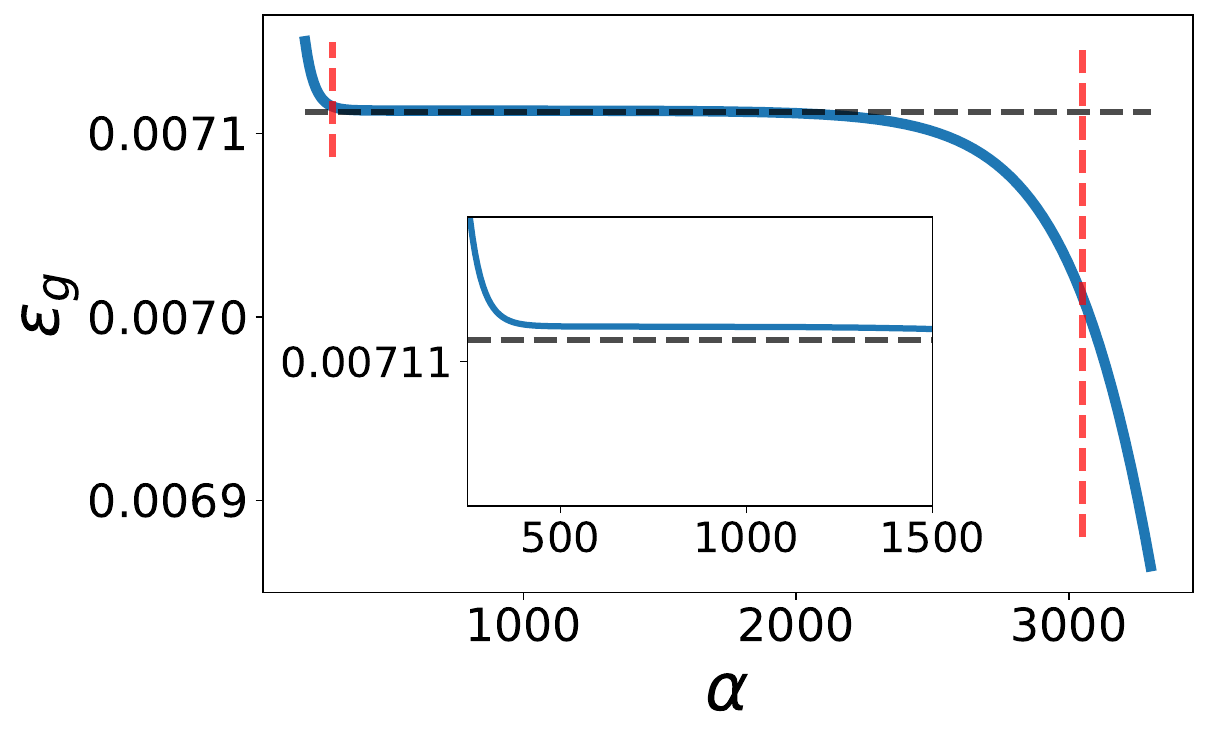}
  \caption{ Symmetric plateau for a non-linear activation function. Left and center: Plateau behavior of the order parameters for $L = 10$, $N = 7000$, $\sigma_J = 0.01$, $\eta = 0.1$, and $M = K = 4$, using one random initialization of the student and teacher vectors. We solve the differential equations in the small learning rate regime, retaining terms up to $\mathcal{O}(\eta)$. The insets display the higher-order order parameters at the plateau. For the student-teacher order parameters, we observe $M$ distinct plateau heights, while the student-student order parameters exhibit a single plateau height with minor statistical deviations in the matrix entries $Q_{ij}^{(l)}$. The dashed horizontal lines in the insets correspond to the plateau heights predicted by Eq.~(\ref{plateau fix 1}).
  Right: Corresponding generalization error $\epsilon_g$ for the same setup. The vertical dashed lines indicate the estimated plateau length based on Eqs.~(\ref{Biehl plateau}) and (\ref{escape time}).}
  \label{fig: plateau height first order}
\end{figure}
Figure \ref{fig: plateau height first order} displays the first-order order parameters as functions of $\alpha$. For the student-teacher order parameters, we observe $M$ distinct plateau heights, while the student-student order parameters exhibit a single plateau height. In Appendix \ref{plateau height appendix}, we show that these unique plateau heights are determined by the sum of off-diagonal elements  $d^{(l)}_n = \sum_{m,m \neq n}^{M-1} T^{(l)}_{nm}$ for each row $n$ of the teacher-teacher matrix. To simplify the analysis, we assume that all diagonal elements are equal to $T^{(l)}$, and all off-diagonal elements are given by $T_{nm}^{(l)} = \frac{1}{M-1}D^{(l)}$, where $D^{(l)}$ represents the average sum of off-diagonal entries. This approximation captures the general behavior of the plateaus. By considering the stationary solutions to Eqs.~(\ref{odes op}), we find the fixed points for $l=1$
\begin{align}
R^{*^{(1)}} =\frac{1}{\sqrt{\frac{M}{T^{(1)}+D^{(1)}}\left(\frac{M T^{(1)}}{T^{(1)}+D^{(1)}} \left(1+\frac{1}{T^{(1)}}\right)-1\right)}}  , && Q^{*^{(1)}} =  \frac{1}{\frac{ MT^{(1)}}{T^{(1)}+D^{(1)}} \left(1+\frac{1}{T^{(1)}}\right)-1} .
\label{plateau fix 1}
\end{align}
Expressions for the fixed points of higher-order order parameters are provided in Appendix \ref{plateau height appendix}.

\subsubsection{Escape from the plateau}
\label{plateau_escape}

To escape from the plateau, the symmetry in each order $l$ of the order parameters must be broken. To model this symmetry breaking, we introduce parameters $S^{(l)}$ and $C^{(l)}$, which indicate the onset of specialization for the student-teacher and student-student order parameters, respectively.
 Specifically, we use the parametrization $R_{im}^{(l)} = R^{(l)} \delta_{im} + S^{(l)} (1 - \delta_{im})$ and $Q_{ij}^{(l)} = Q^{(l)} \delta_{ij} + C^{(l)} (1 - \delta_{ij})$. To study the onset of specialization, we introduce small perturbation parameters $r^{(l)}$, $s^{(l)}$, $q^{(l)}$, and $c^{(l)}$ to represent deviations from the plateau values: $R^{(l)} = R^{*^{(l)}} + r^{(l)}$, $S^{(l)} = S^{*^{(l)}} + s^{(l)}$, $Q^{(l)} = Q^{*^{(l)}} + q^{(l)}$, and $C^{(l)} = C^{*^{(l)}} + c^{(l)}$, where $S^{*^{(l)}} = R^{*^{(l)}}$ and $C^{*^{(l)}} = Q^{*^{(l)}}$.
  Therefore, instead of analyzing the dynamics of the order parameters directly, we focus on the dynamics of these perturbative parameters and linearize the differential equations given in Eq.~(\ref{odes op}).

In Appendix~\ref{plateau escape appendix}, we demonstrate that, due to the structure of the leading eigenvectors of the dynamical system, we can set  $c^{(l)}=q^{(l)}=\frac{2T^{(l)}}{{T^{(1)}+D^{(1)}}} R^{*^{(l)}} \left(r^{(l)}+(M-1)s^{(l)}\right) $ and $s^{(l)}=\frac{-1}{(M-1)}r^{(l)} $. 
This allows us to obtain a reduced dynamical differential equation of the form
\begin{figure}
  \centering
  \includegraphics[width=0.49\linewidth]{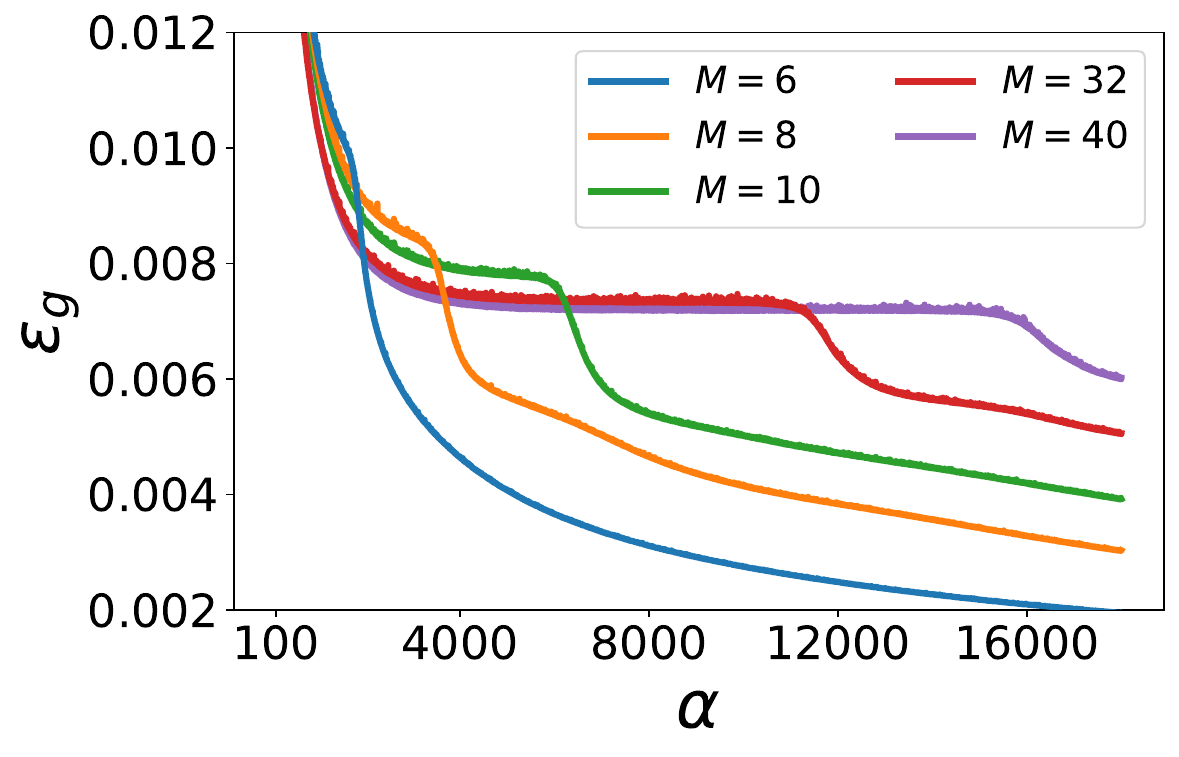} 
  \includegraphics[width=0.49\linewidth]{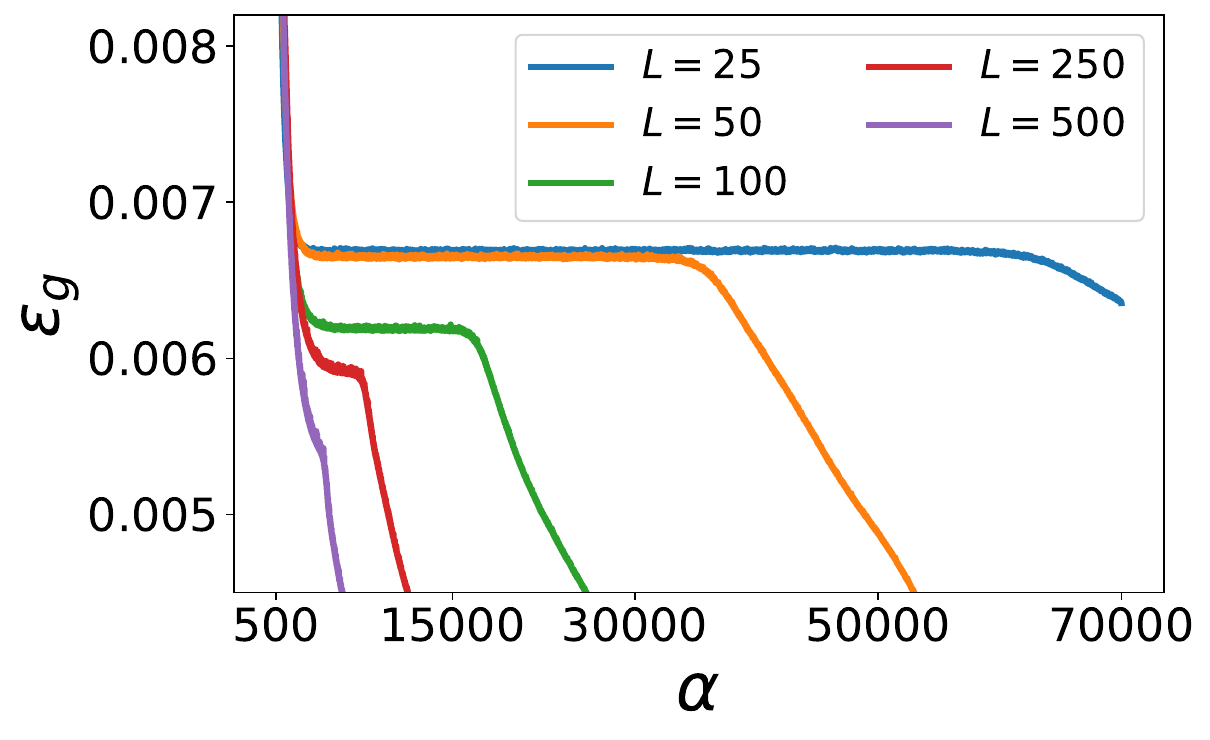} 
  \caption{ Plateau behavior of the generalization error  $\epsilon_g$  obtained by simulations for one random initialization of student and teacher vectors with $K=M$, $\eta=0.01$, $\sigma_J=10^{-6}$ and $\beta=0.25$. Left: $\epsilon_g$ for different student-teacher sizes for $L=N=512$. Right: $\epsilon_g$ for different numbers of distinct eigenvalues $L$ and for $K=M=6$ and $N=500$. }
  \label{fig: plateau length scaling}
\end{figure}

\begin{align}
\frac{d\bm{r}}{d\alpha} = \eta \bm{A}_r\bm{r},
\end{align}
where $\bm{r} = [r^{(1)}, r^{(2)}, \dots, r^{(L)}]^\top$, and $\bm{A}_r \in \mathbb{R}^{L \times L}$ is defined in Appendix~\ref{plateau escape appendix}. After solving these differential equations, we find that the escape from the plateau follows  $\epsilon_g^*-\epsilon_g \propto e^{\frac{\alpha}{\tau_{\mathrm{esc}}}} $, where the escape time $\tau_{\mathrm{esc}}$ is given by
\begin{align}
\tau_{\mathrm{esc}} &=  \frac{\pi}{2\eta}\frac {\sqrt{(M-1)T^{(1)}-D^{(1)}+M} \left(D^{(1)}+ (M+1)T^{(1)}+M\right)^{\frac{3}{2}}} { \left(T^{(2)}-\frac{D^{(2)}}{M-1}\right) \left(D^{(1)}+T^{(1)}\right)} \ .
\label{escape time}
\end{align}
For large $L$, one can show that $T^{(2)} \propto L$.
Therefore, for large $M$ and $L$, the escape time scales as $\tau_{\mathrm{esc}} \sim \frac{M^2}{\eta L}$. This behavior is illustrated in Figure~\ref{fig: plateau length scaling}, where we train a student network with synthetic input data.
Additional numerical results for the plateau length are provided in Appendix~\ref{plateau escape numerics appendix}.

\subsection{asymptotic solution}

In this subsection, we investigate how the generalization error converges to its asymptotic value. To this end, we consider the typical teacher configuration where $\langle T_{nm}^{(l)} \rangle = \delta_{nm} T^{(l)}$, as this configuration effectively captures the scaling behavior of the generalization error. For the asymptotic fixed points of the order parameters, we find $R_{im}^{*^{(l)}} = T^{(l)} \delta_{im}$ and $Q_{ij}^{*^{(l)}} = T^{(l)} \delta_{ij}$.
To model the convergence towards the asymptotic solution, we again distinguish between diagonal and off-diagonal entries, parametrizing the order parameters as $R_{im}^{(l)} = R^{(l)} \delta_{im} + S^{(l)} (1 - \delta_{im})$ and $Q_{ij}^{(l)} = Q^{(l)} \delta_{ij} + C^{(l)} (1 - \delta_{ij})$, similar to the plateau case. We then linearize the dynamical equations for small perturbations around the fixed points, setting $R^{(l)} = T^{(l)} + r^{(l)}$, $S^{(l)} = T^{(l)} + s^{(l)}$, $Q^{(l)} = T^{(l)} + q^{(l)}$, and $C^{(l)} = T^{(l)} + c^{(l)}$, and retain terms up to $\mathcal{O}\left( \eta \right)$.
This yields the following linearized equation
\begin{figure}
  \centering
  \includegraphics[width=0.49\linewidth]{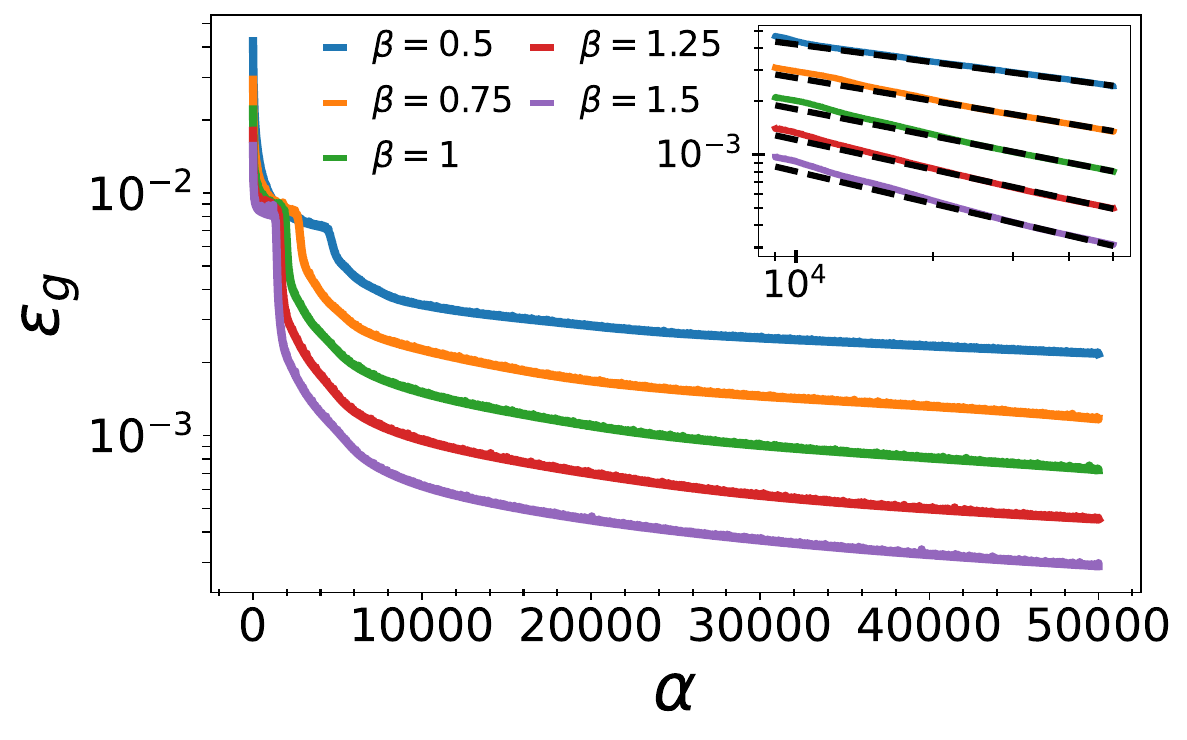}
  \includegraphics[width=0.49\linewidth]{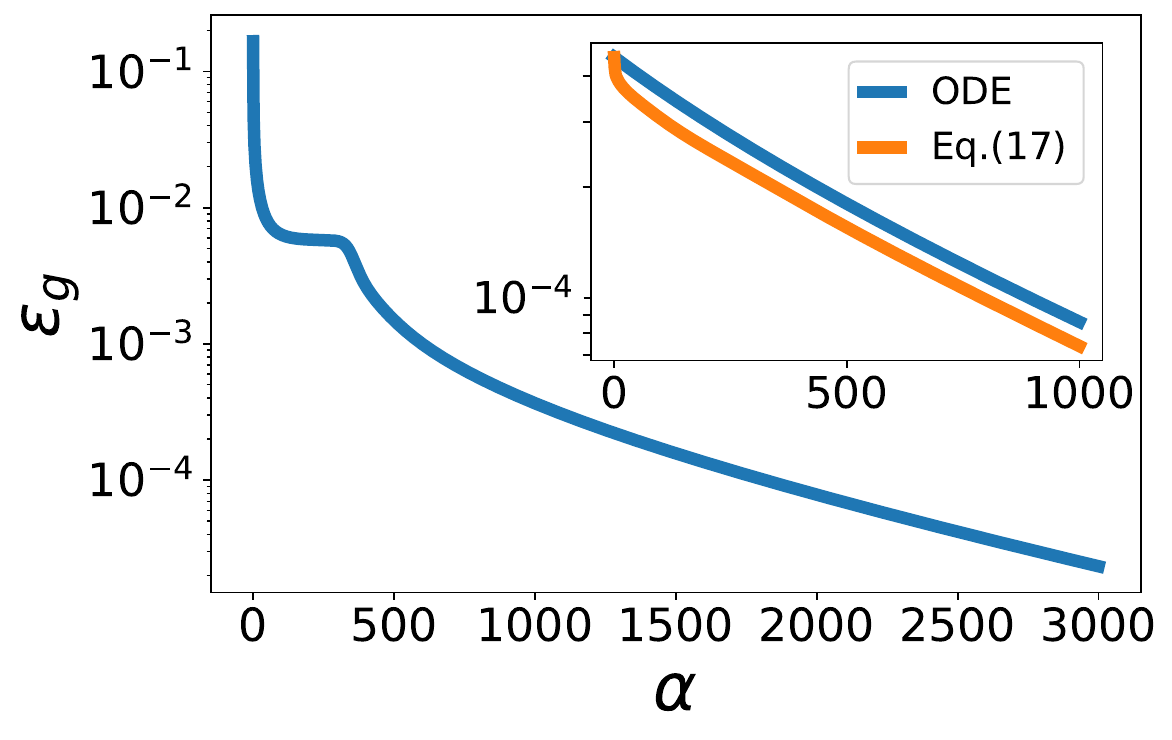}
  \caption{ Scaling behavior of the generalization error $\epsilon_g$ in the asymptotic regime for a non-linear activation function. Left: $\epsilon_g$ as a function of $\alpha$ for $K=M=40$, $\eta=0.01$,  $\sigma_J=10^{-6}$ and $L=N=512$ for simulations averaged over $10$ different initializations. Inset: comparison to a power law with exponent $-\beta/(1+\beta)$. Right: $\epsilon_g$ obtained from equation Eq. (\ref{first order eg}) (orange) in comparison to the solution of differential equations of $\mathcal{O}(\eta)$ (blue) for $K=M=2$, $\eta=0.25$, $\beta=1$ and $L=9$. }
  \label{fig: realizable}
\end{figure}
\begin{align}
\frac{d\bm{x}}{d\alpha} =a \bm{A}_{\mathrm{asym}} \hspace{3pt} \bm{x}
\label{ode asymp}
\end{align}
where $\bm{x}_i = \left( r^{(i - 1)}, q^{(i - 1)}, s^{(i - 1)}, c^{(i - 1)} \right)^\top$, $a = \frac{2 \sqrt{3}}{3 \pi M}$, and $\bm{A}_{\mathrm{asym}} \in \mathbb{R}^{4L \times 4L}$ is defined in Appendix~\ref{asymptotic appendix}.
After solving Eq.~(\ref{ode asymp}), we find for the generalization error
\begin{align}
\epsilon_g &= \frac{1}{6\pi}\sum_{k=1}^L g_k^{(1)}  e^{-a\left(2+\sqrt{3}\right)\lambda_k \alpha} \left( 2\sqrt{3}{v}^{(1)}_{k,2L+2} -4\sqrt{3}{v}^{(1)}_{k,2} +3\left(M-1\right){v}^{(1)}_{k,3L+2} -6\left(M-1\right){v}^{(1)}_{k,L+2}  \right) \nonumber \\
&+g_k^{(2)} e^{-a\left(2-\sqrt{3}\right)\lambda_k \alpha}  \left( 2\sqrt{3}{v}^{(2)}_{k,2L+2} -4\sqrt{3}{v}^{(2)}_{k,2} +3\left(M-1\right){v}^{(2)}_{k,3L+2} -6\left(M-1\right){v}^{(2)}_{k,L+2}  \right) \ ,
\label{first order eg}
\end{align}
where $\lambda_k$ are the eigenvalues of the data covariance matrix, $v_k^{(1)}$ and $v_k^{(2)}$ are two groups of eigenvectors corresponding to the eigenvalues $\left(2 + \sqrt{3} \right) \lambda_k$ and $\left(2 - \sqrt{3}\right) \lambda_k$, and the coefficients $g_k^{(1)}$ and $g_k^{(2)}$ depend on the initial conditions, as detailed in Appendix~\ref{asymptotic appendix}. The asymptotic convergence is governed by the smaller group of eigenvalues $\left(2 - \sqrt{3}\right) \lambda_k$. The weighted sum of exponentials results in a slowdown of the convergence of the generalization error, similar to the linear activation function case. Figure~\ref{fig: realizable} illustrates the generalization error during the late phase of training for different $\beta$. In all configurations, we observe the previously derived scaling $\epsilon_g \propto \alpha^{-\frac{\beta}{1+\beta}}$, consistent with the linear activation function.

\section{Conclusion}

We have provided a theoretical analysis of neural scaling laws within a student-teacher framework using statistical mechanics. By deriving analytical expressions for the generalization error, we demonstrated how power-law spectra in the data covariance matrix influence learning dynamics across different regimes. For linear activation functions, we have established the conditions under which power-law scaling for the generalization error with $\alpha$ emerges and computed the power-law exponent for the scaling of the generalization error with the number of student parameters. For non-linear activations, we presented an analytical formula for the plateau length, revealing its dependence on the number of distinct eigenvalues and the covariance matrix's power-law exponent. In addition,  we found that the convergence to perfect learning transitions from exponential decay to power-law scaling when the data covariance matrix exhibits a power-law spectrum. This highlights the significant impact of data correlations on learning dynamics and generalization performance.

Note added: After completion of this work, we became aware of the preprint \cite{bordelon2024featurelearningimproveneural}, which also studies neural scaling laws in the feature learning regime.

\subsubsection*{Acknowledgments}
This work was supported by the IMPRS MiS Leipzig.
\bibliography{scaling_laws_bib.bib}

\bibliographystyle{iclr2025_conference}
\newpage
\appendix

\section{Differential Equations}
\label{odes}
From the stochastic gradient descent given in Eq. (\ref{difference eq}), one can derive the following differential equations in the thermodynamic limit $N \to \infty$ (see \cite{NEURIPS2019_287e03db})

\begin{align}
\frac{dQ_{ij}^{(l)}}{d{\alpha}} &=  \frac{\eta}{K} \left[\sum_{m=1}^{M}  I_3(x_i, x_j^{(l)}, y_m) - \frac{M}{K}\sum_{k=1}^{K}  I_3(x_i, x_j^{(l)}, x_k)
    +  \sum_{m=1}^{M}  I_3(x_j, x_i^{(l)}, y_m) - \frac{M}{K} \sum_{k=1}^{K}  I_3(x_j, x_i^{(l)}, x_k)\right] \nonumber \\
   & +  \frac{\eta^2}{K^2} \nu_{l+1} \Biggr[ \frac{M}{K^2} \sum_{k,l=1}^{K,K}  I_4(x_i, x_j, x_k, x_l)+  \sum_{n,m=1}^{M,M}  I_4(x_i, x_j, y_n, y_m) 
    - \frac{2}{K} \sum_{k,n=1}^{K,M} I_4(x_i, x_j, x_k, y_n)   \nonumber
\Biggr] \nonumber \\
\frac{dR_{in}^{(l)}}{d{\alpha}} &= \eta \left[
    \sum_{m=1}^{M} I_3(x_i, y_n^{(l+1)}, y_m) - \sum_{j=1}^{K}  I_3(x_i, y_n^{(l+1)}, x_j)
\right],
\label{odes op 2}
\end{align}
with $\nu_{l}= \frac{1}{N} \sum_k^N \lambda_k^l$, $I_3(z_1, z_2, z_3) = \langle g'(z_1) z_2 g(z_3) \rangle \quad  \text{and} \quad I_4(z_1, z_2, z_3, z_4) = \langle g'(z_1) g'(z_2) g(z_3) g(z_4) \rangle $. In this setting, the generalization error becomes
 
\begin{align}
\epsilon_g = \frac{1}{2} \left[ \sum_{m,n=1}^{M} I_2(y_n^{(1)}, y_m^{(1)}) + \sum_{i,j=1}^{K} I_2(x_j^{(1)}, x_i^{(1)}) - 2 \sum_{i=1}^{K} \sum_{n=1}^{M} I_2(x_i^{(1)}, y_n^{(1)}) \right].
\end{align}

where \(I_2(z_1, z_2) := \langle g(z_1) g(z_2) \rangle\). Thereby, the $I_2. I_3$ and $I_4$ are integrals over the generalized pre-activations $x_i^{(l)}$ and $y_n^{(l)}$. Thereby, $z_i$ are normally distributed variables and stand for either $x_i^{(l)}$ or $y_n^{(l)}$. Therefore, the integrals $I_2, I_3$, and $I_4$, are multivariate Gaussian expectation values that are determined by the expectation values and covariance matrix of the generalized pre-activations. $I_2$ is a two-dimensional Gaussian integral. For example, for $I_2(x_j^{(1)}, x_j^{(1)}) $, on obtains the following covariance matrix

\begin{align}
    \bm{C}\left(i,j\right)=\begin{pmatrix} 
        Q_{ii}^{(1)} & Q_{ij}^{(1)}  \\ 
        Q_{ij}^{(1)} & Q_{jj}^{(1)} \\
    \end{pmatrix}.
    \end{align}
    
The resulting elements of the covariance matrices depend on the higher-order order parameters. $I_3$ is a three-dimensional Gaussian integral, and an example of the covariance matrix is

    \begin{align}
    \bm{C}^{(l)}\left(i,j,n\right)=\begin{pmatrix} 
        Q_{ii}^{(1)} & Q_{ij}^{(1)} & R_{in}^{(1)} \\ 
        Q_{ij}^{(l+1)} & Q_{jj}^{(2l+1)} & R_{jn}^{(l+1)} \\ 
        R_{in}^{(1)} & R_{jn}^{(l+1)} & T_{nn}^{(1)} 
    \end{pmatrix},
\end{align}

for $I_3(x_j^{(1)}, x_j^{(l)},y_n^{(1)}) $. Note that $I_3$ depends on higher-order $l$ as compared to $I_2$ and $I_4$, which only depend on the first-order. Furthermore, $I_4$ is a four-dimensional Gaussian integral and depends, for example, on the following covariance matrix for $I_4(x_j^{(1)}, x_j^{(1)},y_n^{(1)},y_m^{(1)}) $

\begin{align}
    \bm{C}\left(i,j,n,m\right)=\begin{pmatrix} 
        Q_{ii}^{(1)} & Q_{ij}^{(1)} & R_{in}^{(1)} & R_{im}^{(1)} \\ 
        Q_{ij}^{(1)} & Q_{jj}^{(1)} & R_{jn}^{(1)} & R_{jm}^{(1)} \\ 
        R_{in}^{(1)} & R_{jn}^{(1)} & T_{nn}^{(1)} & T_{nm}^{(1)} \\ 
        R_{im}^{(1)} & R_{jm}^{(1)} & T_{nm}^{(1)} & T_{mm}^{(1)} 
    \end{pmatrix}.
\end{align}
The specific differential equations for the linear and error function activation are provided in their corresponding subsections.

\section{Linear activation}
\subsection{Solution of order parameters} 
\label{Solution of order parameters}
The linear activation function leads to the following teacher output $\zeta(\boldsymbol{B},\boldsymbol{\xi})= \frac{1}{\sqrt{M}}\sum_{n=1}^M \frac{\bm{\xi}\bm{B}_n}{\sqrt{N}}$. This makes it possible to rewrite the output as $\zeta(\boldsymbol{B},\boldsymbol{\xi})= \frac{1}{\sqrt{M}}\sum_{n=1}^M \frac{\bm{\xi}\bm{B}_n}{\sqrt{N}}= \frac{\bm{\xi}}{\sqrt{N}} \frac{1}{\sqrt{M}}\sum_{n=1}^M  \bm{B}_n =\frac{\bm{\xi}\tilde{\bm{B}}}{\sqrt{N}} $, where we have defined a new teacher vector $ \tilde{\bm{B}}=\frac{1}{\sqrt{M}}\sum_{n=1}^M  \bm{B}_n$. Since $\tilde{\bm{B}}$ has the same statistical properties as one random teacher vector $\bm{B}_n$ it makes no difference whether we consider the case $M>1$ and define $\tilde{\bm{B}}$ or $M=1$. The same argument also applies to the linear student network. Therefore, in the following, we analyze the case $K=M=1$. \\
For $K=M=1$, the generalization error becomes
\begin{equation}
\epsilon_g= \frac{1}{2} \left( Q^{(1)}-2 R^{(1)}+T^{(1)}\right).
\label{eg linear}
\end{equation} 
The differential equations for the order parameters are
\begin{align}
\frac{dR^{(l)}}{d\alpha}&= \eta \left[T^{(l+1)}-R^{(l+1)}\right], \nonumber \\  \frac{dQ^{(l)}}{d\alpha}&=  2 \eta \left[R^{(l+1)}-Q^{(l+1)}\right] + \eta^2 T^{(l+1)} \left[T^{(1)}+ Q^{(1)}-2R^{(1)}\right]
\label{linear diff 1}
\end{align}
with $0\leq l \leq L-2$ and for the last component
\begin{align}
\frac{dR^{(L-1)}}{d\alpha}&= \eta \left[\sum_k^{L-1} c_k \left( T^{(k)}-R^{(k)} \right)\right], \nonumber \\
 \frac{dQ^{(L-1)}}{d\alpha}&=  2 \eta \left[\sum_k^{L-1} c_k \left( R^{(k)}-Q^{(k)} \right)\right] + \eta^2 T^{(L)}\left[T^{(1)}+ Q^{(1)}-2R^{(1)}\right].
 \label{linear diff 2}
\end{align}
Thereby, we have exploited the characteristic polynomial for the order parameters for the $L$-th order, e.g. $R^{(L)}= -\sum_{k=0}^{L-1} c_k R^{(k)}$, with the coefficients of the characteristic polynomial $c_k$. The set of coupled linear differential equations given by Eqs. (\ref{linear diff 1}) and (\ref{linear diff 2}) can be written in the following form

\begin{align}
\frac{d}{d\alpha}\begin{pmatrix}
\bm{R} \\ \bm{Q}
\end{pmatrix}= \eta \begin{pmatrix}
\bm{A}_1 & \bm{0}_{L \times L}   \\
-2 \bm{A}_1 -2\eta \bm{U} & 2\bm{A}_1 +\eta \bm{U}
\end{pmatrix} \begin{pmatrix}
\bm{R} \\ \bm{Q}
\end{pmatrix} + \eta \begin{pmatrix}
\bm{u} \\ \eta \bm{u}
\end{pmatrix},
\label{block linear diff}
\end{align}
where  $\bm{u}=\left(T^{(1)},T^{(2)},...,T^{(L)}\right)^\top$, $\bm{U}=\bm{u}\bm{e}_2^\top$ with $\bm{e}_2=\left(0,1,0,...,0\right)^\top$ and 

\begin{align}
\bm{A}_1= \begin{pmatrix}
0 & -1      & 0       & \cdots & 0       & 0 \\
0        & 0 & -1     & \ddots & \vdots  & \vdots \\
0        & 0       & 0& \ddots & 0       & 0 \\
\vdots   & \vdots  & \ddots  & \ddots & -1      & 0 \\
0        & 0       & \cdots  & 0      & 0 & -1 \\
c_0      & c_1     & c_2     & \cdots & c_{L-2} & c_{L-1}
\end{pmatrix} , && \bm{U}=\bm{u}\bm{e}_2^\top.
\end{align}
Therefore, the differential equations for the order parameters are

\begin{align}
\frac{d\bm{R}}{d\alpha} &= \eta \bm{A}_1 \bm{R} + \eta \bm{u} \\
\frac{d\bm{Q}}{d\alpha} &= \eta \bm{A}_3 \bm{R} + \eta \bm{A}_4 \bm{Q} + \eta^2 \bm{u} 
\label{linear diff eq}
\end{align}

with $\bm{A}_3=-2\bm{A}_1-2 \eta \bm{U}$, $\bm{A}_4=2\bm{A}_1+\eta\bm{U}$.\\
Thus, we can solve differential equations for the student-teacher order parameters independent from the student-student and find
\begin{align}
 \bm{R}\left(\alpha\right)  &= e^{\eta\bm{A}_1 \alpha}\bm{R}_0+ e^{\eta \bm{A}_1 \alpha} \bm{A}_1^{-1} \bm{u} - \bm{A}_1^{-1} \bm{u} ,
 \label{formal sol R}
\end{align}
where $\bm{R}_0$ are the student-teacher order parameters at initialization and we set $\bm{R}_0=0$ which is achieved in the large $N$ limit and on average. Before inserting this result into the differential equation for the student-student order parameters, we evaluate Eq. (\ref{formal sol R}). For this, we need to find the eigenvalues of the matrix $\bm{A}_1$ and evaluate  $\bm{A}_1^{-1} \bm{u} $.\\
First, we start with the eigenvalues. In order to find the determinant of $\bm{A}_1-\lambda\bm{I}_{L}$, we apply the Laplace expansion with respect to the last row of $\bm{A}_1-\lambda\bm{I}_{2L}$ and evaluate the determinants of $L$ different $L-1 \times L-1$ smaller matrices. The resulting sub-matrices are triangular, and their determinant is, therefore, simply given by the product of the diagonal entries. After applying the Laplace expansion, we find
\begin{align}
\det\left(\bm{A}_1-\lambda\bm{I}_{L}\right) &= \sum_{i=0}^{L-2} c_i (-1)^{L+1+i} (-1)^{L-1-i}(-\lambda)^i + (c_{L-1}-\lambda) (-1)^{2L} (-\lambda)^{L-1} \nonumber \\
&= \sum_{i=0}^L c_i (-\lambda)^i =0
\label{char A_1}
\end{align}
with $c_L=1$. Since $c_0,...,c_L$ are the coefficients of the characteristic polynomial for the distinct eigenvalues of the data covariance matrix, we now know the roots of Eq. (\ref{char A_1}). Therefore, the eigenvalues of $\bm{A}_1$ are given by the negative eigenvalues of the distinct eigenvalues of the data covariance matrix $\lambda_{A_1,l}= -{\lambda}_{l}$ for $1\leq l \leq L$. By applying the matrix $\bm{A}_1$ on a potential eigenvector $\bm{A}_1 \bm{v}_k= \lambda_k \bm{v}_k$, we find the following conditions for the eigenvector entries 
\begin{align}
v_{k,i}=(-1)^{i-1} \lambda_{k}^{i-1} v_{k,1},
\end{align}
obtained by a recursive method. Furthermore, we can choose $v_{k,1}=1$ for all eigenvectors. The eigenvector matrix $\bm{V}$, for which an eigenvector gives each column, is given by the transpose of the Vandermonde matrix

\begin{align}
\bm{V} = \begin{pmatrix}
1 & 1 & \cdots & 1 \\
\lambda_1 & \lambda_2 & \cdots & \lambda_L \\
\lambda_1^2 & \lambda_2^2 & \cdots & \lambda_L^2 \\
\vdots & \vdots & \ddots & \vdots \\
\lambda_1^{L-1} & \lambda_2^{L-1} & \cdots & \lambda_L^{L-1}
\end{pmatrix}.
\label{Vandermonde eigenmatrix}
\end{align}\\

Second, we evaluate all matrix products given in Eq. (\ref{formal sol R}). Since all eigenvalues are strictly negative, the student-teacher order parameters converge to $\lim_{\alpha \to \infty} \bm{R}\left(\alpha\right) = -\bm{A}_1^{-1} \bm{u} $ that we are going to evaluate using the eigenvector matrix. We insert the eigendecomposition $\bm{A}_1^{-1}=\bm{V} \bm{\Lambda} \bm{V}^{-1} $ into the asymptotic solution with $\Lambda_{kj}=-\delta_{lj} \frac{1}{\lambda_l} $ and find for the entries of the student-teacher order parameters
\begin{align}
\lim_{\alpha \to \infty} R_{i} &= -\sum_j^L V_{ij} \sum_k^L  \Lambda_{2,jk} \sum_l^L \left(V^{-1}\right)_{kl} u_{l} \nonumber \\
&= -\sum_j^L V_{ij}   \Lambda_{2,jj} \sum_l^L \left(V^{-1}\right)_{jl} u_{l} \nonumber \\
&= -\sum_j^L  \lambda_j^{(i-1)} \frac{1}{-\lambda_j} \sum_l^L \left(V^{-1}\right)_{jl} u_{l} \nonumber \\
&= \sum_j^L   \lambda_j^{(i-2)} \sum_l^L \left(V^{-1}\right)_{jl} u_{l} \nonumber \\
\end{align}
and further, evaluate
\begin{align}
\sum_l^L  \left(V^{-1}\right)_{jl} u_{l}  &= \sum_{l=1}^{L} \left(V^{-1}\right)_{jl} T^{(l)} \nonumber \\
&=\sum_{l=1}^{L} \left(V^{-1}\right)_{jl} \frac{1}{L} \sum_a^L \lambda_a^l \nonumber \\
&=\frac{1}{L} \sum_a^L \lambda_a \sum_{l=1}^{L} \left(V^{-1}\right)_{jl} \lambda_a^{l-1} \nonumber \\
&=\frac{1}{L} \sum_a^L \lambda_a \sum_{l=1}^{L} \left(V^{-1}\right)_{jl} V_{l,a} \nonumber \\
&=\frac{1}{L} \lambda_j \nonumber \\
\end{align}
where we have used $\sum_{l=1}^L  \left(V^{-1}\right)_{jl} V_{l,a}  = \delta_{j,a}$ obtained by the definition of the product between a matrix with its inverse. Thus, we find
\begin{align}
\lim_{\alpha \to \infty} R_{i}  = \frac{1}{L}\sum_j^L  \lambda_j^{(i-1)} =T^{(i-1)}.
\end{align}
Note that $R_{i}= R^{(i-1)}$ and therefore $ \lim_{\alpha \to \infty} R^{(i)}= T^{(i)}$. \\ 
Next, we want to evaluate $e^{\bm{A}_1 \alpha} \bm{A}_1^{-1} \bm{u}$ and define
\begin{align}
F_i &= \sum_j^L V_{ij} \sum_k^L  \exp\left(-\eta\lambda_j\alpha\right) \delta_{jk} \sum_l^L \Lambda_{kl} \sum_m^L \left(V^{-1}\right)_{lm}u_m \nonumber \\
&= \sum_j^L V_{ij}  \exp\left(-\eta\lambda_j\alpha\right) \frac{1}{-\lambda_j} \sum_m^L \left(V^{-1}\right)_{lm}u_m \nonumber \\
&= - \frac{1}{L} \sum_j^L \lambda_j^{i-2} \exp\left(-\eta\lambda_j\alpha\right) \lambda_j \nonumber \\
&= - \frac{1}{L} \sum_j^L  \exp\left(-\eta\lambda_j\alpha\right) \lambda_j^{i-1}   \nonumber \\
\end{align}
which leads to $\langle \bm{R}\left(\alpha\right)\rangle_{J_{a,0}, B_a}   = \bm{F} - \bm{A}_1^{-1} \bm{u} = \bm{T}+\bm{F} $. Thus, we obtain for the expectation value of $R^{(1)}$
\begin{align}
\langle R^{(1)} \rangle_{J_{a,0}, B_a} &= \langle R_{2} \rangle_{J_{a,0}, B_a}  \nonumber \\
&= 1-\frac{1}{L} \sum_a^L \exp\left(-\eta\lambda_a\alpha\right) \lambda_a.
\label{sol lin first order R }
\end{align}
\\
As a next step, we insert the result given by Eq. (\ref{formal sol R}) for $\langle \bm{R}_0\rangle =0  $ into the differential equations for the student-student order parameters given by Eq. (\ref{linear diff eq}), in order to obtain a new expression
\begin{align}
\frac{d\bm{Q}}{d\alpha} = \eta \bm{A}_3 \left( e^{\eta \bm{A}_1 \alpha} \bm{A}_1^{-1} \bm{u} - \bm{A}_1^{-1} \bm{u} \right) + \eta \bm{A}_4 \bm{Q} + \eta^2 \bm{u}
\label{diff lin Q 1}
\end{align}
In order to simplify the differential equation, we evaluate $\bm{A}_3 \bm{A}_1^{-1} \bm{T}$. The inverse of $\bm{A}_1$ can be obtained analytically, where we find
\begin{align}
\bm{A}_1^{-1}= \begin{pmatrix}
\frac{c_1}{c_0} & \frac{c_2}{c_0}       & \frac{c_3}{c_0}        & \cdots & \frac{c_{L-1}}{c_0}        &\frac{1}{c_0}  \\
-1       & 0 & 0     & \ddots & \vdots  & \vdots \\
0        & -1       & 0& \ddots & 0       & 0 \\
\vdots   & \vdots  & \ddots  & \ddots & 0      & 0 \\
0        & 0       & \cdots  & -1     & 0 & 0 \\
0     & 0     & 0    & \cdots & -1 & 0
\end{pmatrix} 
\end{align}
and obtain for the matrix-vector product $-\bm{A}_3 \bm{A}_1^{-1} \bm{u}= 2\left(\eta-1\right) \bm{u}$. We can further simplify $\eta \bm{u}-\bm{A}_3 \bm{A}_1^{-1} \bm{u} = \left(\eta-2\right) \bm{u}$ and insert this result in Eq. (\ref{diff lin Q 1}) in order to obtain
\begin{align}
\frac{d\bm{Q}}{d\alpha} = \eta \bm{A}_3 e^{\bm{A}_1 \alpha} \bm{A}_1^{-1} \bm{u} + (2-\eta) \bm{u} + \eta \bm{A}_4 \bm{Q}.
\label{diff lin Q 2}
\end{align}
The solution of this differential equation is given by
\begin{align}
 \bm{Q}\left(\alpha\right) &= e^{\eta\bm{A}_4 \alpha}\bm{Q}_0+e^{\eta\bm{A}_4 \alpha} (\bm{A}_1 - \bm{A}_4)^{-1} \left(e^{(\eta\bm{A}_1 - \eta\bm{A}_4) \alpha} - \bm{I}\right) \bm{A}_3 \bm{A}_1^{-1} \bm{u}  \\
 & \quad+ (2-\eta) e^{\eta\bm{A}_4 \alpha} \bm{A}_4^{-1} \left(\bm{I} - e^{-\eta\bm{A}_4 \alpha}\right)\bm{u} 
 \label{sol lin Q 1}
\end{align}
with $\bm{Q}_0$ are the student-student order parameters at initialization. Note that the initial value $\bm{Q}_0$ does not vanish in the thermodynamic limit and is also not zero on average in contrast to $\bm{R}_0$. By the definition of the matrices, we find $\bm{A}_1 - \bm{A}_4= \frac{1}{2} \bm{A}_3$ making it straightforward to evaluate their relations $(\bm{A}_1 - \bm{A}_4)^{-1} \bm{A}_3= 2 \bm{I}_L $. Furthermore, in order to estimate $(2-\eta) \bm{A}_4^{-1} \bm{u}$, we need to know the inverse of $\bm{A}_4$ for which we find
\begin{align}
\bm{A}_4^{-1}= \begin{pmatrix}
\frac{c_1}{2c_0}-\frac{\eta}{2-\eta} & \frac{c_2}{2c_0}       & \frac{c_3}{2c_0}        & \cdots & \frac{c_{L-1}}{2c_0}        &\frac{1}{2c_0}  \\
\frac{1}{\eta-2}      & 0 & 0     & \ddots & \vdots  & \vdots \\
\frac{\eta T^{(2)}}{2(\eta-2)}         & -\frac{1}{2}       & 0& \ddots & 0       & 0 \\
\vdots   & \vdots  & \ddots  & \ddots & 0      & 0 \\
\frac{\eta T^{(L-2)}}{2(\eta-2)}        & 0       & \cdots  & -\frac{1}{2}     & 0 & 0 \\
\frac{\eta T^{(L-1)}}{2(\eta-2)}     & 0     & 0    & \cdots & -\frac{1}{2}  & 0
\end{pmatrix} .
\end{align}
Thus, we obtain for the product $(2-\eta) \bm{A}_4^{-1} \bm{u} =-\bm{T} $. Therefore, we can simplify Eq. (\ref{sol lin Q 1}) to
\begin{align}
 \bm{Q}\left(\alpha\right) &= \bm{T}+e^{\eta \bm{A}_4 \alpha}\bm{Q}_0+\left( e^{\eta \bm{A}_4 \alpha} - 2e^{\eta \bm{A}_1 \alpha} \right)\bm{T}
 \label{sol lin Q 2}
\end{align}

\subsubsection{small learning rates}
For small learning rates, we can approximate $\bm{A}_4 \approx 2 \bm{A}_1$ and immediately find for the first order student-student order parameters
\begin{align}
\langle Q^{(1)} \rangle_{J_{a,0}, B_a} &= \langle Q_{2} \rangle_{J_{a,0}, B_a}  \nonumber \\
&= 1+\left(1+\sigma_J^2\right)\frac{1}{L} \sum_a^L \exp\left(-2\eta\lambda_a\alpha\right) \lambda_a - 2\frac{1}{L} \sum_a^L \exp\left(-\eta\lambda_a\alpha\right) \lambda_a,
 \label{sol lin first order Q }
\end{align}
where we have exploit that $\langle \bm{Q}_0 \rangle_{J_{a,0}, B_a}= \sigma_J^2 \bm{T}$. After inserting Eqs. (\ref{sol lin first order R }) and (\ref{sol lin first order Q }) into Eq. (\ref{eg linear}), we obtain for the generalization error
\begin{align}
\langle \epsilon_g \rangle_{J_{a,0}, B_a} =  \left(1+\sigma_J^2\right) \frac{1}{2 L} \sum_{a=1}^L \lambda_a \exp(-2\eta \lambda_a \alpha) ,
\label{eps perceptron time 2}
\end{align}
leading to Eq. (\ref{eps perceptron time}) in the main text.\\

\subsection{Scaling with time}
\label{Euler-Maclaurin with time}
In order to evaluate the sum for the generalization error given in Eq. (\ref{eps perceptron time 2}), we consider the Euler-Maclaurin approximation. The formula for the approximation is given by
\begin{equation}
\sum_{i=m}^{n} f(i) = \int_{m}^{n} f(x)\, dx + \frac{f(n) + f(m)}{2} + \sum_{k=1}^{\left\lfloor \frac{p}{2} \right\rfloor} \frac{B_{2k}}{(2k)!} \left(f^{(2k-1)}(n) - f^{(2k-1)}(m)\right) + R_p,
\label{Euler Mclaurin}
\end{equation}
where \( B_{2k} \) is the \( k \)th Bernoulli number. For the difference of the sum from the integral, one can find 
\[
\begin{aligned}
\sum_{i=m}^{n}f(i) - \int_{m}^{n} f(x)\,dx &= \frac{f(m) + f(n)}{2} + \int_{m}^{n} f'(x) P_1(x)\,dx \\
&= \frac{f(m) + f(n)}{2} + \frac{1}{6} \frac{f'(n) - f'(m)}{2!} + \int_{m}^{n} f'''(x) \frac{P_3(x)}{3!}\,dx,
\end{aligned}
\]
where \( P_k(x) \) are the periodized Bernoulli polynomials of order \( k \). These are defined as:
\[
P_k(x) = B_k(x - \lfloor x \rfloor),
\]
where \( B_k(x) \) are the Bernoulli polynomials and \( \lfloor x \rfloor \) denotes the floor function. For our case, the function and its derivative are
\[
f(k, \alpha, \beta) = \frac{\left(1+\sigma_J^2\right)}{L} \frac{\lambda_+ }{k^{\beta+1}} \exp\left(-\frac{2\eta \lambda_+ \alpha}{k^{\beta+1}}\right)
\]
\[
f'(k) = \frac{\left(1+\sigma_J^2\right)\lambda_+ }{L} \exp\left(-\frac{2\eta \lambda_+ \alpha}{k^{\beta + 1}}\right) \left( \frac{2\eta \lambda_+ \alpha (\beta + 1)}{ k^{2\beta + 3}} -\frac{\beta + 1}{k^{\beta + 2}}  \right).
\]
We want to approximate the sum by the integral and, therefore, estimate their maximal difference. Due to the exponential pre-factor in all functions, the maximal deviation of the integral from the sum is obtained at initialization \( \alpha=0 \) or for \( \eta \to 0 \). Therefore, we make the following ansatz
\[
\sum_{i=m}^{n}f(i) - \int_{m}^{n} f(x)\,dx < \lim_{\eta \to 0} \frac{f(m) + f(n)}{2} + \int_{m}^{n} f'(x) P_1(x)\,dx.
\]
Our goal is to express the term \( \int_{m}^{n} f'(x) P_1(x)\,dx \) for \( \eta \to 0 \), which represents the correction to the integral, in a more explicit form. To do this, we consider the integral over the interval \([k, k+1)\), where \( k \) is an integer. This allows us to handle the floor function more easily, as it is constant over each interval. Within each interval, we can simplify the expression for \( \int_{k}^{k+1} f'(x) P_1(x)\,dx \), which can be written as:
\[
\Delta_k = \frac{L }{\left(1+\sigma_J^2\right)\lambda_+} \lim_{\eta \to 0} \int_{k}^{k+1} f'(x) P_1(x)\,dx.
\]
Substituting the expression for \( f'(x) \) and taking the limit as \( \eta \to 0 \), \( \Delta_k \) becomes:
\[
\Delta_k = \frac{1}{2(k + 1)^{\beta + 1}} + \frac{1}{2k^{\beta + 1}} + \frac{1}{\beta(k + 1)^\beta} - \frac{1}{\beta k^\beta}.
\]
This expression for \( \Delta_k \) provides a simplified form for the correction term associated with the Euler-Maclaurin approximation in the limit as \( \eta \) approaches zero. To estimate the total correction across the entire interval from \( 1 \) to \( L \), we sum this expression from \( k = 1 \) to \( L-1 \) for large $L$. For the first term, we obtain
by reindexing
\[
\frac{1}{2}\sum_{k=1}^{L-1} \frac{1}{(k + 1)^{\beta + 1}} = \frac{1}{2} \zeta(\beta + 1) - \frac{1}{2},
\]
where \( \zeta(\beta + 1) \) is the Riemann zeta function evaluated at \( \beta  + 1 \). The second term is straightforward and sums to:
\[
\frac{1}{2}\sum_{k=1}^{L-1} \frac{1}{k^{\beta + 1}} = \frac{1}{2} \zeta(\beta + 1).
\]
For the difference of the third term from the fourth, we find
\[
\sum_{k=1}^{L-1} \frac{1}{\beta(k + 1)^\beta} -\frac{1}{\beta k^\beta} = -\frac{1}{\beta}
\]
Combining all these results, the sum of the integral correction term over all intervals from \( 1 \) to \( L-1 \) is:
\[
\sum_{k=1}^{L-1} \Delta_k =\zeta(\beta + 1) - \frac{1}{2} - \frac{1}{\beta}
\]
where \( \zeta(\beta + 1) \) is the Riemann zeta function evaluated at \( \beta + 1 \).\\
Thus, we can upper bound the error by 
\[
\sum_{i=m}^{n}f(i) - \int_{m}^{n} f(x)\,dx < \frac{\left(1+\sigma_J^2\right)\lambda_+ }{2L} \left[ \frac{1}{L^{\beta+1}}+ \zeta(\beta  + 1) -\frac{1}{\beta }\right] \\
\]
This "worst case" upper bound works excellent for moderate input sizes $N \sim \mathcal{O}\left(10^1\right)$ as well.\\
The behavior of the difference with $\alpha$ and finite $\eta$ can also be estimated. If we consider a learning rate larger than $\eta>\frac{1}{2\lambda_+ \alpha}$, which can still be seen as small since $\lambda_+ \sim N$ for $\alpha  \gtrsim  1$, then the difference will be mainly governed by the $0$th order contributions. Especially $f(k=1)$ plays the most important role and we find
\begin{align}
\sum_{i=m}^{n}f(i) - \int_{m}^{n} f(x)\,dx < \frac{1}{2} f(1) = \frac{\left(1+\sigma_J^2\right)\lambda_+ }{2L}  \exp\left(-2\eta \lambda_+ \alpha\right).
\label{0th order correc}
\end{align}

\begin{figure}
  \centering
  \includegraphics[width=0.49\linewidth]{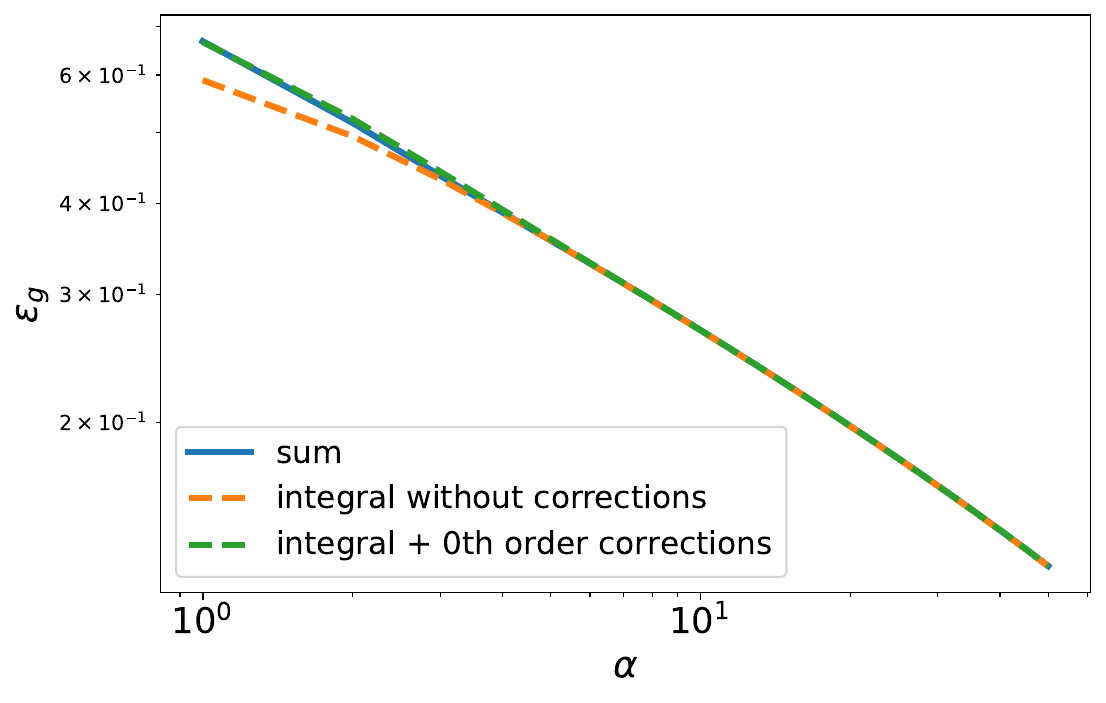}
  \includegraphics[width=0.49\linewidth]{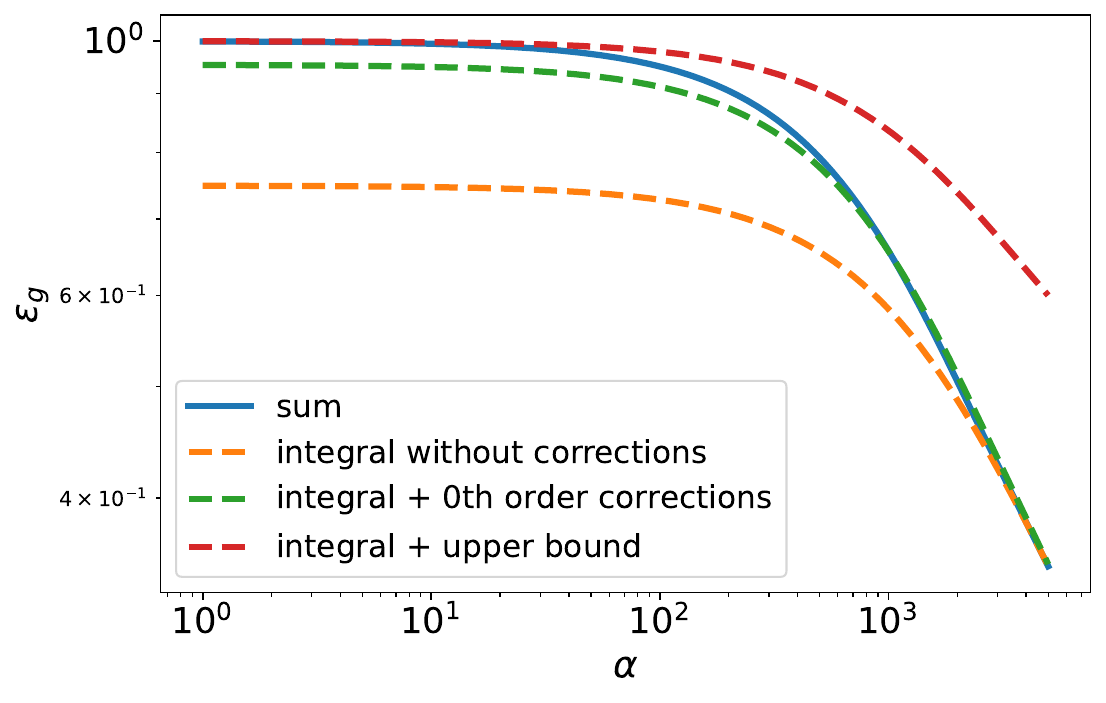}
  \caption{Generalization error evaluated by Eq. (\ref{eps perceptron time 2}) (full blue) and Eq. (\ref{Euler Mclaurin}) including different order of corrections (dashed) for $N=128$ , $K=M=1$, $\sigma_J^2=1$ and $\beta=0.5$. Left:  For $\eta=\frac{1}{2\lambda_+} \approx 0.001$, the upper bound of the error can be tightened and is given by the $0$th order corrections. Right: For learning rates $\eta<\frac{1}{2\lambda_+}$ the upper bound includes higher order corrections up to the worst case bound for small $\alpha < \alpha= \frac{1}{2\eta\lambda_+}$ (red dashed). However, for $\alpha> \frac{1}{2\eta\lambda_+}$, the $0$th order upper bound is again valid. Here, we have chosen $\eta=0.00001$ and therefore the $0$th order begins to be valid at $\alpha= \frac{1}{2\eta\lambda_+} \approx 950$. }
  \label{fig: euler corrections}
\end{figure}

Since we have chosen $\eta>\frac{1}{2\lambda_+ \alpha}$, the difference decreases exponentially in time. For smaller values of the learning rate $\eta<\frac{1}{2\lambda_+ \alpha}$ higher order must be included up to the worst case bound as given in Figure \ref{fig: euler corrections}. However, for the condition $\eta<\frac{1}{2\lambda_+ \alpha}$ or equivalently $\alpha<\frac{1}{2\eta\lambda_+}$ no meaningful learning has occurred. 

\begin{figure}
  \centering
  \includegraphics[width=0.49\linewidth]{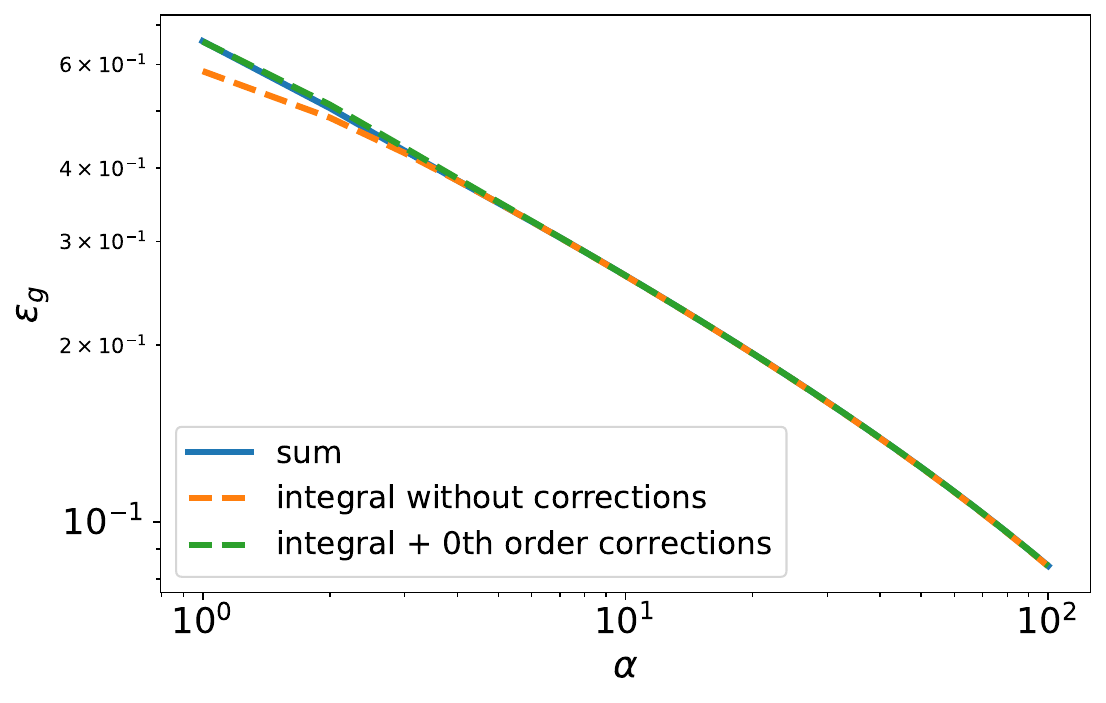}
  \includegraphics[width=0.49\linewidth]{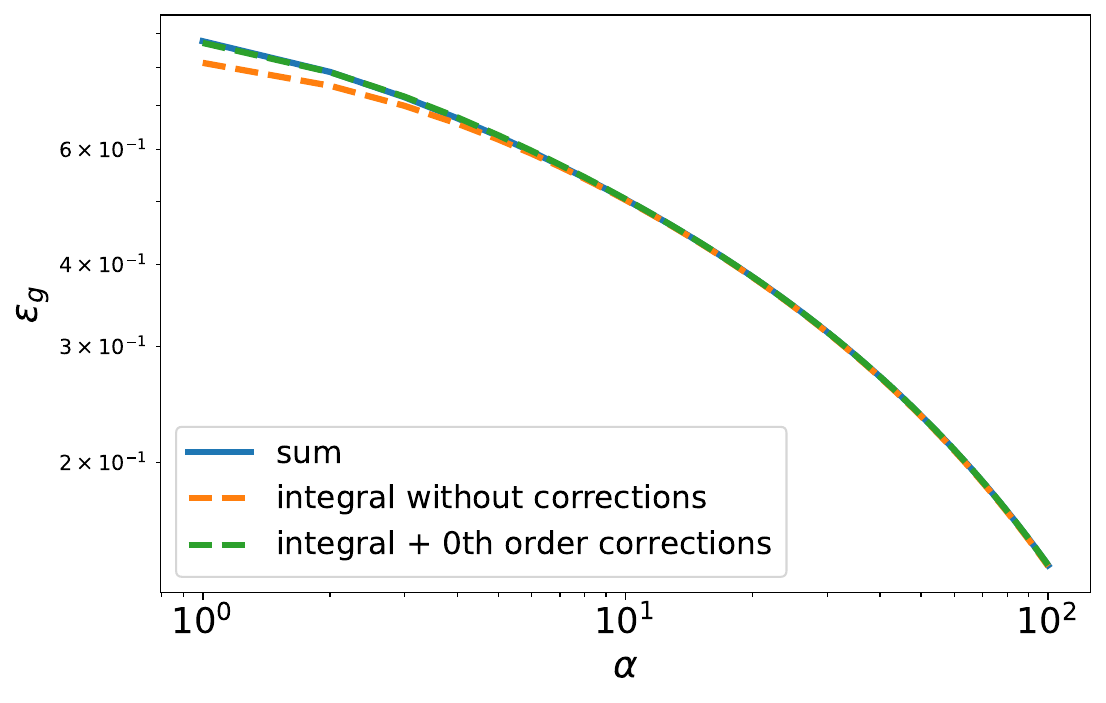}
  \caption{Generalization error evaluated by Eq. (\ref{eps perceptron time 2}) (full blue) and Eq. (\ref{Euler Mclaurin}) including different order of corrections (dashed) for $N=128$ , $K=M=1$, $\sigma_J^2=1$ and $\eta=0.01$  Left:  For $\beta=0.5$. Right: For $\beta=0.01$. }
  \label{fig: euler corrections 2}
\end{figure}

In the right plot of Figure \ref{fig: euler corrections}, we have chosen $\eta=0.00001$ in order to test the boundaries of our approximation. However, such a learning rate is very untypical since no learning would occur over many time orders. Typically, the learning rates that we use in our simulations are larger than $0.001$, and $0$th order corrections are enough to consider (cf. Figure \ref{fig: euler corrections 2}). \\
 
Since the $0$th order corrections decrease exponentially with $\alpha$, we use the integral in order to approximate the generalization error $\epsilon_g(L;\alpha) \approx \int_{m}^{n} f(x)\,dx $ for $m=1$ to $n=L$. This integral can be solved analytically and we obtain
  \begin{align}
 \epsilon_g(L;\alpha) \approx  \frac{\left(1+\sigma_J^2\right)\lambda_+}{2L}\frac{\left(2\eta\lambda_+ \alpha\right)^{-\frac{\beta}{1+\beta}}}{1+\beta} \Biggl[ \Gamma \left(\frac{\beta}{1+\beta}, \frac{2\eta\lambda_+ \alpha}{L^{\beta+1}}\right)-\Gamma \left(\frac{\beta}{1+\beta}, 2\eta\lambda_+ \alpha\right)\Biggr]
 \label{approx eps 2}
 \end{align}
where $\Gamma \left(s, x \right)$ is the incomplete gamma function. In the expression of the generalization error, we can observe a power-law factor and want to clear under which conditions the generalization error and the function given by Eq. (\ref{approx eps 2}) shows a scaling according to $\epsilon_g \sim  \alpha^{-\frac{\beta}{1+\beta}} $. For this, we note that the incomplete gamma function $\Gamma(s,z)$ can be approximated by
\begin{align}
\Gamma(s,z) \approx \Gamma(s)- \frac{z^s}{s}
\label{approx gamma}
\end{align}
for $z \ll 1$. For our setup, we identify $s=\frac{\beta}{\beta+1}$ and $z=\frac{2\eta\lambda_+\alpha}{L^{1+\beta}}$ for the first $\Gamma$ function in the brackets given in Eq. (\ref{approx eps 2}).

\begin{figure}
  \centering
  \includegraphics[width=0.49\linewidth]{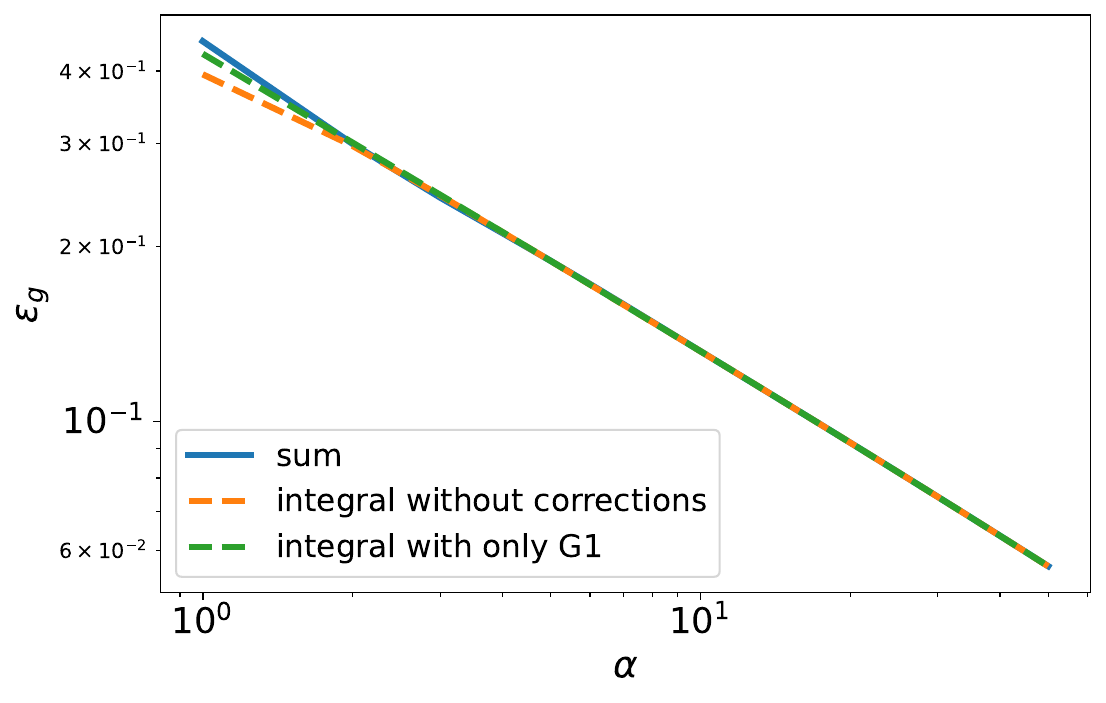}
  \includegraphics[width=0.49\linewidth]{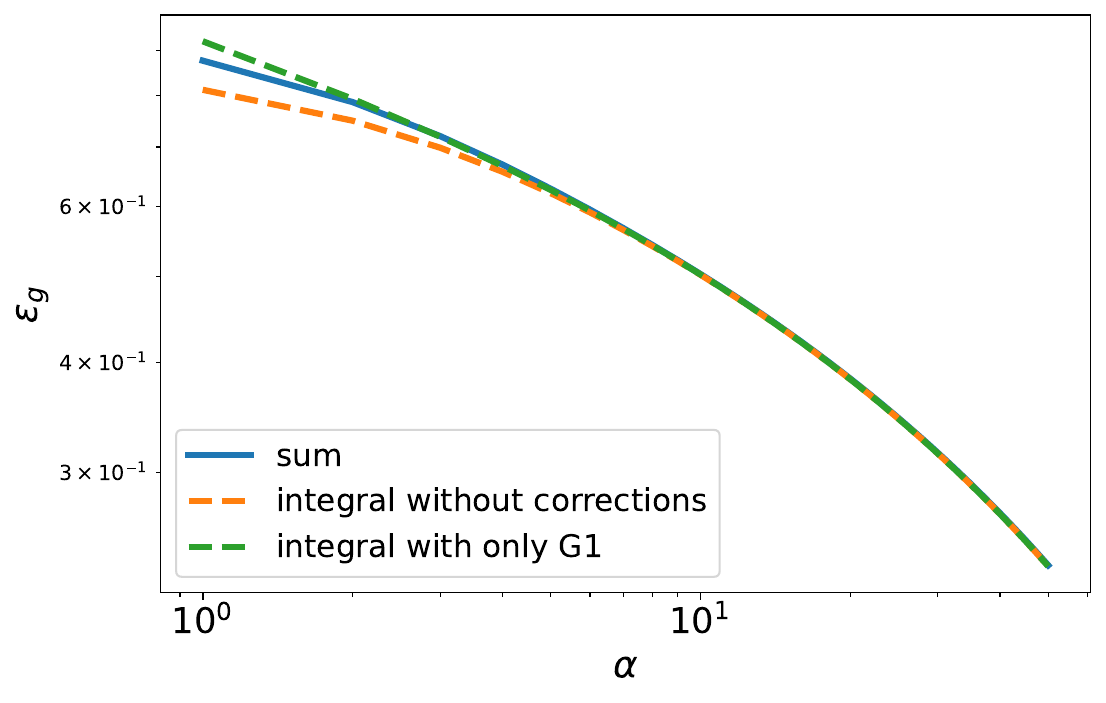}
  \caption{Generalization error evaluated by Eq. (\ref{eps perceptron time 2}) (full blue), Eq. (\ref{approx eps 2}) (dashed orange)  and (\ref{approx eps 3}) (dashed green) for $N=128$ , $K=M=1$, $\sigma_J^2=1$ and $\eta=0.01$.  Left:  For $\beta=1$. Right: For $\beta=0.01$. }
  \label{fig: euler corrections 3}
\end{figure}

For the second gamma function within the brackets in (\ref{approx eps 2}), we note that its argument is $L^{1+\beta}$ larger than for the first gamma function. Furthermore, from the previous discussion based on empirical observations, we know that meaningful learning happens for $\alpha > \frac{1}{2\eta\lambda_+}$. Thus, we can introduce a scaled time variable by $\tilde{\alpha}=2\eta\lambda_+\alpha$ and insert this into the second gamma function $\Gamma\left(\frac{\beta}{1+\beta},2\eta\lambda_+\alpha\right)=\Gamma\left(\frac{\beta}{1+\beta},\tilde{\alpha}\right)$ decreasing exponentially fast with $\tilde{\alpha}$. Therefore, we can neglect the second gamma function compared to the first one since both operate in different time scales as presented in Figure \ref{fig: euler corrections 3}. Note that the condition $\alpha > \frac{1}{2\eta\lambda_+}$ ensures that the zeroth-order contributions from the gamma function do not cancel out. For very small learning rates ($\eta \to 0$), both gamma functions can be approximated by Eq. (\ref{approx gamma}), where the zeroth-order contributions are identical ($\Gamma(s)$). Thus, our initially empirical observation $\alpha > \frac{1}{2\eta\lambda_+}$ now has analytical justification. Thus, we further simplify
  \begin{align}
 \epsilon_g(L;\alpha) \approx  \frac{\left(1+\sigma_J^2\right)\lambda_+}{2L}\frac{\left(2\eta\lambda_+ \alpha\right)^{-\frac{\beta}{1+\beta}}}{1+\beta}  \Gamma \left(\frac{\beta}{1+\beta}, \frac{2\eta\lambda_+ \alpha}{L^{\beta+1}}\right)
 \label{approx eps 3}
 \end{align}
valid for $\frac{1}{2\eta \lambda_+}<\alpha$.\\
Thus, for the condition $\Gamma(s)>\frac{z^s}{s}$, we obtain a power-law scaling for the generalization error. This condition is fulfilled for $\alpha<\frac{1}{2\eta \lambda_+} \Gamma\left(\frac{2\beta+1}{1+\beta}\right)^{\frac{1+\beta}{\beta}} L^{1+\beta}$. \\
In conclusion, the generalization error shows a power-law scaling for the condition $\frac{1}{2\eta \lambda_+}<\alpha<\frac{1}{2\eta \lambda_+} \Gamma\left(\frac{2\beta+1}{1+\beta}\right)^{\frac{1+\beta}{\beta}} L^{1+\beta}$. Note that $ \Gamma\left(\frac{2\beta+1}{1+\beta}\right)^{\frac{1+\beta}{\beta}} \lesssim 1$.  Inserting the approximation given by Eq. (\ref{approx gamma}) into Eq. (\ref{approx eps 2}) and neglecting the second gamma term, we obtain
  \begin{align}
 \epsilon_g(L;\alpha) \approx  \frac{\left(1+\sigma_J^2\right)\lambda_+}{2 L}\frac{\left(2\eta\lambda_+ \alpha\right)^{-\frac{\beta}{1+\beta}}}{1+\beta} \Gamma\left(\frac{\beta}{1+\beta}\right)
 \label{approx eps 4}
 \end{align}
We observe that the power-law range is extended with an increasing number of distinct eigenvalues and therefore with the input dimension $N$ in practice and covariance matrix power-law exponent $\beta$ which can be seen very easily for the rescaled version $1 < \tilde{\alpha}< \Gamma\left(\frac{2\beta+1}{1+\beta}\right)^{\frac{1+\beta}{\beta}} L^{1+\beta}$.

\subsubsection{general learning rate}
Here, we reconsider Eq. (\ref{sol lin Q 2}) and want to evaluate the student-student order parameters not in the small learning rate limit. For this, we have to understand how the eigenvectors of $\bm{A}_4$ influence $\bm{T}$. We call the eigenvector matrix of $\bm{A}_4$ simply $\bm{V}_4$ which contains all eigenvectors as its columns and introduce the eigendecomposition $e^{\bm{A}_4 \alpha}= \bm{V}_4 e^{\eta \bm{\Lambda_4}\alpha} \bm{V}_4^{-1} $ with $\bm{\Lambda_4}$ containing the eigenvalues of $\bm{A}_4$ on its diagonal. To find the eigenvalues of $\bm{A}_4$ either analytically or numerically is no longer easy. However, we can pretend to know the eigenvalues of $\bm{A}_4$ and call them $\lambda_{4,k}$ for $1\leq k \leq L$. This makes it possible to find the structure of the eigenvector matrix. \\
Here, we present a more general solution for the matrix $\bm{B}= a\bm{A}_1+\epsilon \bm{U}$ and call the eigenvalues of $\bm{B}$ simply $\lambda_{B,k}$ and define the eigenvectors $\bm{B}\bm{v}_{B,k} = \lambda_{B,k} \bm{v}_{B,k}$ for $k \in \{1,...,L\}$. Note that for $a=2$ and $\epsilon=\eta$, we can reproduce $\bm{B}\left(a=2, \epsilon=\eta\right)=\bm{A}_4$. For each $\lambda_{B,k}$, the corresponding $k$-th eigenvector has the following structure
\[
v_{B,1k} = 1, \quad v_{B,2k} = \frac{\lambda_{B,k}}{\epsilon - a}, \quad v_{B,3k} = \frac{1}{a} \frac{1}{ \epsilon - a} (\epsilon T^2 - \lambda_{B,k}),
\]
\[
v_{B,ik} = -\frac{1}{a^{L-2} (\epsilon - a)} \left( -\lambda_{B,k}^{L-1} + \epsilon \sum_{i=1}^{L-2} (-1)^i a^{L-2-i} \lambda_{B,k}^i T^{L-i} \right).
\]
The eigenvectors have an interesting property. Like for the companion matrix $\bm{A}_1$ and its Vandermonde eigenvector matrix $\bm{V}_1$, the first $l-1$ entries of the eigenvalue equation $Bv_{B,k} = \lambda_{B,k} v_{B,k}$ give a zero by construction $\left(Bv_{B,k} - \lambda_{B,k} v_{B,k}\right)_l=0$ for $l=1,...,L-1$. Only the last entry $l=L$ of the eigenvalue equation provides information about the eigenvalues of $\bm{B}$
\[
(\bm{B} \bm{v}_{B,k} - \lambda_{B,k} \bm{v}_{B,k})_L = \sum_{j=1}^L c_{B,j} \lambda_{B,k}^j.
\]
This is exactly the characteristic polynomial for $\bm{B}$ where we have identified 
\begin{align}
c_{B,0} &= a c_0, \quad c_{B,L} = \frac{(-1)^{L+1}}{a^{L-2}} \frac{1}{\epsilon - a} \\
c_{B,j} &= \frac{1}{a^{j-1}} \frac{(-1)^{j+1} }{\epsilon - a} \left(ac_j + \epsilon \sum_{l=0}^{L-1-i} c_{L-l} T^{(L+1-j-l)}\right)
\label{coeff poly B}
\end{align}
as the shifted coefficients of $\bm{B}$ compared the coefficients $c_j$ of $\bm{A}_1$ for $j=1,...,L-1$. Since we have assumed that $\lambda_{B,k}$ is an eigenvalue of $\bm{B}$, the right-hand side of Eq. has to be zero, and the $c_{B,k}$ are indeed the new coefficients. Since we know the shifted coefficients, we can numerically calculate the corresponding eigenvalues as the roots of the new characteristic polynomial.

\begin{figure}
  \centering
  \includegraphics[width=0.49\linewidth]{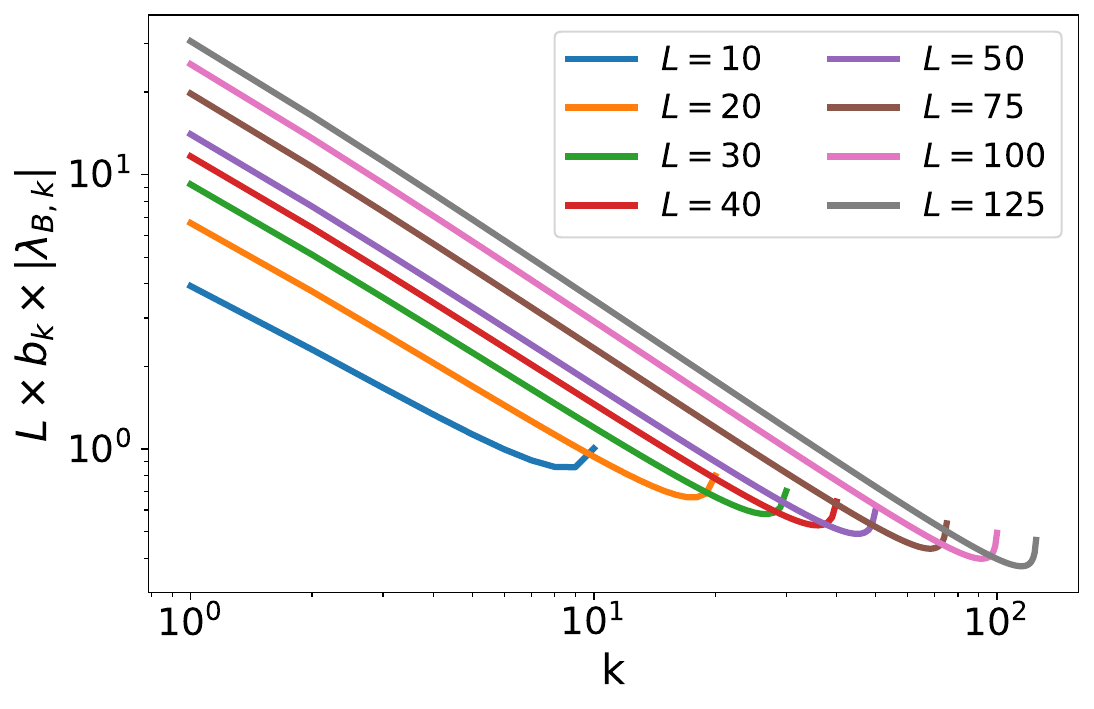}
  \includegraphics[width=0.49\linewidth]{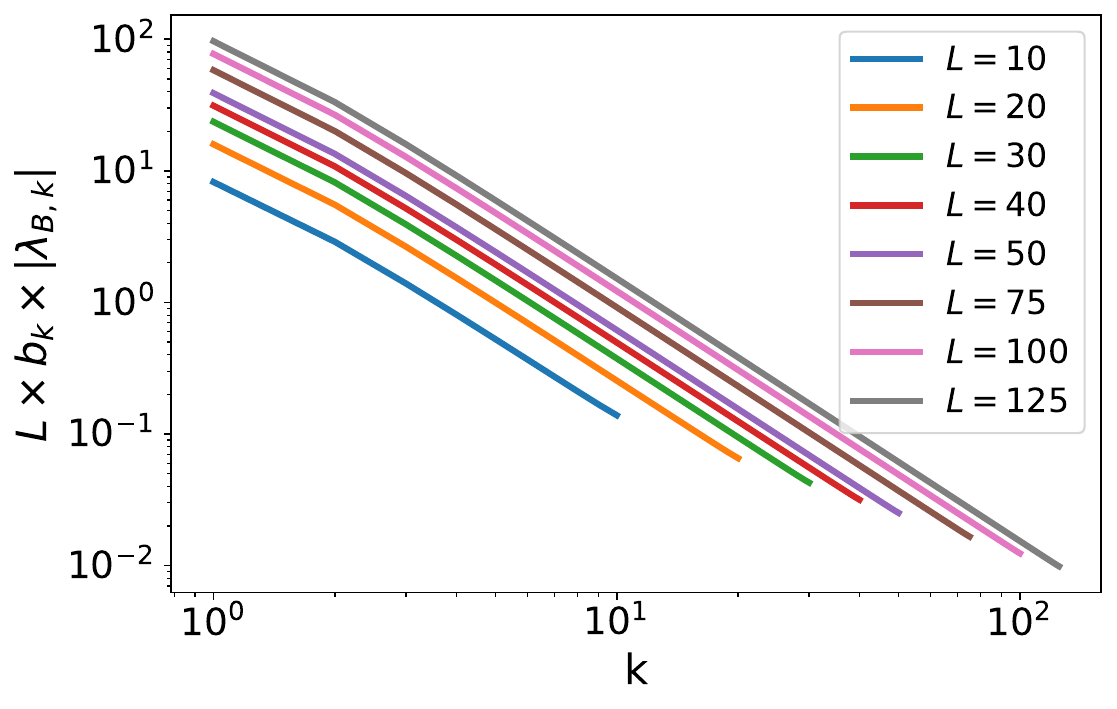}
  
  \caption{Coefficients $b_k$ defined by Eq. (\ref{b_k}) multiplied with $\lambda_{B,k}$ for $\eta=1.0$. Left: $\beta=0.1$. Right:  $\beta=1.0$.}
  \label{fig: mult 2}
\end{figure}

\begin{figure}
  \centering
  \includegraphics[width=0.49\linewidth]{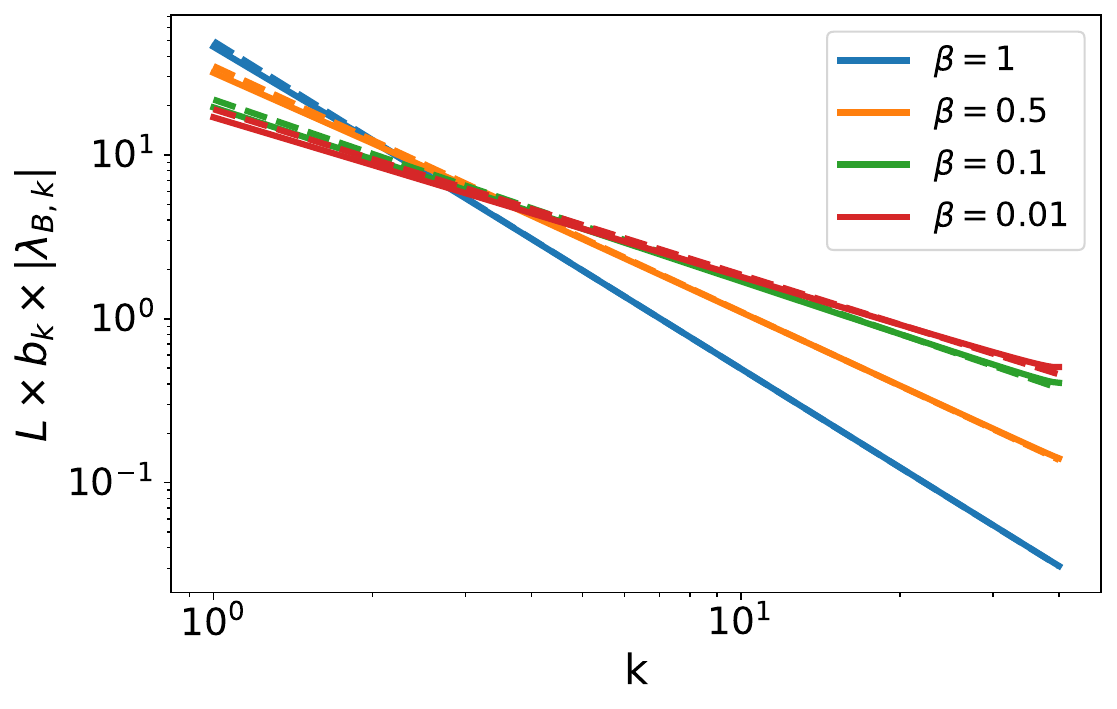}
  \includegraphics[width=0.49\linewidth]{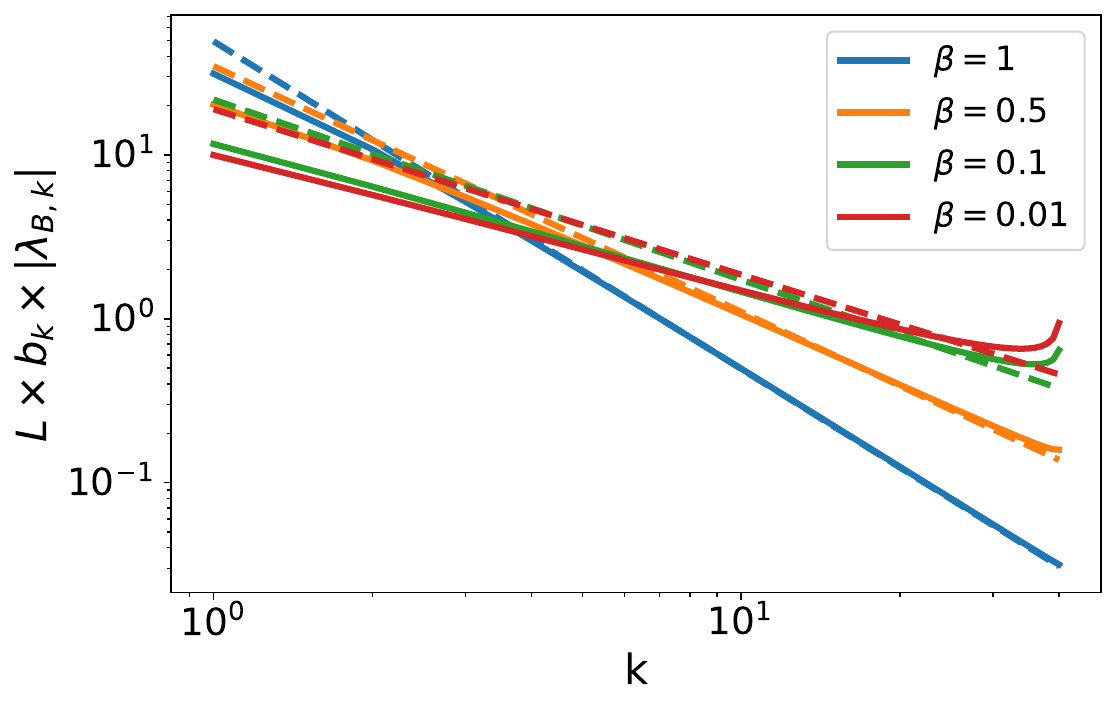}
  \includegraphics[width=0.49\linewidth]{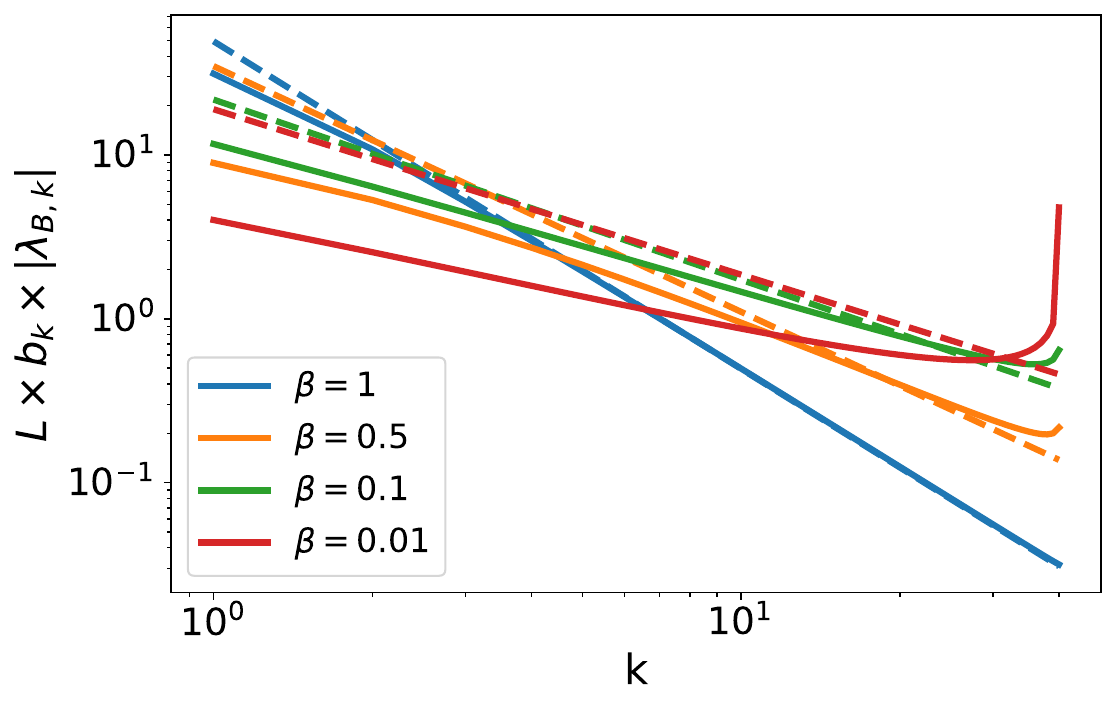}
  
  \caption{Coefficients $b_k$ defined by Eq. (\ref{b_k}) multiplied with $\lambda_{B,k}$ and $L$ for $L=40$ (solid) and eigenvalues of the data covariance matrix (dashed). Upper left: $\eta=0.1$. Upper right: $\eta=0.5$. Center: $\eta=1.0$.}
  \label{fig: mult 1}
\end{figure}

\begin{figure}
  \centering
  \includegraphics[width=0.49\linewidth]{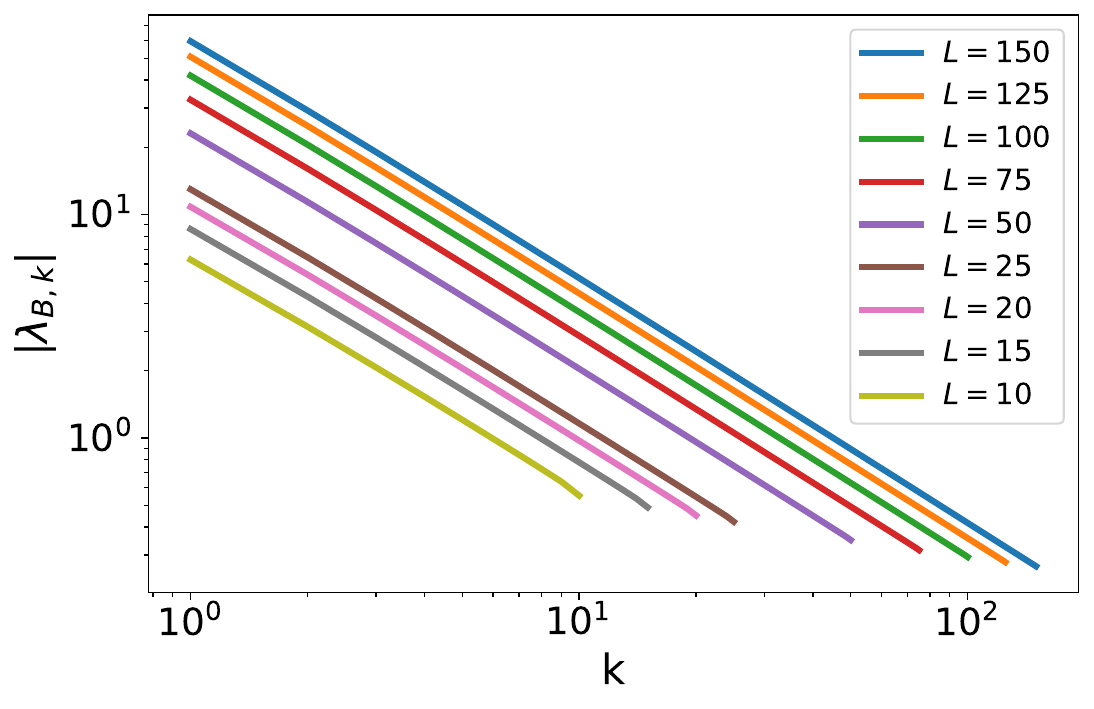}
  \includegraphics[width=0.49\linewidth]{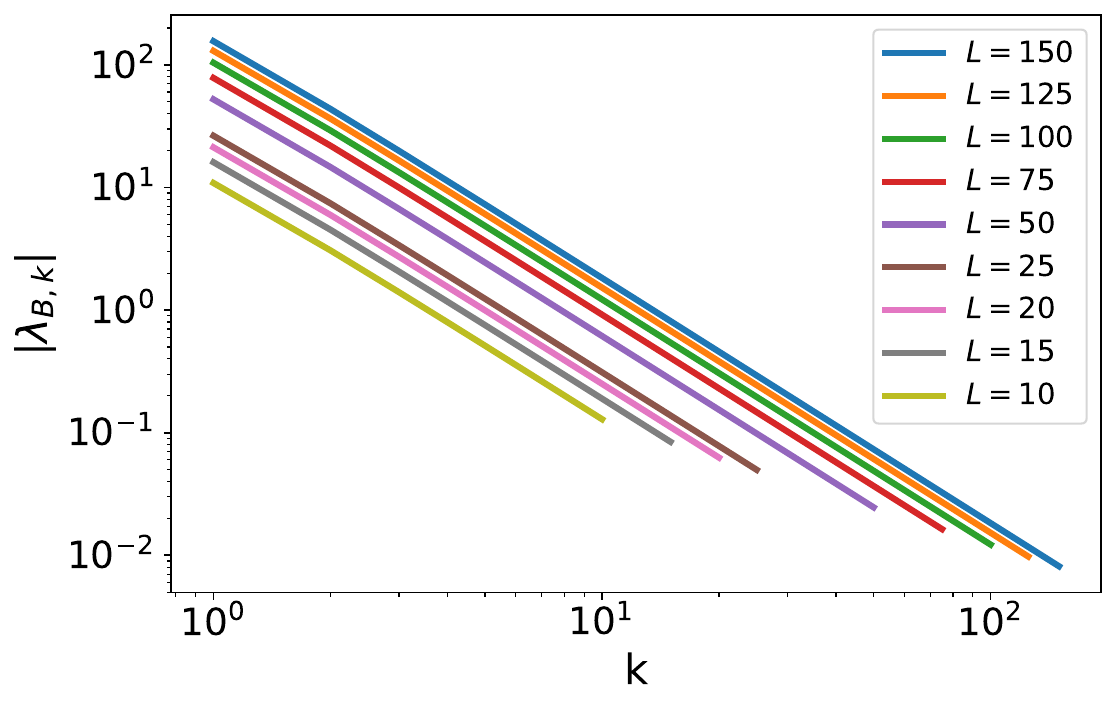}\\
  \includegraphics[width=0.49\linewidth]{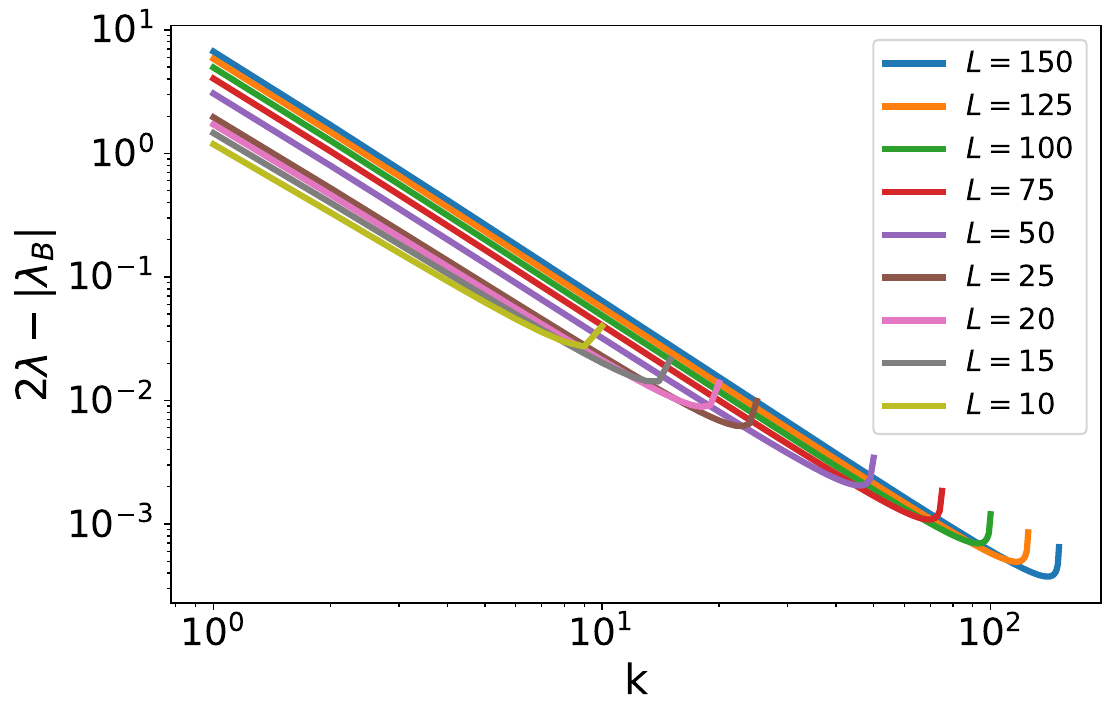}
   \includegraphics[width=0.49\linewidth]{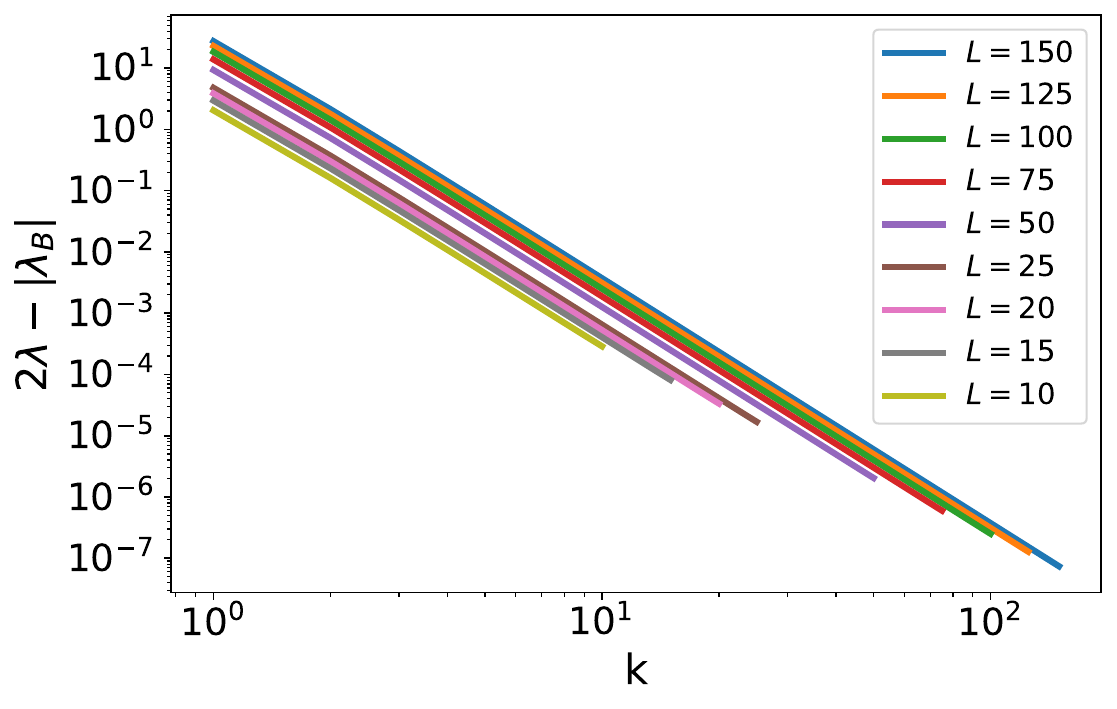}
 
  \caption{Spectrum of matrix $\bm{B}$ for $a=2$ (upper figures) and corresponding difference with the spectra of the data covariance matrix. Left: $\epsilon=1$ and $\beta=0.1$. Right: $\epsilon=0.5$ and $\beta=1.0$.  }
  \label{fig: Julia linear eigenvalues 1}
\end{figure}

\begin{figure}
  \centering
  \includegraphics[width=0.49\linewidth]{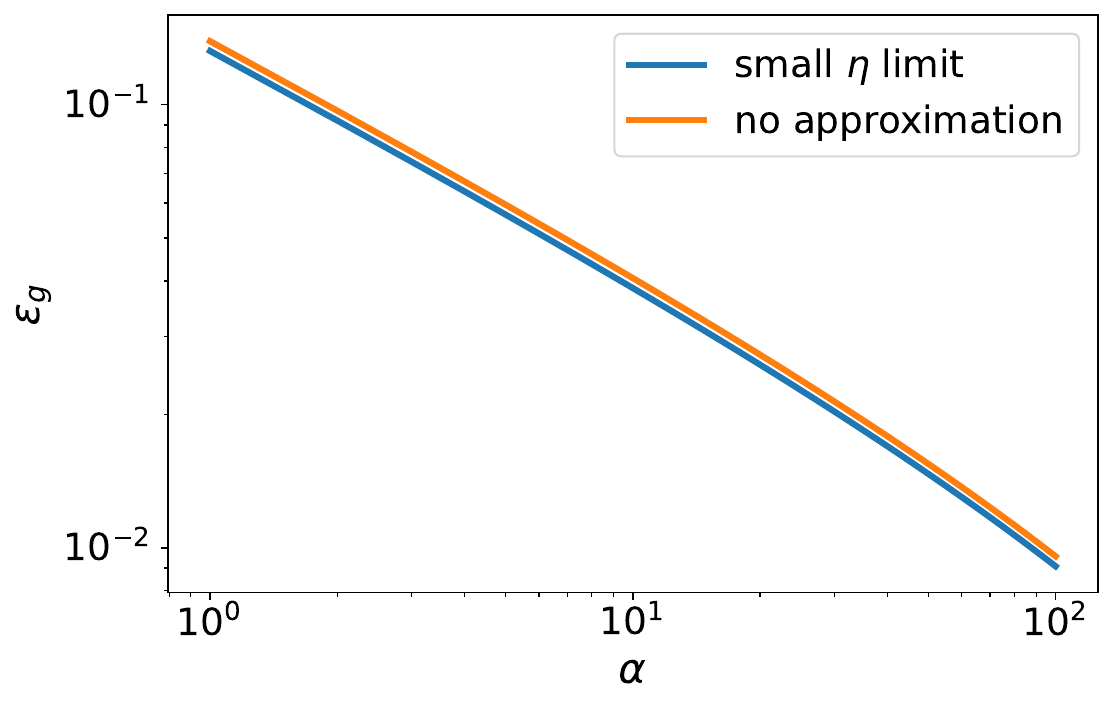}
  \includegraphics[width=0.49\linewidth]{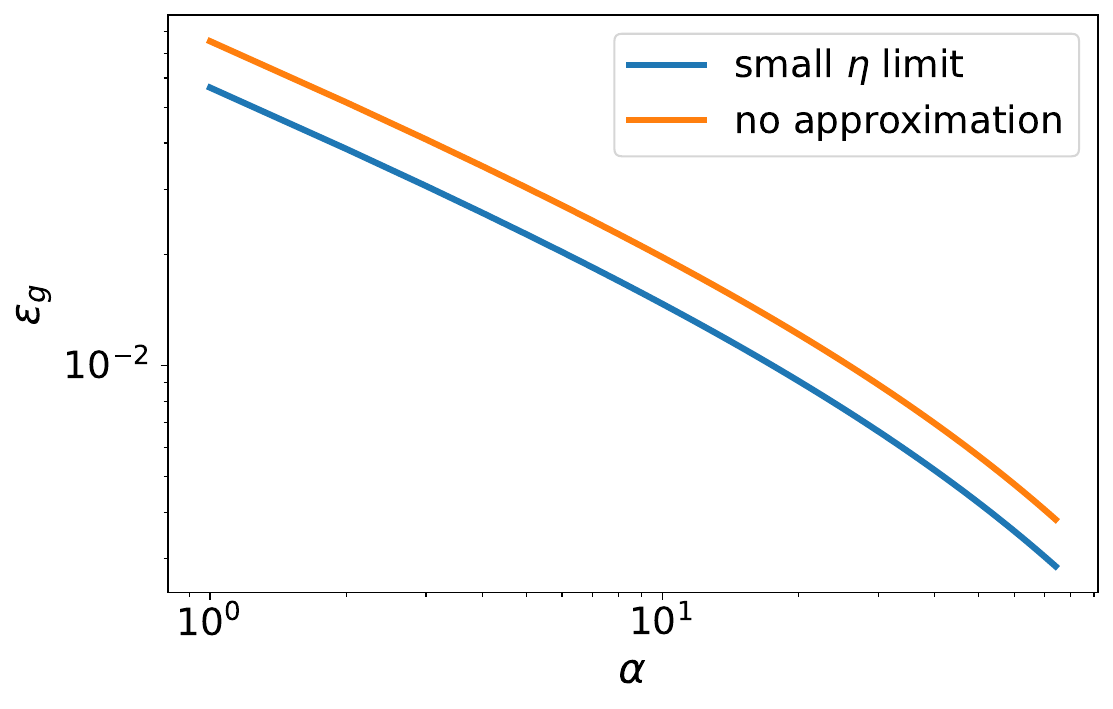}
  \includegraphics[width=0.49\linewidth]{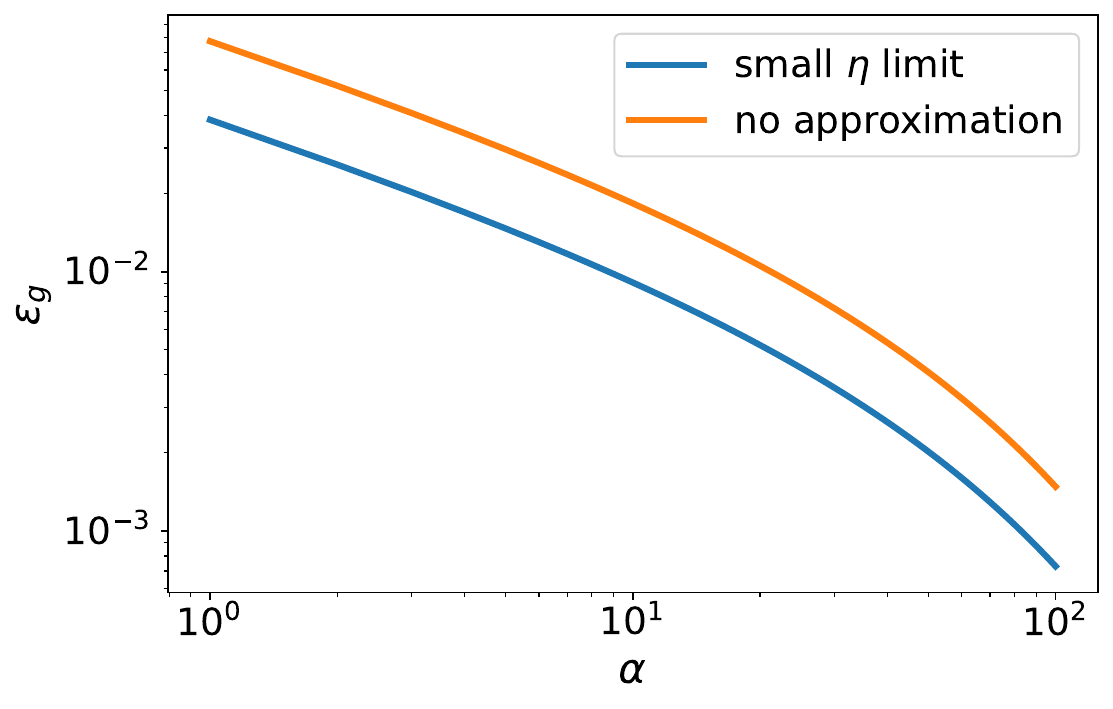}
  \caption{Generalization error evaluated by Eq. (\ref{eps perceptron time}) (blue) and Eq. (\ref{eps perceptron time full}) (orange) for $N=128$ , $K=M=1$, $\sigma_J^2=0$ and $\beta=1$. Upper left: $\eta=0.1$.  Upper Right: $\eta=0.5$. Bottom: $\eta=1.0$. }
  \label{fig: eg0 vs eg 2}
\end{figure}

Most important is the second entry of the eigenvector matrix $\bm{V}_4$ since this will have an influence on the generalization error. Next, we evaluate $\bm{V}_4^{-1}\bm{T}$

\begin{align}
\langle \epsilon_g \rangle_{J_{a,0}, B_a} =  \left(1+\sigma_J^2\right) \frac{1}{2 L} \sum_{k=1}^L b_k \lambda_{B,k} \exp(-2\eta \lambda_{B,k} \alpha) ,
\label{eps perceptron time 3}
\end{align}
with $b_k=\sum _l^L \left(V_B^{-1}\right)_{kl} T^{(l)}$ and $\lambda_{B,k} $ are the eigenvalues of $\bm{B}$ for $a=2$ and $\epsilon=\eta$. Thus, in order to calculate the generalization error, we have to find an expression for $\bm{V}_B^{-1}$. We know that $\bm{B}$ shows some properties of a companion matrix. Therefore, we perform a similarity transformation $\bm{B}= \bm{S} \bm{B}_2 \bm{S}^{-1}$. Thereby, $\bm{B}_2$ has again a companion matrix structure similar to $\bm{A}_1$, but it has $c_{B,i}$ for its last row entry for $i=0,...,L-1$ given by Eq. (\ref{coeff poly B}). For the transformation matrix $\bm{S}$, we obtain
\[
S = \frac{1}{ (a-\eta)}\begin{bmatrix}
a-\eta & 0 & 0 & 0 & \cdots & 0 \\
0 & -1 & 0 & 0 & \cdots & 0 \\
0 & -T^2 \frac{\eta}{a} & \frac{1}{a} & 0 & \cdots & 0 \\
0 & -T^3 \frac{\eta}{a} & T^2 \frac{\eta}{a^2} & -\frac{1}{a^2} & \ddots & 0 \\
\vdots & \vdots & \vdots & \vdots & \ddots & \vdots \\
0 & -T^{L-1} \frac{\eta}{a} & T^{L-2} \frac{\eta}{a^2} & -T^{L-3} \frac{\eta}{a^3} & \cdots & (-1)^{L-1} \frac{1}{a^{L-1}}
\end{bmatrix}
\]
having a triangular structure. The entries of the inverse of a triangular matrix can be calculated by this simple relation
\[
S^{-1}_{ii} = \frac{1}{S_{ii}}
\]
\[
S^{-1}_{ij} = -\frac{1}{S_{ii}} \sum_{k=i}^{j-1} S_{ik} S^{-1}_{kj}, \quad \text{for } i > j
\]
Note that the eigenvector matrix $\bm{V}_{B2}$ is again a Vandermonde matrix, meaning that we know the entries and their inverse. Therefore, we obtain 
\begin{align}
b_k= \sum _l^L \left(\left(\bm{S} \bm{V}_{B2}\right)^{-1}\right)_{kl} T^{(l)}.
\label{b_k}
\end{align}
In practice, we no longer have to directly invert any matrix, which allows us to consider higher values of $L$ and more different covariance matrix power-law exponents $\beta$. Calculating the inverse of $\bm{V}_B$ is numerically very unstable and just possible for small numbers of distinct eigenvalues $L$. \\
Figure \ref{fig: mult 2} compares $b_k$ multiplied with $\lambda_{B,k}$ with the eigenvalues of the data covariance matrix $\lambda_k$ for various $L$. For small learning rates, the differences in the spectra are very small and increase with increasing learning rates until the trend of power-law decay is destroyed. Note that $b_k=\frac{1}{L}$ and $\lambda_{B,k} = -a\lambda_k$ if we just consider the differential equations up to order $\mathcal{O}(\eta)$. Figure \ref{fig: mult 1} shows the same behavior but for different $\beta$. The smaller $\beta$ becomes, the larger the divination from a consistent power-law scaling. Figure \ref{fig: Julia linear eigenvalues 1} shows the spectrum of $\lambda_{B,k}$ and its difference from $\lambda_{k}$. We also observe in the spectra that the pure power-law behavior is perturbed. However, this effect seems to just increase the generalization error without changing its scaling, as shown in Figure \ref{fig: eg0 vs eg 2}. 

\subsection{Feature scaling}
\label{Feature scaling appendix}
Instead of considering the statistical mechanics approach, we study directly the stochastic gradient descent here
\begin{align}
\bm{J}^{\mu+1}-\bm{J}^{\mu}=-\eta \nabla_{\bm{J}} \epsilon \left(\bm{J}^\mu,\bm{\xi}^\mu\right).
\label{difference eq appendix}
\end{align}
From this difference equation, one can derive a Langevin equation for the weight dynamics in the thermodynamic limit $N \to \infty$ (see \cite{Rotskoff2022, PhysRevResearch.6.L022049}). For the continuous equation describing the trajectories of the weights, we obtain
\begin{equation}
\frac{\mathrm{d}\bm{J}}{\mathrm{d}\alpha}=-\eta\nabla_{\bm{J}} \epsilon_g + \frac{\eta}{\sqrt{N}} \bm{\gamma}
\label{Langevin}
\end{equation}
where $\bm{\gamma}$ is a random vector describing the path fluctuations with $\left<\bm{\gamma}\right>=0$ and $\left<\gamma_i\left(\alpha\right) \gamma_j\left(\alpha^\prime\right)\right>=C_{ij} \delta\left(\alpha-\alpha^\prime\right)$. \\
For small learning rates or a proper scaling of the learning rate with the network size, we can neglect path fluctuations and approximate the stochastic process by its mean path. This leads to the following differential equation
\begin{align}
 \frac{\mathrm{d}\bm{J}}{\mathrm{d}\alpha} &\underset{\eta \to 0}{\approx}  -\eta \nabla_{\bm{J}} \epsilon_g \nonumber \\
 & = -\eta \bm{\Sigma} \left(  \bm{J} -  \bm{B}\right).
\label{mean path}
\end{align}
For a diagonal a covariance matrix $\Sigma_{kl}=\delta_{kl} \lambda_{kl}$, the solution of Equation (\ref{mean path}) is
\begin{align}
J_{k} = \exp\left(-\eta \lambda_{k} \alpha \right) \left(J_k^0-B_k\right) + B_k
\label{sol perceptron diag}
\end{align}
for $k \in \{1,...,L\}$ and initial weight component $J_k^0$. Thus, the individual weights are learned exponentially fast, and each component $J_{k} $ converges thereby to the component of the teacher. For the first-order order parameters, we find
\begin{align}
R^{(1)} = \sum_{k=1}^N \frac{J_k \lambda_k B_k}{N} = \frac{1}{N} \sum_{k=1}^N \left( \exp(-\eta \lambda_k \alpha) \left(J_{k}^0 - B_k\right)+ B_k \right)  \lambda_k B_k \\
Q^{(1)} = \sum_{k=1}^N \frac{J_k \lambda_k J_k}{N} = \frac{1}{N} \sum_{k=1}^N \left( \exp(-\eta \lambda_k \alpha) \left(J_{k}^0 - B_k\right)+ B_k \right)^2  \lambda_k 
\end{align}
and for their corresponding expectation value
\begin{align}
\langle R^{(1)} \rangle_{J_{k}^0, B_k} &= 1 - \frac{1}{N} \sum_{k=1}^N \lambda_k \exp(-\eta \lambda_k \alpha) \\
\langle Q^{(1)} \rangle_{J_{k}^0, B_k} &= 1 + \frac{1+\sigma_J^2}{N} \sum_{k=1}^N \lambda_k \exp(-2\eta \lambda_k \alpha) - \frac{2}{N} \sum_{k=1}^N \lambda_k \exp(-\eta \lambda_k \alpha)
\label{expect QR perceptron}
\end{align}
For the expectation value of the generalization error, we obtain
\begin{align}
\langle \epsilon_g \rangle_{J_{k}^0, B_k} &=  \frac{1+\sigma_J^2}{2N} \sum_{k=1}^N \lambda_k \exp(-2\eta \lambda_k \alpha) \\
&= \frac{1+\sigma_J^2}{2L}\sum_{k=1}^L \lambda_k \exp(-2\eta \lambda_k \alpha)
\label{eps perceptron time 2 method appendix}
\end{align}
where we have exploit that $\frac{N}{L} \in \mathbb{N}$ as assumed for our setup (see Section \ref{Setup section}). Note that Equation (\ref{eps perceptron time 2 method appendix}) that we derived from the approximated Langevin equation, is the same as the generalization error derived from the statistical mechanics approach in the small learning rate limit $\eta \to 0$ given by Equation (\ref{eps perceptron time}). Therefore, neglecting the higher order of the learning rate in the differential equations is equivalent to neglecting path fluctuations of the stochastic gradient descent.\\
Next, in order to model how the generalization error scales as more and more feature directions are explored, we assume that $N_l$ components of the student are already learned. Thereby, the other $N-N_l$ components are kept fixed and random. Here, we want to investigate the generalization error as a function of $N_l$. For this, we make the ansatz:
\begin{align}
J_k = \begin{cases} B_k & \text{for } k = 1, \dots, N_l \\ J_{k}^0& \text{else} \end{cases}
\label{cases appendix}
\end{align}
Therefore, we obtain two different parts for the order parameters
\begin{align}
R^{(1)} = \frac{1}{N} \sum_{k=1}^N \lambda_k J_k B_k = \frac{1}{N} \left(\sum_{k=1}^{N_l} \lambda_k B_k^2 + \sum_{k=N_l+1}^N \lambda_k J_{k}^0 B_k\right) \nonumber \\
Q^{(1)} = \frac{1}{N} \sum_{k=1}^N \lambda_k J_k J_k = \frac{1}{N} \left(\sum_{k=1}^{N_l} \lambda_k B_k^2 + \sum_{k=N_l+1}^N \lambda_k J_{k}^0 J_{k}^0\right)
\end{align}
and their expectation values become
\begin{align}
\langle R^{(1)} \rangle_{J_{k,0}, B_k} &= \frac{1}{N} \sum_{a=1}^{N_l} \lambda_a, \nonumber \\ 
\langle Q^{(1)} \rangle_{J_{k,0}, B_k}& = \frac{1}{N} \sum_{k=1}^{N_l} \lambda_k + \frac{\sigma_J^2}{N} \sum_{k=N_l+1}^N \lambda_k 
\label{feature scaling expect first order}
\end{align}
Next, we insert Equation (\ref{feature scaling expect first order}) into the expression of the generalization error and obtain
\begin{align}
\langle \epsilon_g \rangle_{J_{k,0}, B_k}  &= \frac{1+\sigma_J^2}{2}\left( 1 - \frac{1}{L} \sum_{k=1}^{N_l} \lambda_k \right) \nonumber \\
&=\frac{1+\sigma_J^2}{2}\frac{1}{L} \sum_{k=N_l}^{L} \lambda_k,
\label{eg params}
\end{align}
where we have exploit that $\sum_k^N \lambda_k=N$ and $\frac{N}{L} \in \mathbb{N}$. Next, we use the definition of the eigenvalues $\lambda_i = \lambda_+ \left(\frac{1}{i}\right)^{1+\beta}$ and define the partial sum  $S(N_l) = \frac{\lambda_+}{L} \sum_{k=N_l}^{L}  \left(\frac{1}{i}\right)^{1+\beta}$ which leads to
\begin{align}
\epsilon_g = \frac{1+\sigma_J^2}{2} S(N_l) 
\end{align}
In order to approximate the sum, we use the Euler-Maclaurin formula defined in Equation (\ref{Euler Mclaurin}) and replace the sum with an integral. For $\beta > 0$, we find
\begin{align}
S(N_l) &\approx  \frac{\lambda_+}{L} \int_{N_l}^{L} \frac{1}{i^{\beta+1}} \, di \nonumber \\
&=  \frac{\lambda_+}{N \beta} \left(\frac{1}{N_l^\beta} - \frac{1}{L^\beta}\right)
\end{align}
Thus, we obtain
\begin{align}
\langle \epsilon_{g,\mathrm{asymp}} \rangle \approx \frac{1+\sigma_J^2}{2} \frac{\lambda_+}{L \beta} \left(\frac{1}{N_l^\beta} - \frac{1}{L^\beta}\right).
\label{asymp eigenbasis appendix}
\end{align}
The zeroth-order error of this approximation, which we consider as the worst-case, is $\delta \epsilon_g = \frac{1+\sigma_J^2}{4}\frac{\lambda_+}{L} \left( \frac{1}{N_l^{1+\beta}}+\frac{1}{L^{1+\beta}}\right)$. \\
If we do not assume that $N_l$ eigenvalues are already converged, then we can make the ansatz
\begin{align}
\tilde{J}_k = \begin{cases} J_k & \text{for } k = 1, \dots, N_l \\ J_{k}^0& \text{else} \end{cases}
\label{cases appendix 2}
\end{align}
where $J_k$ are given in Equation (\ref{sol perceptron diag}). Therefore, we again obtain two different parts for the order parameters
\begin{align}
R^{(1)} &= \frac{1}{N} \sum_{k=1}^N \lambda_k \tilde{J}_k B_k \nonumber \\
&= \frac{1}{N} \left(\sum_{k=1}^{N_l} \lambda_k J_k + \sum_{k=N_l+1}^N \lambda_k J_{k}^0 B_k\right) \nonumber \\
&=\frac{1}{N} \sum_{k=1}^N \lambda_k \left[ \exp\left(-\eta \lambda_{k} \alpha \right) \left(J_k^0-B_k\right) + B_k\right]B_k  + \frac{1}{N} \sum_{k=N_l+1}^N \lambda_k J_{k}^0 B_k \nonumber\\
Q^{(1)} &= \frac{1}{N} \sum_{k=1}^N \lambda_k \tilde{J}_k\tilde{J}_k \nonumber \\
&= \frac{1}{N} \left(\sum_{k=1}^{N_l} \lambda_k J_k^2 + \sum_{k=N_l+1}^N \lambda_k J_{k}^0 J_{k}^0\right) \nonumber \\
&= \frac{1}{N} \left(\sum_{k=1}^{N_l} \lambda_k \left[ \exp\left(-\eta \lambda_{k} \alpha \right) \left(J_k^0-B_k\right) + B_k\right]^2 + \sum_{k=N_l+1}^N \lambda_k J_{k}^0 J_{k}^0\right) 
\end{align}
and their expectation values become
\begin{align}
\langle R^{(1)} \rangle_{J_{k}^0, B_k} = &\frac{1}{N} \sum_{k=1}^{N_l}  \lambda_k -\frac{1}{N} \sum_{k=1}^{N_l} \lambda_k \exp\left(-\eta \lambda_{k} \alpha \right), \nonumber \\ 
\langle Q^{(1)} \rangle_{J_{k}^0, B_k} =& \frac{1+\sigma_J^2}{N} \sum_{k=1}^{N_l} \lambda_k \exp\left(-2\eta \lambda_{k} \alpha \right) -\frac{2}{N} \sum_{k=1}^{N_l} \lambda_k \exp\left(-\eta \lambda_{k} \alpha \right)  \nonumber \\
&+ \frac{1}{N} \sum_{k=1}^{N_l} \lambda_k + \frac{\sigma_J^2}{N} \sum_{k=N_l+1}^N \lambda_k 
\label{feature scaling expect first order 2}
\end{align}
Note that Equation (\ref{feature scaling expect first order 2} reduces to Equation (\ref{feature scaling expect first order}) for $\alpha \to \infty$. Next, we insert Equation (\ref{feature scaling expect first order 2}) into the expression of the generalization error and obtain
\begin{align}
\langle \epsilon_g \rangle_{J_{k}^0, B_k} =\frac{1+\sigma_J^2}{2L}\left[\sum_{k=1}^{N_l} \lambda_k \exp\left(-2\eta \lambda_{k} \alpha \right) +\sum_{k=N_l}^{L} \lambda_k \right]
\label{eg final form params 2}
\end{align}
Note that Equation (\ref{eg final form params 2}) is basically a combination of Equation (\ref{eg params}) and Equation (\ref{eps perceptron time 2 method appendix}). If we set $N_l=N$ in Equation (\ref{eg final form params 2}), then we can reproduce Equation (\ref{eps perceptron time 2 method appendix}), and if we let $\alpha \to \infty$ in Equation (\ref{eg final form params 2}), then we reach the asymptotic generalization error given by Equation (\ref{eg params}). Therefore, we can repeat the analysis of this and the previous subsection separately and finally obtain the generalization error for small learning rates
\begin{align}
\epsilon_{g}\left(N_l,\alpha\right) \underset{\eta \to 0}{\approx} & \frac{1+\sigma^2}{2} \frac{\lambda_+}{L} \Biggl[ \frac{1}{\beta}\left(\frac{1}{N_l^\beta}-\frac{1}{L^\beta}\right) \nonumber \\
&+\frac{\left(2\eta\lambda_+ \alpha\right)^{-\frac{\beta}{1+\beta}}}{1+\beta} \Biggl[ \Gamma \left(\frac{\beta}{1+\beta}, \frac{2\eta\lambda_+ \alpha}{L^{\beta+1}}\right) -\Gamma \left(\frac{\beta}{1+\beta}, 2\eta\lambda_+ \alpha\right)\Biggr]\Biggr].
\label{ful eg}
\end{align}
\\
\begin{figure}[t]
  \centering
  \includegraphics[width=0.49\linewidth]{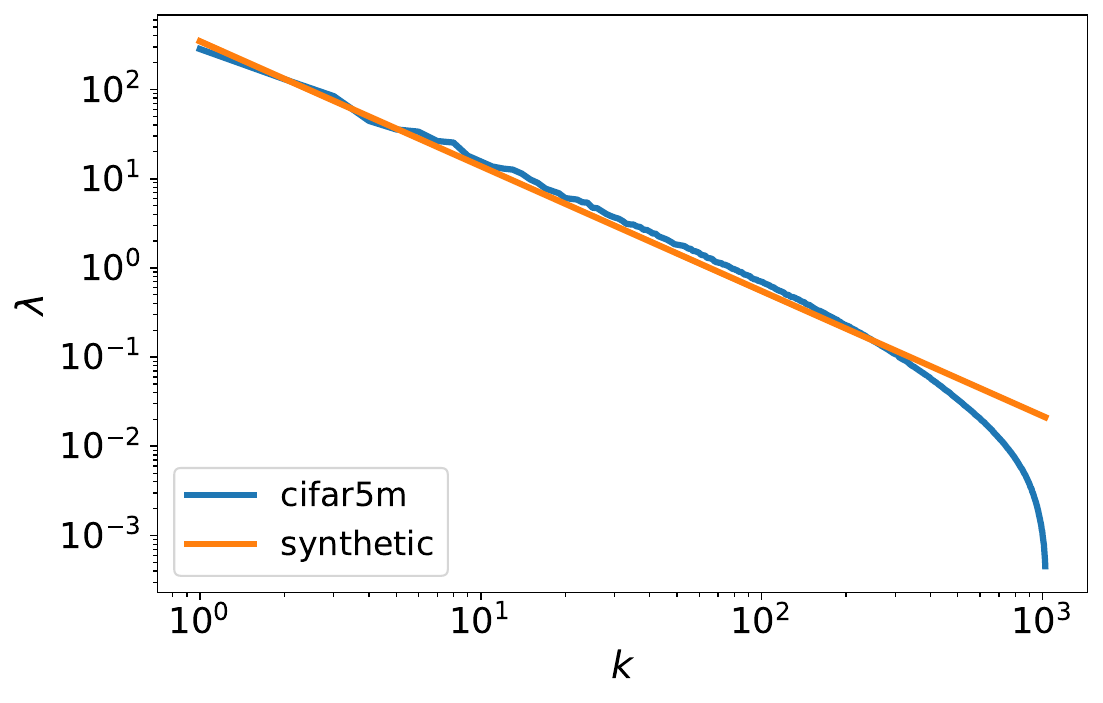}
  \caption{Left: Spectra of the feature-feature covariance matrix of CIFAR-5m data set vs. Eq. (\ref{data generating model}) for $\beta=0.3$ and $N=L=1024$.}
  \label{fig: cifar-5m}
\end{figure}
Next, we consider a general $\bm{\Sigma}$, which is no longer diagonal. The solution of the Langevin equation is
\begin{align}
\tilde{\mathbf{J}} = \exp(-\eta \mathbf{D} \alpha) (\tilde{\mathbf{J}}^{(0)} - \tilde{\mathbf{B}}) + \tilde{\mathbf{B}},
\label{parametrize student vector}
\end{align}
where we have introduced the eigendecomposition $\bm{\Sigma} = \mathbf{W} \mathbf{D} \mathbf{W}^\top$ and expressed the weights in the eigenbasis $  \tilde{\mathbf{J}} = \mathbf{W}^\top \mathbf{J},  \tilde{\mathbf{B}} = \mathbf{W}^\top \mathbf{B}$. Therefore, we obtain for the order parameters
\begin{align}
Q^{(1)} = \frac{\mathbf{J}^\top \bm{\Sigma} \mathbf{J}}{N} = \frac{\mathbf{J}^\top \mathbf{W} \mathbf{D} \mathbf{W}^\top \mathbf{J}}{N} = \frac{\tilde{\mathbf{J}}^\top \mathbf{D} \tilde{\mathbf{J}}}{N} \nonumber \\
R^{(1)} = \frac{\mathbf{J}^\top \bm{\Sigma} \mathbf{B}}{N} = \frac{\mathbf{J}^\top \mathbf{W} \mathbf{D} \mathbf{W}^\top \mathbf{B}}{N} = \frac{\tilde{\mathbf{J}}^\top \mathbf{D} \tilde{\mathbf{B}}}{N}
\end{align}
Note that the expectation values for the order parameters and, therefore, for the generalization error are still the same. \\
We use the parametrization from Eq. (\ref{parametrize student vector}) to obtain the results shown in the right panel of Figure \ref{fig: eg lin Nl}. Thereby, we test our prediction for the generalization error from Eq. (\ref{ful eg}) to a student network trained on CIFAR-5M images using approximately $10^6$ (see \cite{nakkiran2021the} for details on the dataset). We use only the first channel of the images resulting in a total input dimension of $N=1024$ after flattening. To approximate the true covariance matrix $\bm{\Sigma}$, we numerically estimate the feature-feature covariance matrix based on the input examples from the training dataset. During training, we update only the first entries of $\tilde{\bm{J}}$, while resetting the remaining entries to their initial values after each iteration. Based on the spectra of the feature-feature covariance matrix depicted in Figure \ref{fig: cifar-5m}, we estimate $\beta \approx 0.3$ and use this spectrum to evaluate Eq. (\ref{parametrize student vector}).

\subsubsection{Student training beyond covariance eigenbasis}
In this Subsection, we consider the generalization evaluated for a student with $N_l$ trainable weights and $N-N_l$ random weights. Thereby, we no longer train the weights of the student in the eigenbasis of the data-covariance matrix. In order to distinguish between trainable and untrainable components, we decompose the student vector \(\bm{J}\) in two parts: the trainable components \(\tilde{\bm{J}}\) and the non-trainable components \(\hat{\bm{J}}\). For the covariance matrix \(\bm{\Sigma}\), we identify the following structure in block matrix form

\begin{align}
  \bm{\Sigma} = \begin{pmatrix} \tilde{\bm{\Sigma}} & \bm{\Sigma}_{\text{cross}} \\ \bm{\Sigma}_{\text{cross}}^\top & \hat{\bm{\Sigma}} \end{pmatrix},  
\end{align}
where
\begin{itemize}
    \item \(\tilde{\bm{\Sigma}} \in \mathbb{R}^{N_l \times N_l}\) represents the submatrix of the covariance matrix, acting exclusively on the subspace spanned by the first \(N_l\) components of the vector \(\bm{J}\), denoted as \(\tilde{\bm{J}}\). 
    
    \item \(\hat{\bm{\Sigma}} \in \mathbb{R}^{(N-N_l) \times (N-N_l)}\) refers to the submatrix corresponding to the subspace of the remaining \(N-N_l\) components of \(\bm{J}\), denoted as \(\hat{\bm{J}}\). 
    
    \item \(\bm{\Sigma}_{\text{cross}} \in \mathbb{R}^{N_l \times (N-N_l)}\) represents the cross-covariance matrix part, describing the interactions between the subspace of the first \(N_l\) components and the complementary subspace of the last \(N-N_l\) components
\end{itemize}

The evolution of the student vector \(\bm{J}\) is governed by the differential equation for small learning rates

\begin{align}
\frac{d \bm{J}_i}{d\alpha} \underset{\eta \to 0}{\approx}  \eta \sum_{k=1}^N \Sigma_{ik} (B_k - J_k).
\end{align}

The solution for the trainable components \(\tilde{\bm{J}}(\alpha)\) at time \( \alpha \) is
given by

\begin{align}
\tilde{\bm{J}}(\alpha) &= e^{-\eta \tilde{\bm{\Sigma}} \alpha} \tilde{\bm{J}}^0 + \tilde{\bm{B}} \left( 1 - e^{-\eta \tilde{\bm{\Sigma}} \alpha} \right)  + \tilde{\bm{\Sigma}}^{-1} \bm{\Sigma}_{\text{cross}} \left( \hat{\bm{B}} - \hat{\bm{J}} \right) \left( 1 - e^{-\eta \tilde{\bm{\Sigma}} \alpha} \right),
\end{align}

where \(\tilde{\bm{J}}^0\) is the initial condition for the trainable components, and the non-trainable components are kept at their random initial values. Moreover, \(\tilde{\bm{B}}\) and \(\hat{\bm{B}}\) are the components of the teacher vector equivalent to the trainable and non-trainable parts of the student vector, respectively. At stationarity (\( \alpha \to \infty \)), the trainable components become
\begin{align}
\tilde{\bm{J}} = \tilde{\bm{B}} + \tilde{\bm{\Sigma}}^{-1} \bm{\Sigma}_{\text{cross}} \left( \hat{\bm{B}} - \hat{\bm{J}} \right).
\end{align}
For the first-order order parameters, we find
\begin{figure}[t]
  \centering
  \includegraphics[width=0.49\linewidth]{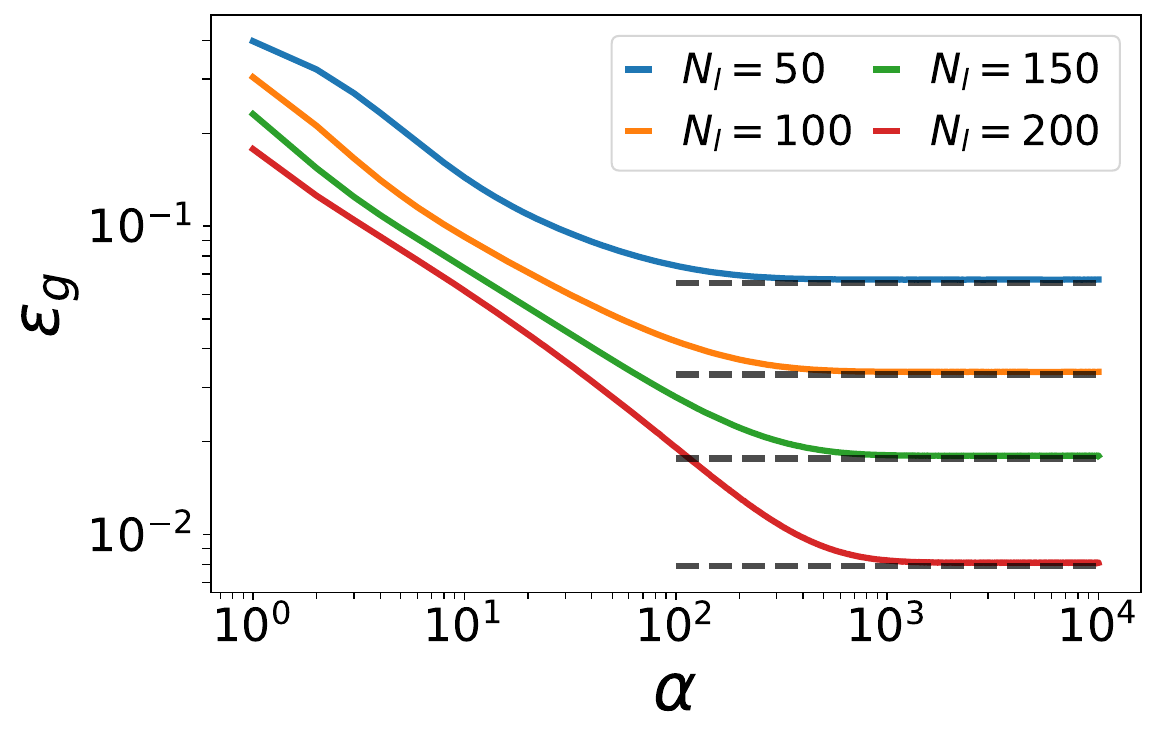}
  \caption{$\epsilon_g$ as a function of $\alpha$ for different trainable input dimensions $N_l$ of the student vector with $L = N = 256$, $K = M = 1$, $\sigma_J = 0.01$, $\eta = 0.05$, and $\beta = 1$. Here, we train the student outside the eigenbasis of the data covariance matrix. We compare results based on simulations (solid curves) to the theoretical prediction from Eq. (\ref{eg random Nl}) (black dashed lines). The student network is trained on synthetic data, where we average over $300$ different initializations of student and teacher vectors.}
  \label{fig: eg random Nl}
\end{figure} 

\begin{align}
R^{(1)} &=\frac{1}{N} \bm{J}^\top \bm{\Sigma} \bm{B} \nonumber \\
&= \frac{1}{N} \left( \tilde{\bm{J}}^\top \tilde{\bm{\Sigma}} \tilde{\bm{B}} + \tilde{\bm{J}}^\top \bm{\Sigma}_{\text{cross}} \hat{\bm{B}} + \hat{\bm{J}}^\top \bm{\Sigma}_{\text{cross}}^\top \tilde{\bm{B}} + \hat{\bm{J}}^\top \hat{\bm{\Sigma}} \hat{\bm{B}} \right) \nonumber \\
Q^{(1)} &=\frac{1}{N} \bm{J}^\top \bm{\Sigma} \bm{J} \nonumber \\
 &= \frac{1}{N} \left( \tilde{\bm{J}}^\top \tilde{\bm{\Sigma}} \tilde{\bm{J}} + 2 \tilde{\bm{J}}^\top \bm{\Sigma}_{\text{cross}} \hat{\bm{J}} + \hat{\bm{J}}^\top \hat{\bm{\Sigma}} \hat{\bm{J}} \right).
\end{align}
Taking the expectation over \(\bm{B}\) and \(\hat{\bm{J}}\) yields

\begin{align}
\langle R^{(1)}\rangle_{B_i,\hat{J}_i} &= \frac{1}{N} \left( \text{Tr}(\tilde{\bm{\Sigma}}) + \text{Tr}\left( \bm{\Sigma}_{\text{cross}}^\top \tilde{\bm{\Sigma}}^{-1} \bm{\Sigma}_{\text{cross}} \right) \right) \nonumber \\
\langle Q^{(1)}\rangle_{B_i,\hat{J}_i} &= \frac{1}{N} \left( \text{Tr}(\tilde{\bm{\Sigma}}) + (1 + \sigma_J^2) \text{Tr}\left( \bm{\Sigma}_{\text{cross}}^\top \tilde{\bm{\Sigma}}^{-1} \bm{\Sigma}_{\text{cross}} \right) + \sigma_J^2 \text{Tr}(\hat{\bm{\Sigma}}) - 2 \sigma_J^2 \text{Tr}\left( \bm{\Sigma}_{\text{cross}}^\top \tilde{\bm{\Sigma}}^{-1} \bm{\Sigma}_{\text{cross}} \right) \right).
\end{align}
Finally, we obtain for the generalization error
\begin{align}
\langle \epsilon_{g,\mathrm{asymp}} \rangle_{B_i,\hat{J}_i} &=  \frac{1}{2} \left[ 1 - \frac{1}{N} \text{Tr}(\tilde{\bm{\Sigma}}) + \frac{1}{N} \sigma_J^2 \text{Tr}(\hat{\bm{\Sigma}}) - \frac{1+\sigma_J^2}{N} \text{Tr}\left( \bm{\Sigma}_{\text{cross}}^\top \tilde{\bm{\Sigma}}^{-1} \bm{\Sigma}_{\text{cross}} \right) \right].
\label{eg random Nl}
\end{align}

\begin{figure}[t]
  \centering
  \includegraphics[width=0.49\linewidth]{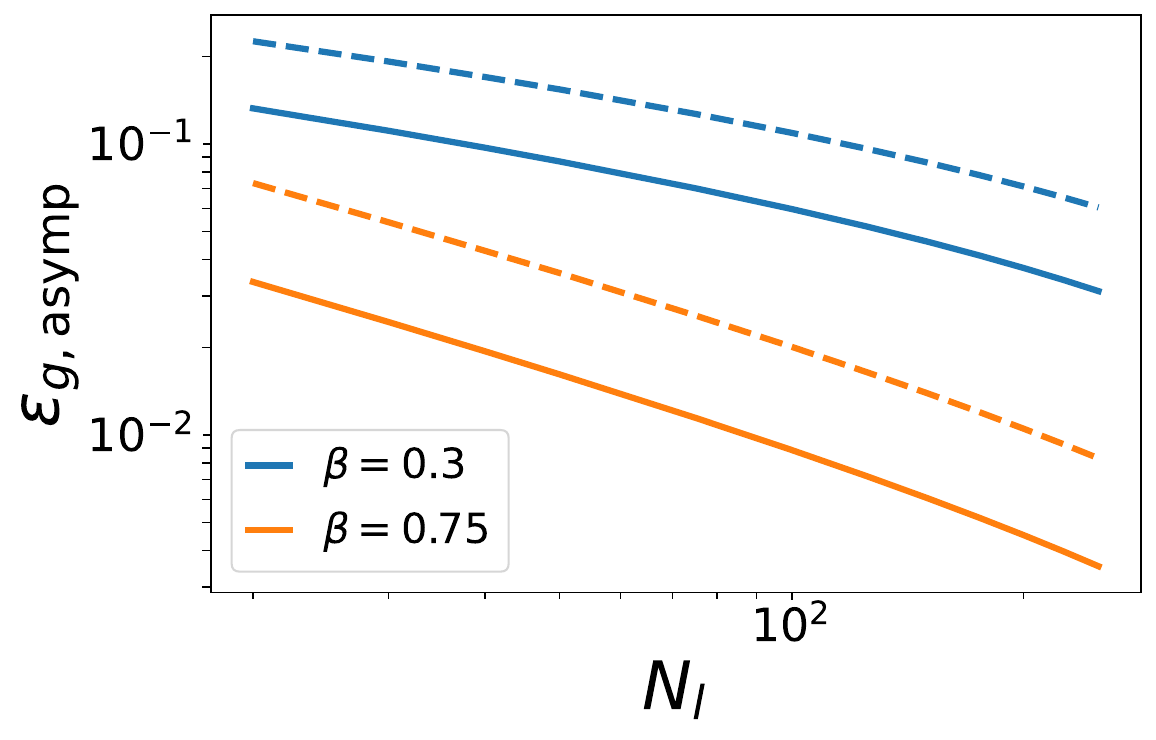}
  \includegraphics[width=0.49\linewidth]{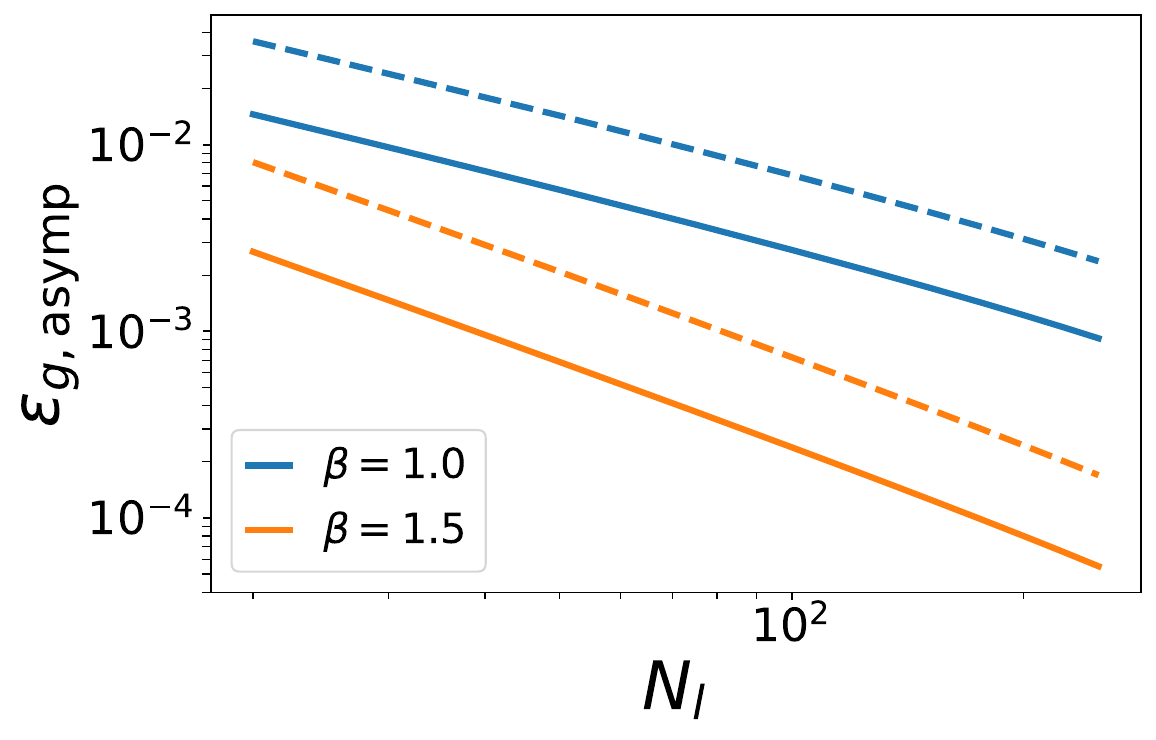}
  \caption{$\epsilon_{g,\mathrm{asymp}}$ as a function of $N_l$ for different $\beta$ and $N=L=1024$. The student is trained with synthetic data. We compare the numerical solution of Eq. (\ref{eg random Nl}) (dashed) with our theoretical prediction for training the student vector in the eigenbasis of the data covariance matrix, as given by Eq. (\ref{asymp eigenbasis appendix}) (solid). The numerical solution is averaged over $100$ different covariance matrices. }
  \label{fig: asymp random vs eigenbasis}
\end{figure}

Note that for diagonal covariance matrices $\Sigma_{ij}= \delta_{ij} \lambda_i$, Eq. (\ref{eg random Nl}) reduces to Eq. (\ref{eg final form params}) for $\alpha \to \infty$. Figure \ref{fig: eg random Nl} presents the generalization error obtained from simulations for various $N_l$ as a function of $\alpha$. In this comparison, the asymptotic solution derived from Eq. (\ref{eg random Nl}) aligns closely with the simulation results, demonstrating excellent agreement.\\

Next, we compare the numerical solution of Eq. (\ref{eg random Nl}) with our theoretical prediction for training the student vector in the eigenbasis of the data covariance matrix, as given by Eq. (\ref{asymp eigenbasis appendix}). The results are presented in Figure \ref{fig: asymp random vs eigenbasis}. For a fixed number of trainable parameters, $N_l$, the generalization error is consistently lower when the student is trained in the eigenbasis of the data covariance matrix. This is expected, as training in the eigenbasis aligns with the directions of the largest $N_l$ eigenvalues, leading to a more efficient learning process. Consequently, the overall generalization error is reduced compared to training outside the eigenbasis. Under the condition $N^\beta \gg N_l^\beta$, we observe the same power-law scaling for the generalization error, $\epsilon_{g,\mathrm{asymp}} \propto N_l^{-\beta}$, in both scenarios. Additionally, for more general configurations, we find that the scaling behavior of the generalization error remains consistent across both setups. The data covariance matrices are generated by $\bm{\Sigma}= \bm{W} \bm{\Lambda} \bm{W}^\top$, where ${\Lambda}_{ij} = \delta_{ij} \lambda_i $ with eigenvalues defined by Eq. (\ref{data generating model}) and $\bm{W} $ is a random orthogonal matrix with zero mean and variance which scales as $\langle W_{ij}^2 \rangle \sim \frac{1}{N}$.

\section{Non-linear activation}
\subsection{Differential equations}
Throughout this work, we consider the error function as our non-linear activation function $g\left(x\right)= \mathrm{erf}\left({\frac{x}{\sqrt{2}}}\right)$. For this activation function, one can solve the integrals $I_2, I_3$ and $I_4$ given in Section \ref{odes} analytically (see \cite{PhysRevE.52.4225,NEURIPS2019_287e03db}). We find
\begin{align}
I_2(C) &= \frac{2}{\pi} \arcsin \left( \frac{C_{12}}{\sqrt{(1 + C_{11})(1 + C_{22})}} \right), \\
I_3(C) &= \frac{2}{\pi}  \frac{1}{\sqrt{(1 + C_{11})(1 + C_{33}) - C_{13}^2}} \left( \frac{C_{23}(1 + C_{11}) - C_{12}C_{13}}{1 + C_{11}} \right), \\
I_4(C) &= \frac{4}{\pi^2} \frac{1}{\sqrt{A_4}} \arcsin \left( \frac{A_0}{\sqrt{A_1} \sqrt{A_2}} \right)
\end{align}

where
\[
A_4 = (1 + C_{11})(1 + C_{22}) - C_{12}^2,
\]
and
\[
A_0 = A_4 C_{34} + C_{23} C_{24} (1 + C_{11}) + C_{13} C_{14} (1 + C_{22}) + C_{12}^2 C_{13}^2 C_{24}^2,
\]
\[
A_1 = A_4 (1 + C_{33}) + C_{23} (1 + C_{11}) - C_{13} (1 + C_{22}) + 2 C_{12} C_{13} C_{23},
\]
\[
A_2 = A_4 (1 + C_{44}) - C_{24} (1 + C_{11}) - C_{14} (1 + C_{22}) + 2 C_{12} C_{14} C_{24},
\]

depending on the precise covariance matrix and its entries $C_{ij}$ as discussed in Section \ref{odes}. After evaluating all necessary covariance matrices, we obtain for $K=M$

\begin{align}
\frac{dR_{in}^{(l)}}{d\alpha}=&\frac{2\eta}{M\pi}\frac{1}{1+Q_{ii}}\Biggl[\sum_m^M \frac{T_{nm}^{(l+1)}\left(1+Q_{ii}\right)-R_{in}^{(l+1)}R_{im}}{\sqrt{\left(1+Q_{ii}\right)\left(1+T_{mm}\right)-R_{im}^2}} -\sum_j^M \frac{R_{jn}^{(l+1)}\left(1+Q_{ii}\right)-R_{in}^{(l+1)}Q_{ij}}{\sqrt{\left(1+Q_{ii}\right)\left(1+Q_{jj}\right)-Q_{ij}^2}} \Biggr] ,\\
 \frac{dQ_{ik}^{(l)}}{d\alpha}=& \frac{2\eta}{M\pi}\Biggl[\frac{1}{1+Q_{ii}}\left(\sum_m^M \frac{R_{km}^{(l+1)}\left(1+Q_{ii}\right)-Q_{ik}^{(l+1)}R_{im}}{\sqrt{\left(1+Q_{ii}\right)\left(1+T_{mm}\right)-R_{im}^2}} -\sum_j^M \frac{Q_{kj}^{(l+1)}\left(1+Q_{ii}\right)-Q_{ik}^{(l+1)}Q_{ij}}{\sqrt{\left(1+Q_{ii}\right)\left(1+Q_{jj}\right)-Q_{ij}^2}}\right) \nonumber \\
&\hspace{-30pt}+ \frac{1}{1+Q_{kk}}\left(\sum_m^M \frac{R_{im}^{(l+1)}\left(1+Q_{kk}\right)-Q_{ik}^{(l+1)}R_{km}}{\sqrt{\left(1+Q_{kk}\right)\left(1+T_{mm}\right)-R_{km}^2}} -\sum_j^M  \frac{Q_{ij}^{(l+1)}\left(1+Q_{kk}\right)-Q_{ik}^{(l+1)}Q_{kj}}{\sqrt{\left(1+Q_{kk}\right)\left(1+Q_{jj}\right)-Q_{kj}^2}}\right)\Biggr] \nonumber \\
 &+ \mathcal{O}\left(\eta^2\right),
\label{odes non appenix}
\end{align}
where we have omitted higher-order terms in $\eta$ for notational simplicity. Similarly, we have also omitted the superscripts for the first-order parameters, meaning that all instances of $R_{in}$, $T_{nm}$, and $Q_{ij}$ without superscripts correspond to their first-order forms: $R^{(1)}_{in}$, $T^{(1)}_{nm}$, and $Q^{(1)}_{ij}$. Note that the differential equations are not closed given in Equation \ref{odes non appenix}. For the last component $L-1$ one has to apply the Cayley-Hamilton theorem for the order parameters as discussed in Section \ref{Setup section}. In the same notation, the generalization error becomes
\begin{align}
    \epsilon_g = \frac{1}{M\pi} \Biggl[ 
\sum_{i,k} \arcsin \left( \frac{Q_{ik}}{\sqrt{1 + Q_{ii}} \sqrt{1 + Q_{kk}}} \right)&
+ \sum_{n,m} \arcsin \left( \frac{T_{nm}}{\sqrt{1 + T_{nn}} \sqrt{1 + T_{mm}}} \right) \nonumber \\
&- 2 \sum_{i,n} \arcsin \left( \frac{R_{in}}{\sqrt{1 + Q_{ii}} \sqrt{1 + T_{nn}}} \right) 
\Biggr],
\end{align}
which just depends on the first-order order parameters. This system is gonna be analyzed in the following.

\subsection{plateau height}
\label{plateau height appendix}

\begin{figure}
  \centering
  \includegraphics[width=0.49\linewidth]{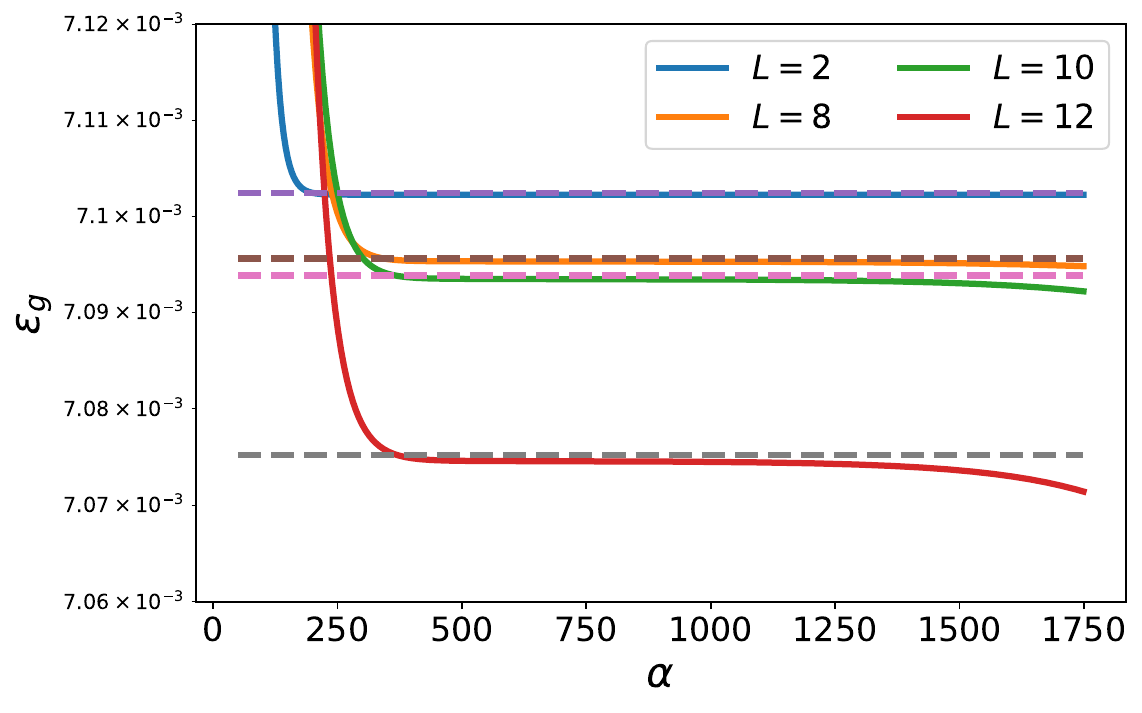}
  \caption{Comparison of the generalization error plateau evaluated by numerical solutions of the differential equations given in Eq. (\ref{odes non appenix}) (solid) and approximated fixed point given by Eq. (\ref{plateau fix approx appendix}) (dashed) for $N=7000$, $K=M=4$, $\sigma_J=0.01$, $\beta=0.25$ and $\eta=0.1$. We solve the differential equations up to $\mathcal{O}\left(\eta\right)$. }
  \label{fig: plateau numerics vs approx}
\end{figure}

\begin{figure}[t]
  \centering
    \includegraphics[width=0.49\linewidth]{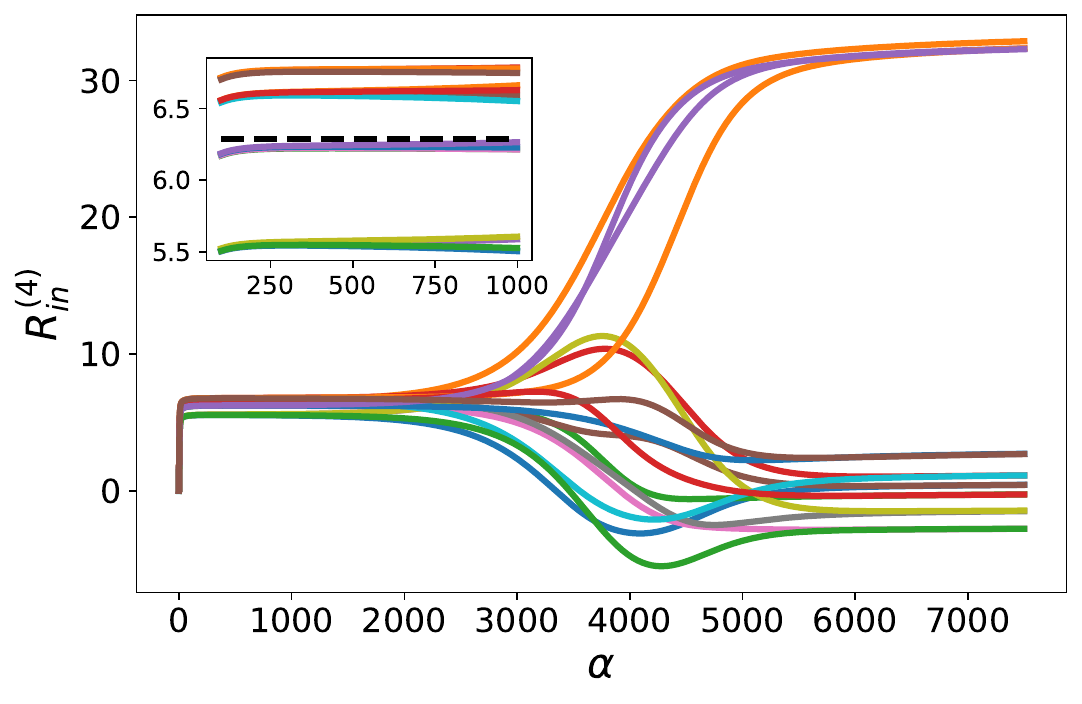}
  \includegraphics[width=0.49\linewidth]{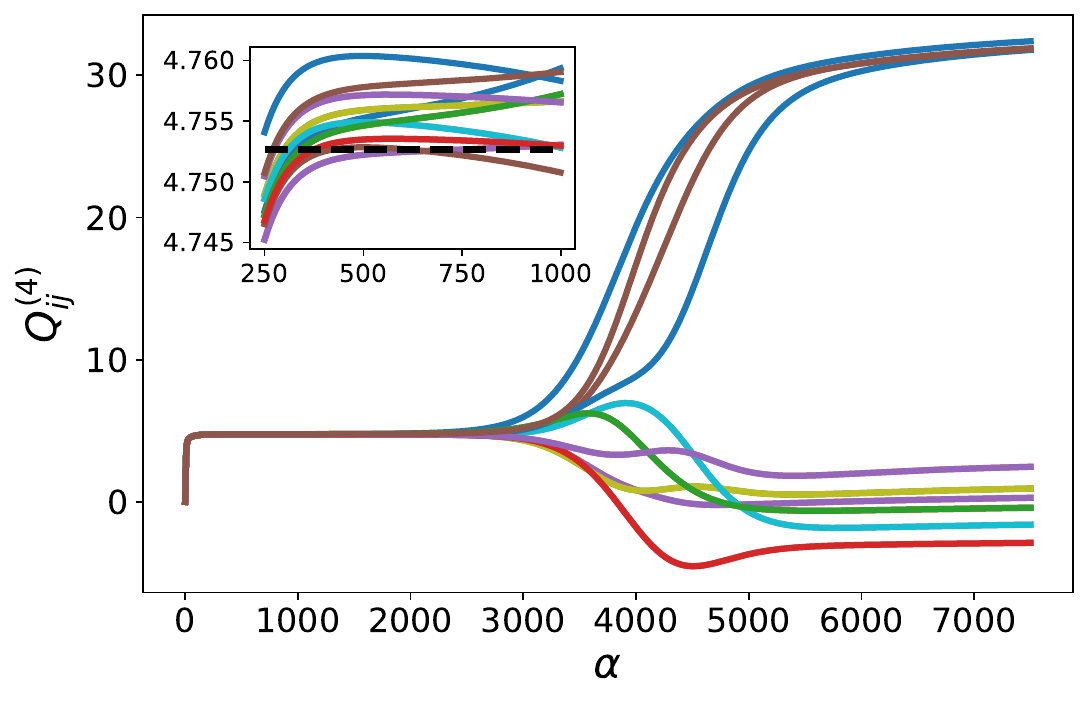} \\
   \includegraphics[width=0.49\linewidth]{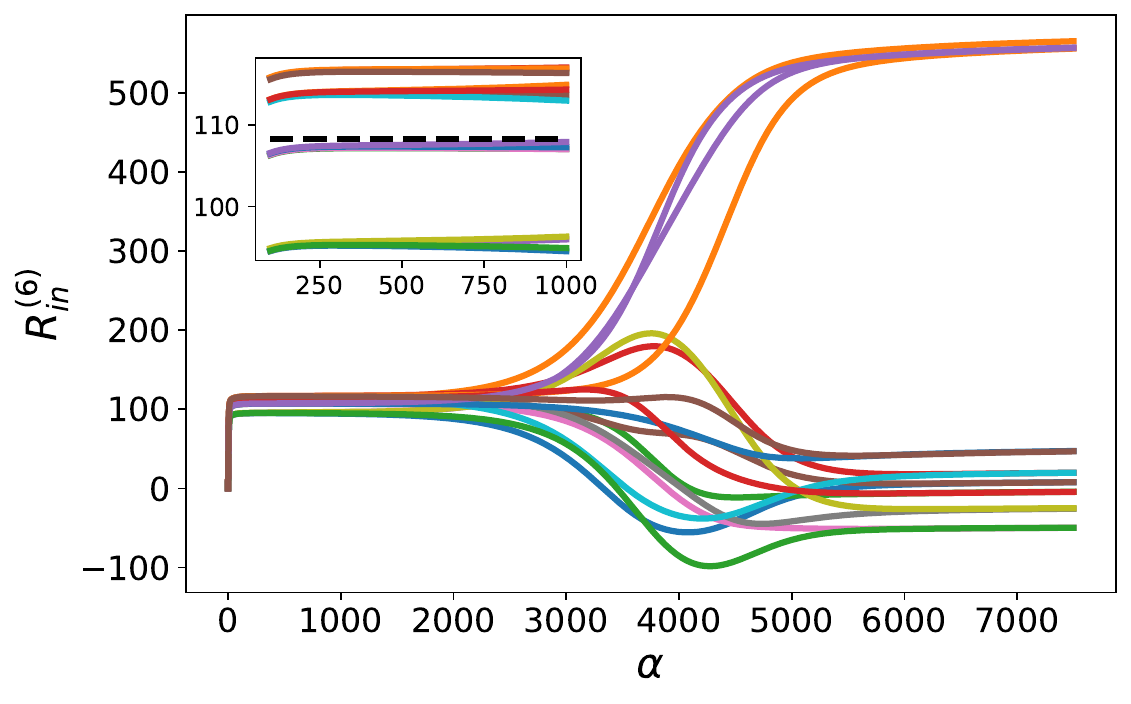}
  \includegraphics[width=0.49\linewidth]{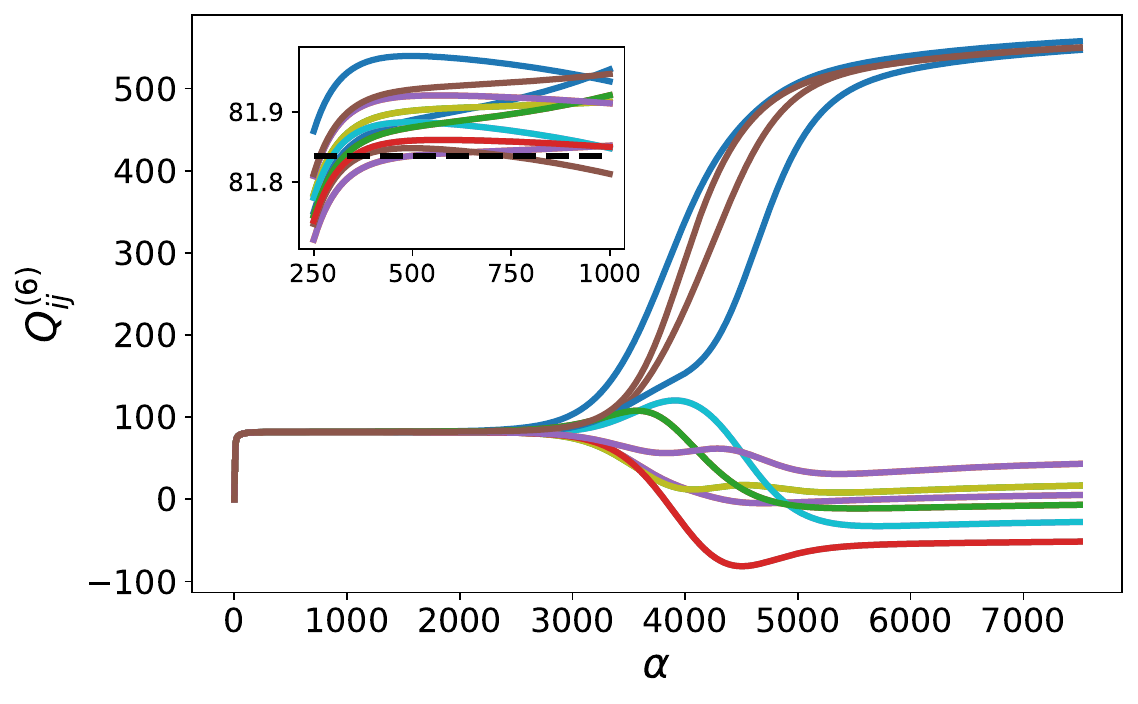}
  \caption{Plateau behavior of the order parameters with $L=10$, $N=7000$, $\sigma_{J}=0.01$, $\eta=0.1$ and $M=K=4$ for one random initialization of student and teacher vectors. We solve the differential equations for the small learning rate case where we consider terms up to $\mathcal{O}\left(\eta\right)$. The inset shows the higher-order order parameters at the plateau. For the student-teacher order parameters, we obtain $M$ different plateau heights. For the student-student order parameters, we observe one plateau height with small statistical deviations for the particular matrix entry $Q_{ij}^{(l)}$. The dashed horizontal lines in the insets show the plateau heights according to Eq. (\ref{plateau fix l appendix}).}
  \label{fig: plateau height higher order appendix}
\end{figure}

Here, we consider the differential equations given in Eq. (\ref{odes non appenix}) for the higher-order order parameters up to the first order in the learning rate $\mathcal{O}\left(\eta\right)$. As already mentioned in the main text, the order parameters are no longer self-averaging, i.e. we cannot replace the random variables $T_{nm}^{(l)}$ by their expectation value in the thermodynamic limit $N \to \infty$. However, without any assumptions on the teacher-teacher order parameters, we find approximately the following fixed point for $l=1$
\begin{align}
R_{in}^{*^{(1)}} \approx \frac{1}{\sqrt{\frac{M}{T_{nn}^{(1)}+d^{(1)}_n}\left(\frac{M T_{nn}^{(1)}}{T_{nn}^{(1)}+d^{(1)}_n} \left(1+\frac{1}{T_{nn}^{(1)}}\right)-1\right)}}  , && Q^{*^{(1)}} \approx \frac{1}{M} \sum_n^M \frac{1}{\frac{M T_{nn}^{(1)}}{T_{nn}^{(1)}+d^{(1)}_n} \left(1+\frac{1}{T_{nn}^{(1)}}\right)-1} 
\label{plateau fix approx appendix}
\end{align}
and for $l \neq 1$
\begin{align}
R_{in}^{*^{(l)}} \approx \frac{T_{nn}^{(l)}}{\sqrt{\frac{M}{T_{nn}^{(1)}+d^{(1)}_n}\left(\frac{M T_{nn}^{(1)}}{T_{nn}^{(1)}+d^{(1)}_n} \left(1+\frac{1}{T_{nn}^{(1)}}\right)-1\right)}}  , && Q^{*^{(l)}} \approx \frac{1}{M} \sum_n^M \frac{T_{nn}^{(l)}}{\frac{M T_{nn}^{(1)}}{T_{nn}^{(1)}+d^{(1)}_n} \left(1+\frac{1}{T_{nn}^{(1)}}\right)-1} ,
\label{plateau fix approx 2 appendix}
\end{align}
with $d^{(1)}_n= \sum_{m,m\neq n}^{M-1} T^{(1)}_{nm}$. Thus, for the student-teacher order parameters, we obtain $M$ different plateau heights for each order $l$ depending on the sum over the off-diagonal entries of the higher-order teacher-teacher order parameters $d^{(1)}_n$ and $T_{nn}^{(1)}$. This approximation is exact if all $T^{(l)}_{nm}=D^{(l)}$ for $n\neq m$ and $T_{nn}=T^{(l)}$ and all plateau heights are the same $R_{in}^{*^{(l)}} =R^{*^{(l)}}$. In Figure \ref{fig: plateau numerics vs approx}, we compare the numerically found generalization error by evaluating the differential equations given in Eq. (\ref{odes non appenix}) up to $\mathcal{O}\left(\eta\right)$ with the generalization error based on the approximately found fixed points given in Eq. (\ref{plateau fix approx appendix}). For small $L$, we find very good agreement between the true plateau and the approximation. \\
In order to proceed, we make the ansatz for the off-diagonal entries $T_{nm}^{(l)}=D^{(l)}=\frac{1}{M}\sum_{n}^M d^{(l)}_n$ for $n\neq m$. This approach preserves the statistical properties of the sum represented by a single parameter $D^{(l)}$. Moreover, for $l=0$, the teacher-teacher order parameters are still self-averaging in the thermodynamic limit, and we can assume $T_{nn}^{(0)}=1$ and $T_{nm}^{(0)}=0$ for large $N$. Thus in summary, we assume $ T_{nm}^{(0)}   = \delta_{nm}$ and $ T_{nm}^{(l)}   = \delta_{nm} T^{(l)}+(1 - \delta_{nm})D^{(l)}$ for $l\neq 0$. After these assumptions, we find new stationary points for our differential equations for $l=1$
\begin{align}
R^{*^{(1)}} =\frac{1}{\sqrt{\frac{M}{T^{(1)}+D^{(1)}}\left(\frac{M T^{(1)}}{T^{(1)}+D^{(1)}} \left(1+\frac{1}{T^{(1)}}\right)-1\right)}}  , && Q^{*^{(1)}} =  \frac{1}{\frac{ MT^{(1)}}{T^{(1)}+D^{(1)}} \left(1+\frac{1}{T^{(1)}}\right)-1} ,
\label{plateau fix approx 1}
\end{align}
and $l \neq 1$
\begin{align}
R^{*^{(0)}} &=\frac{1}{T^{(1)}+D^{(1)}}\frac{1}{\sqrt{\frac{M}{T^{(1)}+D^{(1)}}\left(\frac{M T^{(1)}}{T^{(1)}+D^{(1)}} \left(1+\frac{1}{T^{(1)}}\right)-1\right)}}  , \nonumber \\
Q^{*^{(0)}} &=  \frac{1}{T^{(1)}+D^{(1)}}  \frac{1}{\frac{ MT^{(1)}}{T^{(1)}+D^{(1)}} \left(1+\frac{1}{T^{(1)}}\right)-1} \nonumber \\
R^{*^{(l)}} &=\frac{T^{(l)}}{T^{(1)}}\frac{1}{\sqrt{\frac{M}{T^{(1)}+D^{(1)}}\left(\frac{M T^{(1)}}{T^{(1)}+D^{(1)}} \left(1+\frac{1}{T^{(1)}}\right)-1\right)}}  ,\nonumber \\
Q^{*^{(l)}} &=  \frac{T^{(l)}}{T^{(1)}}\frac{1}{\frac{ MT^{(1)}}{T^{(1)}+D^{(1)}} \left(1+\frac{1}{T^{(1)}}\right)-1}.
\label{plateau fix l appendix}
\end{align}
As one can see in Eqs. (\ref{plateau fix approx 1}), and (\ref{plateau fix l appendix}), we end up in one plateau height for the order parameters $Q_{ij}^{(l)} = Q^{*^{(l)}}$ and $  R_{im}^{(l)} = R^{*^{(l)}}$. In addition to Figure \ref{fig: plateau height first order} given in the main text, we provide the plateau behavior for higher order-order parameters in Figure \ref{fig: plateau height higher order appendix} and compare our newly obtained stationary solutions given in Eq. (\ref{plateau fix l appendix}) with the solutions of the differential equations given in Eq. (\ref{odes non appenix}) up to $\mathcal{O}\left(\eta\right)$. We observe that the student-teacher order parameters defined in given in Eq. (\ref{plateau fix l appendix}) yield an approximation for the mean value of the $M$ groups of order parameters. For the student-student order parameters, we observe a small systematic error which appears to be small compared to the magnitude of $Q^{*^{(l)}}$.\\
Next, we insert Eq. (\ref{plateau fix approx 1}) into the expression of the generalization error and obtain
\begin{align}
\epsilon_g^* = \frac{1}{\pi} \Bigg( 
&M \arcsin\left( \frac{D^{(1)} + T^{(1)}}{M T^{(1)} + M} \right) 
+ (M - 1) \arcsin\left( \frac{D^{(1)}}{T^{(1)} + 1} \right) \nonumber \\
&- 2 M \arcsin\left( \frac{1}{\sqrt{T^{(1)} + 1} 
\sqrt{\frac{-(D^{(1)} M - M^2 - (M^2 - M) T^{(1)})}{(D^{(1)})^2 + 2 D^{(1)} T^{(1)} + (T^{(1)})^2}} 
\sqrt{\frac{M T^{(1)} + M}{(M - 1) T^{(1)} - D^{(1)} + M}}} \right) \nonumber \\
&+ \arcsin\left( \frac{T^{(1)}}{T^{(1)} + 1} \right)
\Bigg)
\label{eg plateau height appendix}
\end{align}
for the plateau height. The right panel in Figure \ref{fig: plateau height first order} shows an example for the estimated plateau height given by Equation (\ref{eg plateau height appendix}) against the numerical solution of the differential equations.

\subsection{Plateau escape}
\label{plateau escape appendix}
In this subsection, we want to present the escape from the plateau. The found stationary equations given by Eqs. (\ref{plateau fix approx 1}) and (\ref{plateau fix l appendix}) are unstable such that after a certain time on the plateau, the generalization error will eventually escape from it. In order to escape from the plateau, the unique solution of the fixed points must be broken for each $l$ at a certain time. We want to study the dynamics in the vicinity of the fixed point and clarify how the generalization error leaves it. For this, introduce parameters $S^{(l)}$ and $C^{(l)}$ indicating the onset of specialization for the student vectors towards one teacher vector. Therefore, parameterized the order parameters by $R_{im}^{(l)} = R^{(l)} \delta_{im} + S^{(l)} (1 - \delta_{im})$, $Q_{ij}^{(l)} = Q^{(l)} \delta_{ij} + C^{(l)} (1 - \delta_{ij})$. Moreover, to study the escape from the plateau, we introduce small perturbation 
 parameters $r^{(l)},s^{(l)},q^{(l)} $ and $c^{(l)}$ modeling the repelling characteristic of the unstable fixed point. Thus we parametrized the order parameters by their stationary solution and a small perturbation $ R^{(l)} = R^{*^{(l)}} + r^{(l)}$, $ S^{(l)} = S^{*^{(l)}} + s^{(l)}$,$ Q^{(l)} = Q^{*^{(l)}} + q^{(l)}$ and $ C^{(l)} = C^{*^{(l)}} + c^{(l)}$  with $S^{*^{(l)}} =R^{*^{(l)}}$ and $C^{*^{(l)}} =Q^{*^{(l)}}$. Next, we insert this parametrization into the differential equations given by Eq. (\ref{odes non appenix}) up to $\mathcal{O}\left(\eta\right)$. In order to study the dynamics in the vicinity of the fixed point, we linearized the dynamical equations in $\left(r^{(l)},s^{(l)},q^{(l)}, c^{(l)}\right)^\top$ around zero.\\
In the following, we present the differential equations, eigenvalues, and eigenvectors for the specific case where $D^{(l)} = 0$ and $T^{(1)} = 1$ for notational simplicity. The full system is too large to display in its entirety. However, we provide insights throughout on how these results generalize. After reducing the system of differential equations, we conclude by presenting the full solution for the general case.\\
After a first-order Taylor expansion in $\left(r^{(l)},s^{(l)},q^{(l)}, c^{(l)}\right)^\top$, we find the following linearized equation

\begin{align}
\frac{d}{d\alpha} \begin{pmatrix}
\bm{r}\\
\bm{s}\\
\bm{q}\\
\bm{c}
\end{pmatrix}=\frac{2}{\pi M} \sqrt{\frac{2M-1}{2M+1}} \bm{A}_{p} \begin{pmatrix}
\bm{r}\\
\bm{s}\\
\bm{q}\\
\bm{c}
\end{pmatrix}
\end{align}
with $r_i=r^{(i-1)}$, $s_i=s^{(i-1)}$, $q_i=q^{(i-1)}$, $c_i=c^{(i-1)}$ and $\bm{A}_{p} = \bm{A}+ \bm{B}$ with $\bm{A}_{\mathrm{p}} \in \mathbb{R}^{4L \times 4L} $. The individual matrices can be written as Kronecker products $
\bm{A} =
\bm{G}\otimes \bm{A}_1, \quad 
\bm{B} =
\bm{H}\otimes \bm{U} $
with $ \bm{U} = \bm{u} \bm{e}_2^\top$ and
\begin{align} 
 \bm{G}&=\begin{bmatrix}
1& M-1 & 0& 0 \\
1& M-1&0&0  \\
-2\sqrt{\frac{M}{2M-1}}&-2\sqrt{\frac{M}{2M-1}}  \left(M-1\right) &2&2 \left(M-1\right)  \\
-2\sqrt{\frac{M}{2M-1}}&-2\sqrt{\frac{M}{2M-1}}  \left(M-1\right) &2&2 \left(M-1\right) 
\end{bmatrix} ,\\[10pt]
 \bm{H} &=  
\begin{bmatrix}
\frac{-M}{\left(2M+1\right)\left(2M-1\right)} &\frac{-2M\left(M-1\right)}{\left(2M+1\right)\left(2M-1\right)} &\frac{1}{2}\frac{2M^2+3M-2}{\sqrt{M}\left(2M+1\right)\sqrt{2M-1}} &\frac{\left(M-1\right) \sqrt{2M-1}}{\sqrt{M}\left(2M+1\right)} \\
\frac{-2M}{\left(2M+1\right)\left(2M-1\right)}  & \frac{-M(2M-3)}{\left(2M+1\right)\left(2M-1\right)}  & \frac{1}{2}\frac{2M^2+3M-2}{\sqrt{M}\left(2M+1\right)\sqrt{2M-1}} & \frac{\left(M-1\right) \sqrt{2M-1}}{\sqrt{M}\left(2M+1\right)}  \\
\frac{-2\sqrt{M}}{\left(2M+1\right)\sqrt{2M-1}} & \frac{-2(M-1)\sqrt{M}}{(2M+1)\sqrt{2M-1}}  & \frac{M+2}{2M+1} &\frac{2(M-1)}{2M+1} \\
\frac{-2\sqrt{M}}{\left(2M+1\right)\sqrt{2M-1}} & \frac{-2(M-1)\sqrt{M}}{(2M+1)\sqrt{2M-1}}   & \frac{M+2}{2M+1} & \frac{2(M-1)}{2M+1}.
\end{bmatrix}
\end{align}

We obtain the following eigenvalues and eigenvectors for $\bm{G}$

\begin{align}
\lambda_{G,1} &= M, &\mathbf{v}_{G,1} &= \left(1, 1, \frac{2\sqrt{M}}{\sqrt{2M - 1}}, \frac{2\sqrt{M}}{\sqrt{2M - 1}}\right)^\top, \\
\lambda_{G,2} &= 2M, &\mathbf{v}_{G,2} &= \left(0, 0, 1, 1\right)^\top, \\
\lambda_{G,3} &= 0, &\mathbf{v}_{G,3} &= \left(1, -\frac{1}{M - 1}, 0, 0\right)^\top, \quad \mathbf{v}_{G,4} = \left(0, 0, 1, -\frac{1}{M - 1}\right)^\top,
\end{align}

and for $\bm{H}$

\begin{align}
\lambda_{H,1} &= \frac{M}{4M^2 - 1}, &\mathbf{v}_{H,1} &= \left(1, -\frac{1}{M - 1}, 0, 0\right)^\top ,\\
\lambda_{H,2} &= \frac{2M}{2M + 1}, &\mathbf{v}_{H,2} &= \left(1, 1, \frac{2\sqrt{M}}{\sqrt{2M - 1}}, \frac{2\sqrt{M}}{\sqrt{2M - 1}}\right)^\top, \\
\lambda_{H,3} &= 0, &\mathbf{v}_{H,3} &= \left(1, 1, 0, \frac{M^{3/2}\sqrt{2M - 1}}{2M^2 - 3M + 1}\right),  \mathbf{v}_{H,4} = \left(0, 0, 1, -\frac{1}{2} \frac{2M^2 + 3M - 2}{2M^2 - 3M + 1}\right)^\top.
\end{align}
Since $\bm{A}$ is of block matrix structure expressed by a Kronecker product, we obtain for its spectrum $\lambda_A = \lambda_G \lambda_{A_1}$ and corresponding eigenvectors $\bm{v}_{A} = \bm{v}_G \otimes \bm{v}_{A_1}$ for which we multiply each eigenvalue of $\bm{A}_1$ with each of $\bm{G}$. The same also applies for the eigenvalues and -vectors of $\bm{B}$: $\lambda_B = \lambda_G \lambda_{A_1}$ and corresponding eigenvectors $\bm{v}_{B} = \bm{v}_G \otimes \bm{v}_{A_1}$. The eigenvalues of $\bm{A}_1$ were already studied in Subsection \ref{Solution of order parameters} and are the negative eigenvalues of the data covariance matrix $-\lambda_k$ with eigenvectors $\bm{v}_k$ summarized by the matrix $\bm{V}$ (cf. Eq. (\ref{Vandermonde eigenmatrix})). Since  $\bm{A}_1$ possesses $L$ eigenvalues and -vectors, we obtain multiple groups of different eigenvalues and -vectors for $\bm{A}$ and $\bm{B}$. The eigenvalues and -vectors of $\bm{U}$ are also already known. We have one eigenvector $\bm{u}$ with eigenvalue $\lambda_u=T^{(2)}$ and $L-1$ eigenvectors $\bm{e}_l$ with zero eigenvalue for $l=1,3,4...,L$. Thereby $\bm{e}_l$ is the $l$th unit vector. In the following, the superscript for the eigenvalues and -vectors indicates the corresponding group. \\

 For $\bm{A}$, the first two groups of eigenvalue combinations $\lambda_{A,k}^{(1)} = -M \lambda_{k}$ with eigenvector $\bm{v}_{A,k}^{(1)}= \left(\bm{v}_k, \bm{v}_k, 2\bm{v}_k\frac{\sqrt{M}}{\sqrt{2M - 1}}, 2\bm{v}_k\frac{\sqrt{M}}{\sqrt{2M - 1}}\right)^\top$ and $\lambda_{A,k}^{(2)} = -2M \lambda_{k}$ with $\bm{v}_{A,k}^{(2)}=\left(0, 0, \bm{v}_k, \bm{v}_k\right)^\top$ are plateau attractive. Their corresponding eigenvalues are negative and their directions are against the breaking of order parameter symmetry. The latter fact can be seen that the first two entries of the eigenvectors and the last two are the same. This would drive the dynamics in the direction corresponding to $r^{(l)}= s^{(l)}$ and $q^{(l)}= c^{(l)}$ which is exactly the plateau condition. The third group of eigenvalue combinations $\lambda_{A,k}^{(3)} = 0$ with eigenvectors $\bm{v}_{A,k}^{(3)}=\left(\bm{v}_k, -\bm{v}_k\frac{1}{M - 1}, 0, 0\right)^\top$ and $\bm{v}_{A,k}^{(4)}=\left(0, 0, \bm{v}_k, -\bm{v}_k\frac{1}{M - 1}\right)^\top$ are neither attractive nor repelling. However, their directions indicate a symmetry breaking in the order parameters, at least for one group $r^{(l)} \neq s^{(l)}$ or $q^{(l)}\neq  c^{(l)}$. \\

For the matrix $\bm{B}$, we observe that $\lambda_{B,1}^{(1)} = 0$, with a total of $4L-2$ distinct eigenvectors. However, the more significant impact comes from the directions associated with non-zero eigenvalues. These eigenvalues play a crucial role in influencing the spectrum of $\bm{A}$, particularly when $\bm{B}$ is viewed as a non-negligible perturbation of $\bm{A}$. The second eigenvalue of $\bm{B}$ is $\lambda_{B,2} = \frac{M T^{(2)}}{4M^2 - 1}$ with eigenvector $ \bm{v}_{B,2}= \left(\bm{u}, -\bm{u} \frac{1}{M - 1}, 0, 0\right)^\top$. For the third one, we obtain 
$\lambda_{B,3}= \frac{2M T^{(2)}}{2M + 1}$ with eigenvector $\bm{v}_{B,3}= \left(\bm{u}, \bm{u}, \frac{2\sqrt{M}}{\sqrt{2M - 1}} \bm{u}, \frac{2\sqrt{M}}{\sqrt{2M - 1}} \bm{u}\right)^\top$. \\

In summary, we obtain two important directions for the escape of the plateau. The first one corresponds to the eigenvectors $\bm{v}_{A,k}^{(3)}$ and $ \bm{v}_{B,2}$ and the second one is in the direction of $ \bm{v}_{B,3}$ and $\bm{v}_{A,k}^{(1)}$. Note that these directions are also present for the sum of $\bm{A}+\bm{B}$ resulting in $\bm{A}_p$. Therefore, we make as a first ansatz $q^{(l)}=c^{(l)}$ since this condition is fulfilled for all important eigendirections. Moreover, we can make the following second ansatz $q^{(l)} =c^{(l)}= 2 R^{*^{(l)}} \left(r^{(l)}+(M-1)s^{(l)}\right) $. The second ansatz is fulfilled by both eigendirections as well. For $D^{(l)}\neq 0$ and general $T^{(l)}$, we find with similar steps the relation $q^{(l)} =c^{(l)}= \frac{2T^{(l)}}{{T^{(1)}+D^{(1)}}} R^{*^{(l)}} \left(r^{(l)}+(M-1)s^{(l)}\right) $. \\
Next, we re-parametrize the dynamical equations under the condition $q^{(l)} =c^{(l)}= 2 R^{*^{(l)}} \left(r^{(l)}+(M-1)s^{(l)}\right) $ and find

\begin{align}
\frac{d}{d\alpha} \begin{pmatrix}
\bm{r}\\
\bm{s}\\
\end{pmatrix}=\frac{2}{\pi M} \sqrt{\frac{2M-1}{2M+1}} \bm{A}_{rs} \begin{pmatrix}
\bm{r}\\
\bm{s}\\ 
\end{pmatrix},
\end{align}

with $\bm{A}_{rs} = \bm{A}+ \bm{B}$ and $\bm{A}_{\mathrm{rs}} \in \mathbb{R}^{2L \times 2L} $. 
The matrices $\bm{A}$ and $\bm{B}$ are re-defined as follows:
\begin{align}
\bm{A} &= \bm{G} \otimes \bm{A}_1, && 
\bm{B} = \bm{H} \otimes \bm{U}
\end{align}
with re-defined $\bm{G}$ and $\bm{H}$
\begin{align} 
\bm{H} &=  \frac{1}{\left(2M+1\right)\left(2M-1\right)}
\begin{bmatrix}
5M-3 & 4M^2-7M+3\\
\frac{4M^2-7M+3}{M-1}& 4M^2-6M+3
\end{bmatrix}, &&  \bm{G} = \begin{bmatrix}
1 & (M-1) \\
1 & (M-1)
\end{bmatrix}
\end{align}

For $\bm{A}$, the eigenvectors $\bm{v}_{A}$ are given by $\bm{v}_{A} = \bm{v}_G \otimes \bm{v}_{A_1}$, where $\lambda_{A_1,k} = -\lambda_k$ and $\bm{v}_{A_1,k} = \bm{v}_k$ for $k \in (1,\dots,L)$. The corresponding groups of eigenvalues are $\lambda_{A,k}^{(1)} = - M \lambda_k$, with eigenvectors $\bm{v}_{A,k}^{(1)} = \begin{pmatrix} \bm{v}_k, \bm{v}_k \end{pmatrix}^\top$. The second group is given by $\lambda_{A}^{(2)} = 0$ and the corresponding eigenvectors are $\bm{v}_{A,k}^{(2)} = \begin{pmatrix} \bm{v}_k, \frac{-\bm{v}_k}{M-1} \end{pmatrix}^\top$. \\
For $\bm{B}$, the first eigenvalue is $\lambda_{B,1} = \frac{MT^{(2)}}{(2M-1)(2M+1)}$, with the corresponding eigenvector $\bm{v}_{B,1} = \begin{pmatrix} \bm{u}, \frac{-\bm{u}}{M-1} \end{pmatrix}^\top$. The second eigenvalue is $\lambda_{B,2} = \frac{2MT^{(2)}}{2M+1}$, with eigenvector $\bm{v}_{B,2} = \begin{pmatrix} \bm{u}, \bm{u} \end{pmatrix}^\top$. Furthermore, we have multiple eigenvectors for the eigenvalue zero. For $m \in (1,3,4,\dots,L)$, we have $\lambda_{B,m} = 0$, with corresponding eigenvectors $\bm{v}_{B,m} = \begin{pmatrix} \bm{e}_m, \frac{-\bm{e}_m}{M-1} \end{pmatrix}^\top$. Similarly, for $n \in (1,3,4,\dots,L)$, $\lambda_{B,n} = 0$, with eigenvectors $\bm{v}_{B,n} = \begin{pmatrix} \bm{e}_n, \bm{e}_n \end{pmatrix}^\top$. \\
Note that all eigenvalues were already encountered for the larger system verifying our analytical ansatz. Moreover, the new eigenvectors are the first two entries of the eigenvectors of the large original system. \\
Due to the special structure of $\bm{A}$, the eigenvector $\bm{v}_{B,1}$ are also eigenvectors of $\bm{A}$, both associated with the eigenvalue $0$, meaning $\bm{A} v_{B,m} = 0$ and $\bm{A} v_{B,1} = 0$. Among the eigenvalues of $\bm{A}_{rs}$, $2L-2$ of them are zero. The first non-zero eigenvalue is $\lambda_{A_{rs},1} = \frac{MT^{(2)}}{(2M-1)(2M+1)}$ which is larger than zero indicating a repelling character for direction $ \bm{v}_{B,1} $ and $\bm{v}_{A,k}^{(2)}$. For the eigenvector $\bm{v}_{B,2}$, we have $\bm{A} \bm{v}_{B,2} = M \begin{pmatrix} \bm{A}_1 \bm{u}, \bm{A}_1 \bm{u} \end{pmatrix}^\top$. Therefore, $\bm{v}_{B,2}$ is an eigenvector of $\bm{A}$, provided that $\bm{u}$ is an eigenvector of $\bm{A}_1$. For the product, we find $\bm{A}_1 \bm{u} = - \bm{u}_+ $, with $\bm{u}_+ = \left( T^{(2)}, T^{(3)}, ..., T^{(L+1)}\right)^\top $. $\bm{A} \left(\bm{v}_{A,k}^{(1)}+\bm{v}_{B,2}\right) = -\lambda_k \begin{pmatrix} \bm{v}_k, \bm{v}_k\end{pmatrix}^\top - M \begin{pmatrix} \bm{u}_+, \bm{u}_+\end{pmatrix}^\top$ and this group of eigenvalues is therefore negative. \\
Finally, we obtain one important eigendirection showing an eigenvalue larger than zero. This direction corresponds to  $ \bm{v}_{B,1} $ and $\bm{v}_{A,k}^{(2)}$. Therefore, we make the following last ansatz $s^{(l)} = -\frac{r^{(l)}}{M-1}$ in order to reduce the system for a second time. Note that the exact same relation also holds for $D^{(l)}\neq 0$ and general $T^{(1)}$.\\

For the final form of the differential equations, we return to the case where $D^{(l)} \neq 0$ and $T^{(1)} \neq 1$, as the expressions are now more manageable to display and no longer excessively large. The final re-definition of the dynamical system yields
\begin{align}
\frac{d\bm{r}}{d\alpha} = \eta g_r \tilde{\bm{U}}\bm{r},
\end{align}
with $g_r= \frac{2}{\pi}\frac{ \left(D^{(1)}+T^{(1)}\right)}{\sqrt{(M-1)T^{(1)}-D^{(1)}+M} \left(D^{(1)}+ (M+1)T^{(1)}+M\right)^{\frac{3}{2}}} $, $\tilde{\bm{U}}= \tilde{\bm{u}} \bm{e}_2^\top  $ and $ \tilde{\bm{u}}= \left(T^{(1)}-\frac{D^{(1)}}{M-1}, T^{(2)}-\frac{D^{(2)}}{M-1},...,T^{(L)}-\frac{D^{(L)}}{M-1}\right)^\top$. Note that we define $\bm{A}_r=g_r \tilde{\bm{U}}$ for the main text. Since $\tilde{\bm{U}}$ is a rank-1 matrix, we obtain $L-1$ zero eigenvalues and one eigenvalue
\begin{align}
    \lambda_{r}=  T^{(2)}-\frac{D^{(2)}}{M-1}
\end{align} 
larger than zero. Thus, $\lambda_r$ drives the escape from the plateau. We can solve the differential equation directly and find for the first-order perturbation parameter
\begin{align}
r^{(1)} = e^{\eta g_r \alpha} \, r^{(1)}_0,
\end{align}
where $r^{(1)}_0 = r^{(1)}(\alpha_0)$ and $\alpha_0$ denotes an arbitrary time at the plateau. For the escape of the generalization error within our re-defined dynamical system, we find
\begin{align}
\epsilon_g^* - \epsilon_g &= \frac{\left((M-1)T^{(1)}+M-D^{(1)}\right)\left(D^{(1)}+T^{(1)}\right)}{8\pi M\left(M-1\right)\left(T^{(1)}-\frac{\left(D^{(1)}+T^{(1)}\right)^2}{4M^2}\right)^{\frac{3}{2}}}  e^{\frac{\alpha}{\tau_{\mathrm{esc}}}} \, r_0^{(1)^2}
\label{pl escape formula}
\end{align}
where we have introduced the escape time
\begin{align}
\tau_{\mathrm{esc}} &= \frac{1}{\eta g_r \lambda_{r}} \nonumber \\
&=  \frac{\pi}{2\eta}\frac {\sqrt{(M-1)T^{(1)}-D^{(1)}+M} \left(D^{(1)}+ (M+1)T^{(1)}+M\right)^{\frac{3}{2}}} { \left(T^{(2)}-\frac{D^{(2)}}{M-1}\right) \left(D^{(1)}+T^{(1)}\right)}.
\end{align}
Furthermore, we can approximate $T^{(2)}=\frac{1}{L}\operatorname{Tr}(\Sigma^2) = \frac{\lambda_+^2}{L}  \sum_i^L \frac{1}{i^{2(1+\beta)}} \approx \frac{\lambda_+^2}{L} \propto L$ for large $L$. The same applies to $D^{(2)}$. For large $M$ and $L$, we find $\tau_{\mathrm{esc}} \propto \frac{M^2}{\eta T^{(2)}}$.

\subsection{numerically estimated plateau length}
\label{plateau escape numerics appendix}
\begin{figure}
  \centering
  \includegraphics[width=0.49\linewidth]{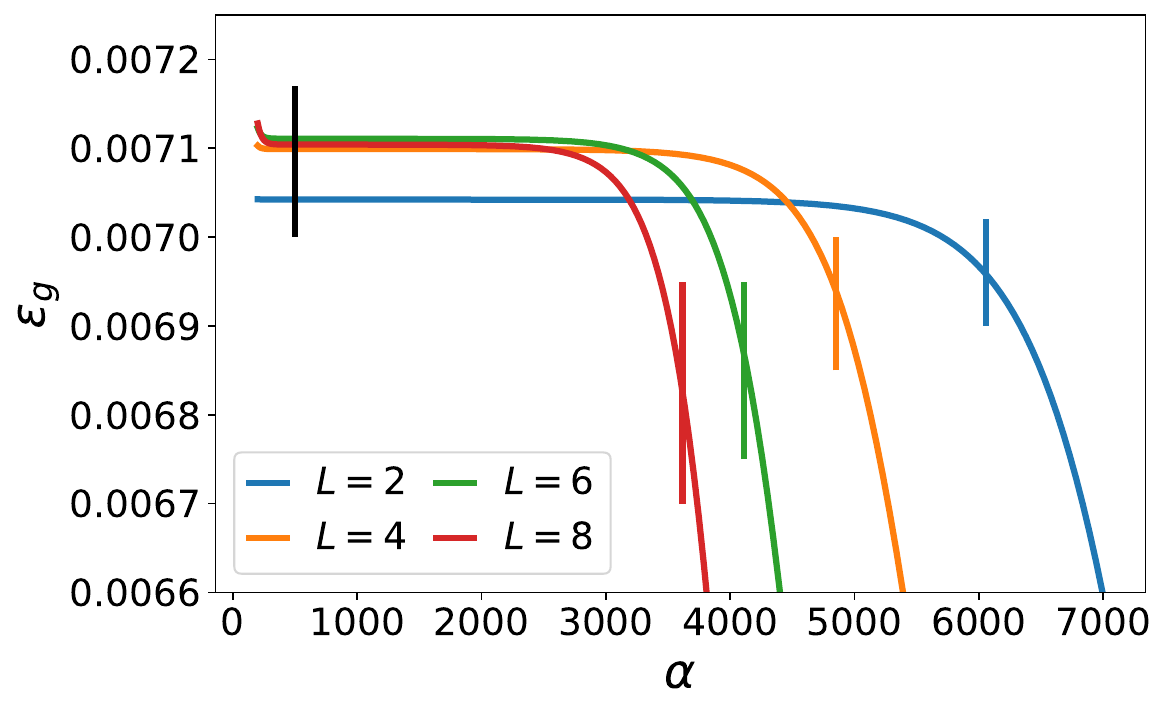}
  \includegraphics[width=0.49\linewidth]{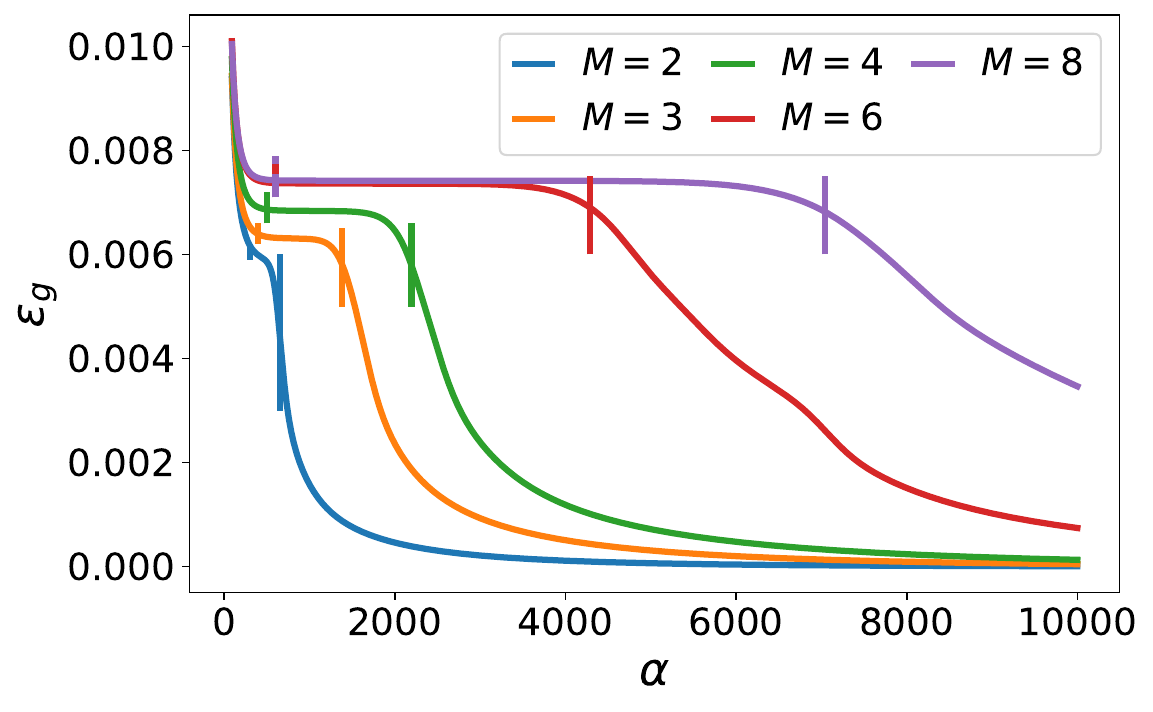}
  \caption{Plateau phase of the generalization error evaluated by numerical solutions of the differential equations for one random initialization (solid) and plateau length estimations given by Eq. (\ref{Biehl plateau 2}) (verticle lines) for $N=7000$, $K=M=4$, $\sigma_J=0.01$, $\beta=0.25$ and $\eta=0.1$. The black verticle line indicates the arbitrarily chosen plateau start $\alpha_0=500$, and the colored verticle lines show $\alpha_P$ for different $L$. We retain terms up to $\mathcal{O}(\eta)$ for the differential equations. }
  \label{fig: plateau length estimation}
\end{figure}

In this subsection, we demonstrate how to combine the analytically derived formula from Eq. (\ref{Biehl plateau}) with our calculated escape time (Eq. (\ref{escape time})) to estimate the plateau length. The plateau escape is described by the equation
\begin{align}
\alpha_P-\alpha_0= \tau_{\mathrm{esc}} \left(D-\frac{1}{2}\ln\left(\sigma_J^2\right)+\frac{1}{2}\ln\left(N\right)\right),
\label{Biehl plateau 2}
\end{align}
where $D$ is a constant of order $\mathcal{O}(1)$, dependent on the variance at initialization and the plateau; $\alpha_0$ is an arbitrary starting point on the plateau; and $\tau_{\mathrm{esc}}$ is the escape time from the plateau.
To estimate the constant $D$, we interpret the results of \cite{Biehl_1996} and find $D = \ln\left(\frac{B}{c}\right)$, where $B$ represents the deviation of the order parameter responsible for the plateau escape at $\alpha_P$ from its value at $\alpha_0$. Thereby, $c$ is a proportionality constant between the fluctuations at the plateau and at initialization. In our case, the order parameter that drives the plateau escape is the first-order student-teacher order parameter (cf. Subsection \ref{plateau escape appendix}). Thus, we define $B = \vert R^{(1)}(\alpha_0) - R^{(1)}(\alpha_P)\vert$. Additionally, to estimate $\alpha_P - \alpha_0$, we set $R^{(1)}(\alpha_P) = e R^{*(1)}$, following the definition of the escape time for the generalization error.  Next, we estimate $\sigma_{P} = c \sigma_J$, where the plateau variance $\sigma_{P}$ is derived from numerical simulations of $R^{*(1)}$ at the plateau. \\
For the example shown in Figure \ref{fig: plateau height first order}, we set $\alpha_0 = 300$ and find $\sigma_{P} \approx 0.000279$, with $c \approx 0.0279$ (since $\sigma_J = 0.01$ is given), $B \approx 0.33$, and $D \approx 2.47$. For the escape time, we find $\tau_{\mathrm{esc}} \approx 239$ by averaging the diagonal terms of $T_{nm}^{(1)}$ to obtain $T^{(1)}$ and using the averaged sum of the off-diagonal entries in order to estimate $D^{(1)}$ and $D^{(2)}$. Finally, we obtain $\alpha_P - \alpha_0 \approx 2751$.\\
This procedure provides valuable insight into how the plateau length behaves with respect to various parameters. Figure \ref{fig: plateau length estimation} shows the generalization error based on the solution of the differential equations and presents additional examples for the plateau length estimation.

\subsection{asymptotic solution}
\label{asymptotic appendix}

Here, we want to investigate how the generalization error converges to its asymptotic value in more detail. For this, we consider the typical teacher configuration $\langle T_{nm}^{(l)}\rangle = \delta_{nm} T^{(l)}$ since this configuration already captures the scaling behavior of the generalization error. For the asymptotic fixed points of the order parameters, we find $R_{im}^{(l)}=T_{im}^{(l)} \delta_{in}$ and $ Q_{ij}^{(l)}=T_{ij}^{(l)} \delta_{ij}$. Here, we distinguish again between diagonal and off-diagonal entries for $R_{im}^{(l)}=R^{(l)}\delta_{im}+S^{(l)}(1-\delta_{im})$ and $Q_{ij}^{(l)}=Q^{(l)}\delta_{ij}+C^{(l)}(1-\delta_{ij})$ as for the plateau case. Furthermore, we linearized the dynamical equations for small perturbation around its fixed point $R^{(l)}=T^{(l)}+r^{(l)}$, $S^{(l)}=T^{(l)}+s^{(l)}$, $Q^{(l)}=T^{(l)}+q^{(l)}$, and $C^{(l)}=T^{(l)}+c^{(l)}$. \\
We find the following dynamic equations
\begin{align}
\frac{d}{d\alpha} \begin{pmatrix}
\bm{r}\\
\bm{s}\\
\bm{q}\\
\bm{c}
\end{pmatrix}=\frac{2\sqrt{3}}{3\pi M} \bm{A}_{a} \begin{pmatrix}
\bm{r}\\
\bm{s}\\
\bm{q}\\
\bm{c}
\end{pmatrix}
\end{align}
with $r_i=r^{(i-1)}$, $s_i=s^{(i-1)}$, $q_i=q^{(i-1)}$, $c_i=c^{(i-1)}$ and $\bm{A}_{a} = \tilde{\bm{A}}+ \tilde{\bm{B}}+\tilde{g} \tilde{\bm{C}}$ and $\tilde{g}= \eta\frac{2\sqrt{3}\left(\sqrt{45}+5\left(M-1\right)\right)}{15\pi M}$. The individual matrices can be written as Kronecker products
$
\tilde{\bm{A}} =
\bm{G}\otimes \bm{A}_1, \quad 
\tilde{\bm{B}} =
\bm{H}\otimes \bm{U},\quad 
\tilde{\bm{C}} =
\bm{F}\otimes \bm{U}
$
with
\begin{align} 
 \bm{G}&=\begin{bmatrix}
1& \frac{\sqrt{3}}{2} & 0& 0 \\
\frac{\sqrt{3}}{2}& 1&0&0  \\
-2& -\sqrt{3}&2&\sqrt{3}  \\
-\sqrt{3}& -2&\sqrt{3}&2  
\end{bmatrix} , \quad
 \bm{H} =  
\begin{bmatrix}
-\frac{1}{3}& -\frac{\sqrt{3}}{4} &\frac{1}{2} & \frac{\sqrt{3}}{4} \\
0& 0 & \frac{\sqrt{3}}{8} & 0 \\
-\frac{2}{3}& -\frac{\sqrt{3}}{2} &1 & \frac{\sqrt{3}}{2} \\
0& 0 & \frac{\sqrt{3}}{4} & 0 
\end{bmatrix}, \\[5pt]
\bm{F}& = \begin{bmatrix}
0 & 0 & 0 & 0 \\[2ex]
0 & 0 & 0 & 0 \\[2ex]
-2 & -\dfrac{f_1 \, f_3}{\sqrt{3}} & 1 & \dfrac{f_1 \, f_3}{2\sqrt{3}} \\[2ex]
-\dfrac{f_1 \, f_4}{\sqrt{3}} & -\dfrac{f_5 \, f_6 }{2} & \dfrac{f_1 \, f_4}{2\sqrt{3}} & \dfrac{f_5 \, f_6 }{4}
\end{bmatrix}
\end{align}
where
$f_1 = \sqrt{6} + M - 2 ,f_2 = \sqrt{45} + 5(M - 1) ,f_3 = \dfrac{15(M - 1)}{b} ,f_4 = \dfrac{15}{b} , f_5 = 4(M - 2)\sqrt{6} + 3M^2 - 15M + 26 ,f_6 = \dfrac{5}{b}$. Thereby, $\bm{A}_1$ and $\bm{U}_1$ are the same matrices as for the linear activation function case. Therefore, the linearized version of the dynamical equation for the non-linear activation function resembles the dynamical equation for the linear activation. However, we encounter an additional "perturbation" by $\tilde{\bm{B}}$, whereas $\tilde{\bm{C}}$ describes the influence of higher-order terms in the learning learning rate. Moreover, compared to the linear case, the differential equation has more additional variables due to correlations between different student and teacher vectors.  In order to analyze the behavior of the dynamical system, we need to determine the eigenvalues and eigenvectors of all sub-matrices. Here, we analyze the system for first order in the learning rate $\mathcal{O}\left(\eta\right)$ neglecting the contribution by $\tilde{g} \tilde{\bm{C}}$.\\

The eigenvalues of $\bm{G}$ are $\lambda_{G,1}=2-\sqrt{3} $ with eigenvector $\bm{v}_{G,1}=(0, 0, 1, -1)^\top$, $\lambda_{G,2}= 2+\sqrt{3} $ with eigenvector $\bm{v}_{G,2}=(0, 0, 1, 1)^\top$, $\lambda_{G,3}=1-\frac{1}{2} \sqrt{3} $ with eigenvector $\bm{v}_{G,3}=(1, -1, 2, -2)^\top$, $\lambda_{G,4}= 1+\frac{1}{2} \sqrt{3} $ with eigenvector $\bm{v}_{G,4}=(1, 1, 2, 2)^\top$. The eigenvalues of $\bm{A}_1$ were already studied in  Subsection \ref{Solution of order parameters} and are the negative eigenvalues of the data covariance matrix $-\lambda_k$ with eigenvectors $\bm{v}_k$ summarized by the matrix $\bm{V}$ (cf. Eq. (\ref{Vandermonde eigenmatrix})). Since $\tilde{\bm{A} }$ is of block matrix structure expressed by a Kronecker product, we obtain for its spectrum $\lambda_{\tilde{A}} = \lambda_G \lambda_{A_1}$ and corresponding eigenvectors $\bm{v}_{\tilde{A}} = \bm{v}_G \otimes \bm{v}_{A_1}$ for which we multiply each eigenvalue of $\bm{A}_1$ with each of $\bm{G}$. Thus, we obtain four different groups of eigenvalues for $\tilde{\bm{A}}$ and in total $4L$ eigenvalues. The first group is obtained by multiplying the first eigenvalue of $\bm{G}$ with all eigenvalues of $\bm{A}_1$ leading to $\lambda_{\tilde{{A}},k}^{(1)}=-\left(2-\sqrt{3}\right) \lambda_k$ with eigenvector $\bm{v}_{\tilde{{A}},k}^{(1)}=(0, 0, \bm{v}_k, -\bm{v}_k)^\top$. With the same procedure, we obtain for the other groups $\lambda_{\tilde{{A}},k}^{(2)}=-\left(2+\sqrt{3}\right) \lambda_k$ with eigenvector $\bm{v}_{\tilde{{A}},k}^{(2)}=(0, 0, \bm{v}_k, \bm{v}_k)^\top$, $\lambda_{\tilde{{A}},k}^{(3)}=\left(1-\frac{1}{2} \sqrt{3}\right) \lambda_k$ with eigenvector $\bm{v}_{\tilde{{A}},k}^{(3)}=(\bm{v}_k, -\bm{v}_k, 2\bm{v}_k, -2\bm{v}_k)^\top$, and $\lambda_{\tilde{{A}},k}^{(4)}=-\left(1+\frac{1}{2} \sqrt{3}\right) \lambda_k$ with eigenvector $\bm{v}_{\tilde{{A}},k}^{(4)}=(\bm{v}_k, \bm{v}_k, 2\bm{v}_k, 2\bm{v}_k)^\top$. The upper index for the eigenvalues and -vectors indicates the corresponding group. \\

The eigenvalues of $\bm{H}$ are $\lambda_{H,1}= \frac{1}{3}-\frac{\sqrt{43}}{12}$ with eigenvector $\bm{v}_{H,1}=\left(1,- \frac{\sqrt{3}}{9} \left(\sqrt{43}+4 \right),2, - \frac{2\sqrt{3}}{9} \left(\sqrt{43}+4 \right) \right)^\top$, $\lambda_{H,2}= \frac{1}{3}+\frac{\sqrt{43}}{12}$ with eigenvector $\bm{v}_{H,2}=\left(1,\frac{\sqrt{3}}{9} \left(\sqrt{43}+4 \right),2, \frac{2\sqrt{3}}{9} \left(\sqrt{43}+4 \right) \right)^\top$, and $\lambda_{H,3}= 0$ with eigenvectors $\bm{v}_{H,3}=\left(1,0,0, \frac{2\sqrt{3}}{9}\right)^\top$ and $\bm{v}_{H,4}=\left(0,1,0, 1\right)^\top$. For the matrix $\bm{U}$, we have just one eigenvalue distinct from zero $\lambda_{U,1}=T^{(2)}$ with eigenvector $\bm{u}$ since $\bm{U}$ has rank one. The remaining eigenvectors are given by $\bm{v}_{U,l}=\bm{e}_l$ for $l \in (1,3,4,...,L)$ and especially $ l \neq 2$. Thus, we obtain two different eigenvalues distinct from zero and $4L-2$ zero eigenvalues for $\tilde{\bm{B}}$. The eigenvalues distinct from zero are $\lambda_{\tilde{{B}},1}=\left(\frac{1}{3}-\frac{\sqrt{43}}{12} \right)T^{(2)}$ and $\bm{v}_{\tilde{{B}},1}=\left(\bm{u},- \frac{\sqrt{3}}{9} \left(\sqrt{43}+4 \right) \bm{u},2\bm{u}, - \frac{2\sqrt{3}}{9} \left(\sqrt{43}+4 \right) \bm{u}\right)^\top$ and $\lambda_{\tilde{{B}},2}=\left(\frac{1}{3}+\frac{\sqrt{43}}{12} \right)T^{(2)}$ and $\bm{v}_{\tilde{{B}},2}=\left(\bm{u}, \frac{\sqrt{3}}{9} \left(\sqrt{43}+4 \right) \bm{u},2\bm{u}, \frac{2\sqrt{3}}{9} \left(\sqrt{43}+4 \right) \bm{u}\right)^\top$. Then, we have two eigenvectors with zero eigenvalue $\bm{v}_{\tilde{{B}},3}=\left(\bm{u},0,0, \frac{2\sqrt{3}}{9} \bm{u}\right)^\top$ and $\bm{v}_{\tilde{{B}},4}=\left(0,\bm{u},0, \bm{u}\right)^\top$. Further eigenvectors have the structure $\bm{e}_{l} \otimes \bm{v}_{H,1} $, $\bm{e}_{l} \otimes \bm{v}_{H,2} $, $\bm{e}_{l} \otimes \bm{v}_{H,3} $ and $\bm{e}_{l} \otimes \bm{v}_{H,4} $.\\

\begin{figure}
  \centering
  \includegraphics[width=0.329\linewidth]{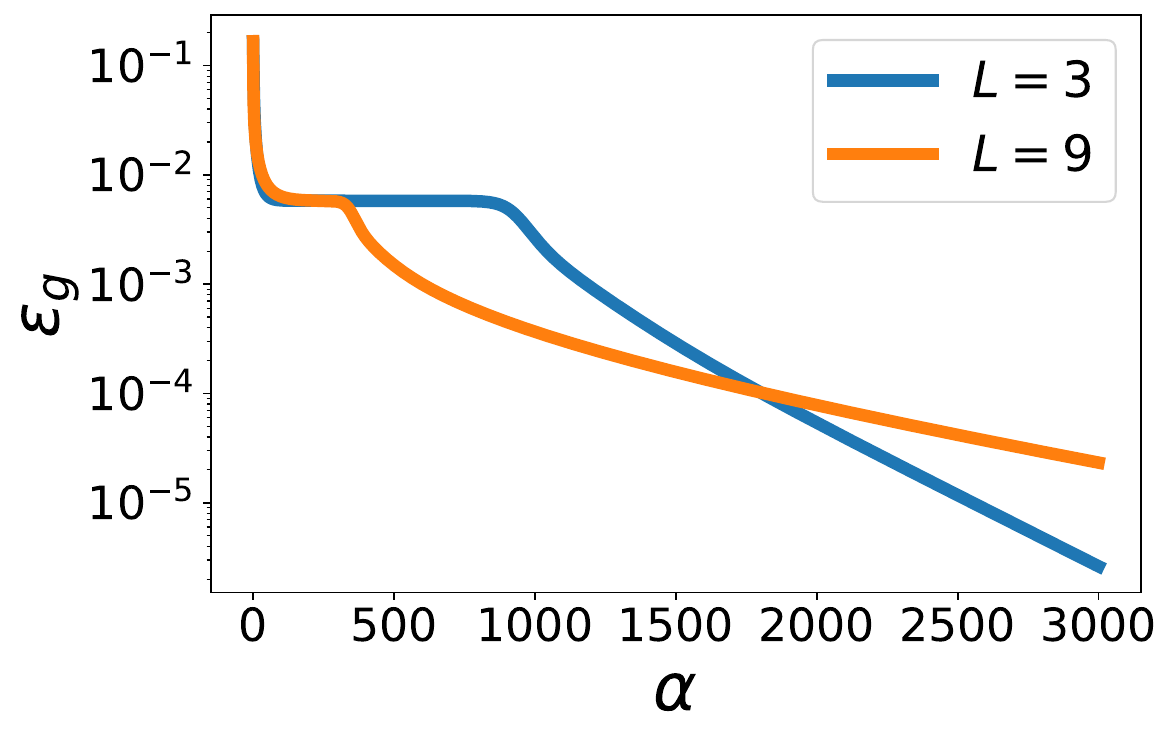}
  \includegraphics[width=0.329\linewidth]{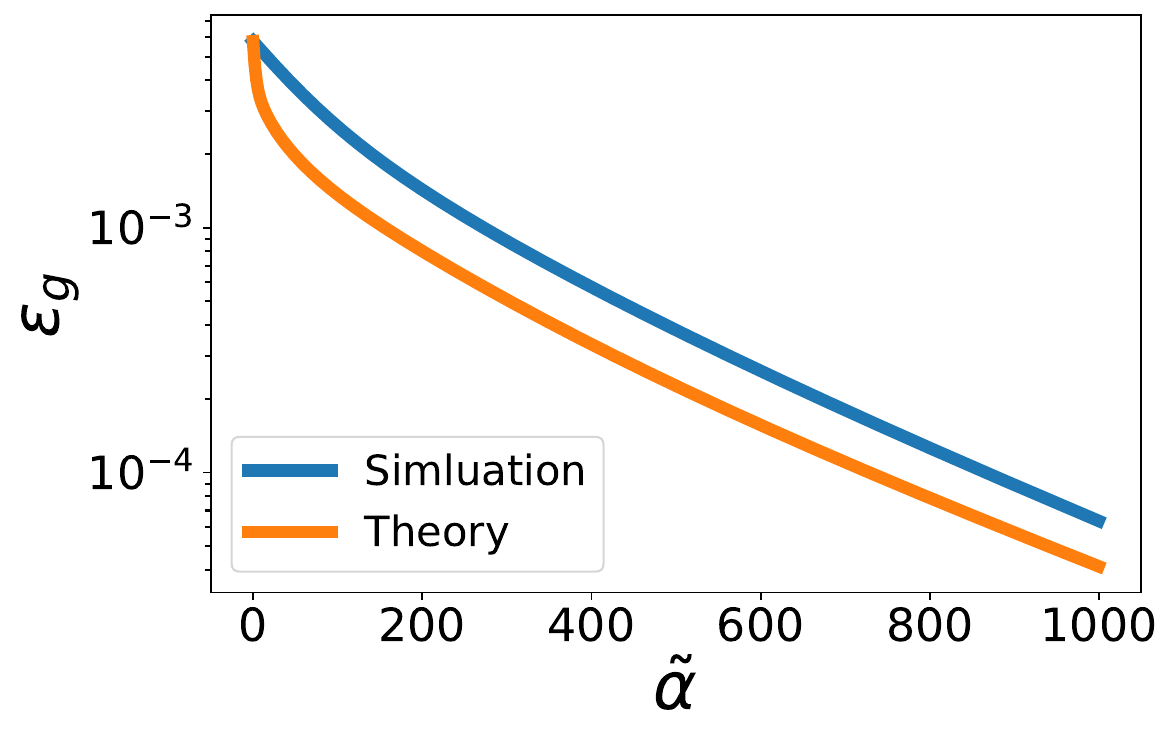}
  \includegraphics[width=0.329\linewidth]{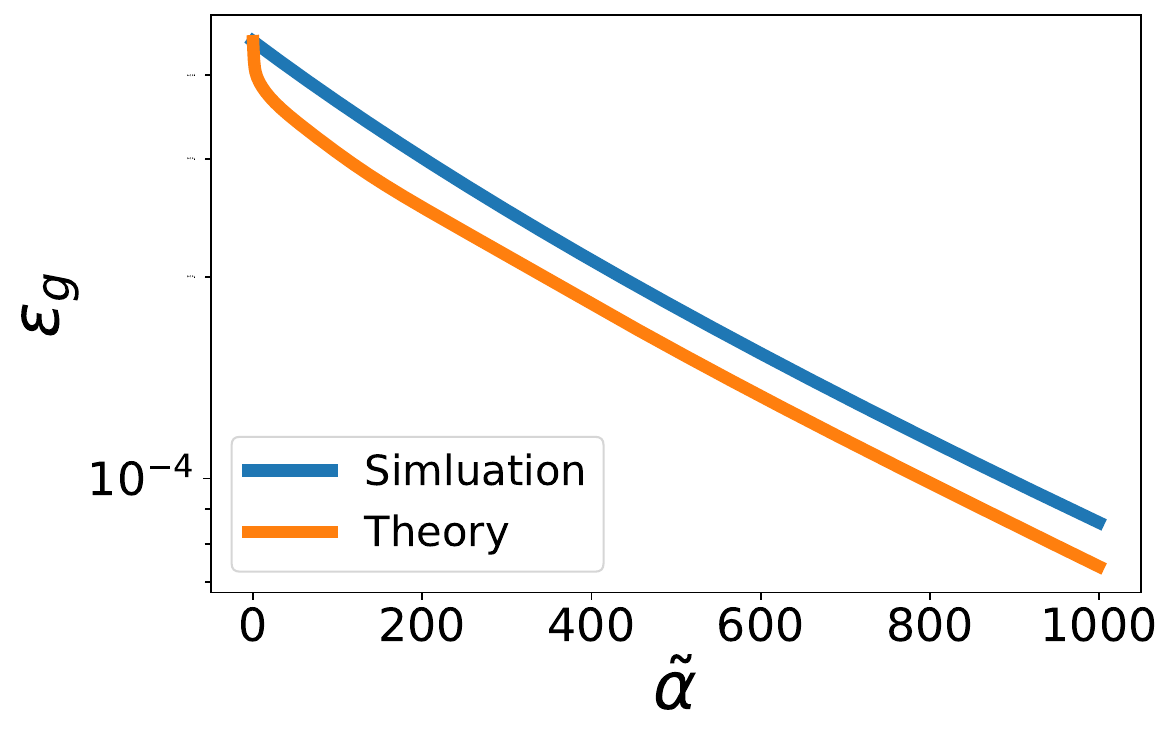}
  \caption{Left: Generalization error as a function of $\alpha$ obtained via the solution of differential equations of the order parameters up to $\mathcal{O}(\eta)$ and $K=M=2$, $\eta=0.25$, $\sigma_J=0.01$, $N=9000$ and $\beta=1$. Center and Right: Comparison of the first-order Taylor expansion of the generalization error based on Eq. (\ref{first order eg rsqc}) for numerically obtained $r^{(1)}, s^{(1)}, q^{(1)}$ and $c^{(1)}$ (Simulation) with Eq. (\ref{first order eg appendix}) (Theory) where the initial conditions $r^{(1)}(\alpha_0), s^{(1)}(\alpha_0), q^{(1)}(\alpha_0)$ and $c^{(1)}(\alpha_0)$ are obtained from simulations. Thereby, we choose $\alpha_0=1000$ and parameterize $\tilde{\alpha}=\alpha-\alpha_0$. Center: $L=3$. Right: $L=9$. }
  \label{fig: asymp numerics vs approx}
\end{figure}

Strictly speaking, none of the eigenvectors of $\tilde{\bm{A}}$ and $\tilde{\bm{B}}$ are the same and we cannot calculate the spectrum of their sum directly. However, we notice that $\tilde{\bm{B}}$ has just two eigenvalues distinct from zero and that its corresponding eigenvectors $\bm{v}_{\tilde{{B}},1}$ and $\bm{v}_{\tilde{{B}},2}$ are of the same structure as the last two groups of $\tilde{\bm{A}}$ namely $\bm{v}_{\tilde{{A}}}^{(3)}$ and $\bm{v}_{\tilde{{A}}}^{(4)}$. For each of the vectors, the third and the fourth components are twice as large as the first and the second one. Therefore, only the eigenvalues of the last two groups of $\tilde{\bm{A}}$ are influenced by adding the matrix $\tilde{\bm{B}}$ leading to the eigenvectors of $\bm{A}_a$ with the structure $\bm{v}_{A_{a},k}^{(3)} = \left(\bm{z}_k^{(3)},\bm{w}_k^{(3)},2\bm{z}_k,2\bm{w}_k^{(3)}\right)^\top$ and $\bm{v}_{A_{a},k}^{(4)} = \left(\bm{z}_k^{(4)}, \bm{w}_k^{(4)}, 2\bm{z}_k^{(4)}, 2\bm{w}^{(4)}_k\right)^\top$ with vectors $\bm{z}_k^{(3)}$, $\bm{w}_k^{(3)}$, $\bm{z}_k^{(4)}$ and $\bm{w}_k^{(4)}$ that has to be determined. Moreover, since the other eigenvalues of $\tilde{\bm{B}}$ are zero, the eigenvalues of the first and second group $\lambda_{\tilde{{A}},k}^{(1)}$ and $\lambda_{\tilde{{A}},k}^{(2)}$ remain the same for $\bm{A}_a$ as for $\tilde{\bm{A}}$. However, the corresponding eigenvectors are no longer analytically determinable and we have to rely on numerical solutions. All these claims for the spectra and eigenvectors are in excellent agreement with numerical experiments. \\
Next, we Taylor expand the generalization error up to first order in the small perturbation parameters $r^{(1)},s^{(1)},q^{(1)}$ and $c^{(1)}$. We find
\begin{align}
\epsilon_g &= \frac{1}{6\pi}  \left( 2\sqrt{3}q^{(1)} -4\sqrt{3}r^{(1)} +3 \left(M-1\right)c^{(1)} - 6\left(M-1\right) s^{(1)}  \right).
\label{first order eg rsqc}
\end{align}
From this expansion, we observe that the eigen-directions $\bm{v}_{A_{a},k}^{(3)}$ and $\bm{v}_{A_{a},k}^{(4)}$ do not contribute to the generalization error in first-order since their components cancel out. After inserting the expressions for the first and second groups of eigenvectors, we obtain
\begin{align}
\epsilon_g &= \frac{1}{6\pi}\sum_{k=1}^N g_k^{(1)}  e^{-a\left(2-\sqrt{3}\right)\lambda_k \alpha} \left( 2\sqrt{3}{v}^{(1)}_{k,2L+2} -4\sqrt{3}{v}^{(1)}_{k,2} +3\left(M-1\right){v}^{(1)}_{k,3L+2} -6\left(M-1\right){v}^{(1)}_{k,L+2}  \right) \nonumber \\
&+g_k^{(2)} e^{-a\left(2+\sqrt{3}\right)\lambda_k \alpha}  \left( 2\sqrt{3}{v}^{(2)}_{k,2L+2} -4\sqrt{3}{v}^{(2)}_{k,2} +3\left(M-1\right){v}^{(2)}_{k,3L+2} -6\left(M-1\right){v}^{(2)}_{k,L+2}  \right),
\label{first order eg appendix}
\end{align}
where $\bm{{v}_k}^{(1)}, \bm{{v}_k}^{(2)} \in \mathbb{R}^{4L}$ are the eigenvectors of $\bm{A}_{a}$ to the eigenvalues $-\left(2-\sqrt{3}\right)\lambda_k$ and $-\left(2+\sqrt{3}\right)\lambda_k$, respectively. Thereby, $g_k^{(1)}=\sum _l^{4L} \left({\left(\bm{V}^{(1)}\right)}^{-1}\right)_{kl} x_l$, $g_k^{(2)}=\sum _l^{4L} \left({\left(\bm{V}^{(2)}\right)}^{-1}\right)_{kl} x_l$ where $\bm{V}^{(1)}$ and $\bm{V}^{(2)}$ containing the first and second group of eigenvectors $\bm{{v}_k}^{(1)}$ and $\bm{{v}_k}^{(2)}$, respectively and $\bm{x}= \left(\bm{r}(\alpha_0), \bm{s}(\alpha_0), \bm{q}(\alpha_0), \bm{c}(\alpha_0) \right)^\top \in \mathbb{R}^{4L}$ is some reference point at arbitrary chosen $\alpha_0$ in the asymptotic phase.\\
Figure \ref{fig: asymp numerics vs approx} compares the generalization error derived from the first-order Taylor expansion in Eq. (\ref{first order eg rsqc}) where we obtain the parameters $r,q,s,c$ by solutions of the differential equations and our theoretical results based on Eq. (\ref{first order eg appendix}). For both, we use the same initial conditions, with $\alpha_0 = 1000$ as in the numerical solutions. The comparison shows excellent agreement. The small discrepancies between the graphs arise from the arbitrariness of $\alpha_0$ and the chosen initial conditions. Similar to linear activation functions, we observe a slowdown in convergence towards perfect generalization as $L$ increases, which eventually leads to a transition from exponential convergence to power-law scaling. We solve Eq. (\ref{first order eg appendix}) using the Julia programming language; however, we encounter limitations in increasing $L$ due to constraints in numerical precision (see Section \ref{remarks on numerics}).


\section{Remarks on numerical solutions}
\label{remarks on numerics}
Evaluating a large number of distinct eigenvalues becomes computationally challenging as $L$ increases. The expectation value of the teacher-teacher order parameters is given by $\left\langle T_{nm}^{(l)} \right\rangle = \delta_{nm} \frac{1}{N} \operatorname{Tr}(\Sigma^l)$. In this context, the highest order trace term in the differential equations is $L-1$, and for large $L$, we can approximate $\operatorname{Tr}(\Sigma^l)$ as $\lambda_+^{L-1} \sum_{i=1}^L \frac{1}{i^{(1+\beta)(L-1)}} \approx \left(\lambda_+\right)^{L-1}$. Since $\lambda_+$ scales with $L$ for large values of $L$, the expectation values increase in a 'super-exponential' manner with $L$. This growth also applies to the standard deviation of the off-diagonal entries of $T_{nm}^{(L-1)}$, further complicating numerical evaluations as $L$ grows.\\
As a result, numerical investigations are restricted to small values of $L$. This limitation applies to solving the differential equations, evaluating fixed points, and analyzing the generalization error in the asymptotic phase. For instance, to evaluate Eq. (\ref{first order eg}) and generate Figures \ref{fig: mult 2} through \ref{fig: eg0 vs eg 2}, we utilized Julia, a high-level scripting language, with arbitrary precision arithmetic.

\end{document}